\documentclass[a4paper,11pt]{article}

\usepackage{libertinus}        
\usepackage{microtype}         

\usepackage{geometry}
\usepackage{pdflscape}         
\usepackage{setspace}
\usepackage{amsmath, amssymb, amsthm}
\usepackage{mathrsfs}
\usepackage{bm}                
\usepackage{mathtools}         

\usepackage{graphicx}
\graphicspath{{./figures/}}
\usepackage{subcaption}        
\usepackage{wrapfig}           
\usepackage{float}             
\usepackage{rotating}          

\usepackage{tikz}
\usetikzlibrary{shapes,arrows,positioning,calc,patterns}

\usepackage{array, booktabs, multirow}
\usepackage{tabularx, longtable}
\usepackage{dcolumn}           
\usepackage{threeparttable}

\usepackage{enumitem}
\setlist{nosep}                

\usepackage{siunitx}
\DeclareSIUnit\bar{bar}

\usepackage{xcolor}
\definecolor{linkblue}{RGB}{0,102,204}

\usepackage[
   labelfont=bf,
   textfont=small,
   justification=justified,
   singlelinecheck=false,
   format=plain
]{caption}

\usepackage{lineno}

\usepackage[
   style=numeric-comp,
   citestyle=numeric-comp,
   sorting=none,
   backend=biber,
   giveninits=true,
   maxnames=5,
   minnames=3,
   date=year,
   doi=true,
   isbn=false,
   url=false
]{biblatex}
\usepackage[
   colorlinks=true,
   linkcolor=linkblue,
   citecolor=linkblue,
   urlcolor=linkblue,
   bookmarks=true,
   bookmarksnumbered=true,
   pdfstartview=FitH
]{hyperref}

\usepackage[capitalize,nameinlink]{cleveref} 
\usepackage{bookmark}                        

\hypersetup{
   pdftitle={Physics-Informed Neural Networks for Predicting Hydrogen Sorption in Geological Formations: Thermodynamically Constrained Deep Learning Integrating Classical Adsorption Theory},
   pdfauthor={Mohammad Nooraiepour, Mohammad Masoudi, Zezhang Song, Helge Hellevang},
   pdfsubject={Hydrogen Storage, Machine Learning, Geosciences},
   pdfkeywords={PINN, hydrogen sorption, machine learning, clays, shales, coals}
}

\usepackage{authblk}

\title{\Large Physics-Informed Neural Networks for Predicting Hydrogen Sorption in Geological Formations: Thermodynamically Constrained Deep Learning Integrating Classical Adsorption Theory}

\author[1,$\ast$]{Mohammad Nooraiepour}
\author[1,2]{Mohammad Masoudi}
\author[3]{Zezhang Song}
\author[1]{Helge Hellevang}

\affil[1]{%
   Faculty of Mathematics \& Natural Sciences, University of Oslo,
   P.O. Box 1047 Blindern, 0316 Oslo, Norway%
}
\affil[2]{%
   SINTEF Industry, Applied Geoscience Department,
   7465 Trondheim, Norway%
}
\affil[3]{%
   College of Geosciences, China University of Petroleum,
   Beijing 102249, China%
}
\affil[$\ast$]{%
   \textit{Corresponding author:} \href{mailto:mohammad.nooraiepour@geo.uio.no}{mohammad.nooraiepour@geo.uio.no}%
}

\date{}

\usepackage{comment}           
\usepackage{balance}           

\begin{document}

\maketitle

\begin{abstract}
Accurate prediction of hydrogen sorption in fine-grained geological materials
is essential for evaluating underground hydrogen storage capacity, assessing caprock integrity, and characterizing hydrogen migration in subsurface energy
systems. Classical isotherm models succeed at the individual sample level but
fail when generalized across heterogeneous populations, with
the coefficient of determination collapsing from 0.80--0.90 for single-sample
fits to 0.09--0.38 for aggregated multi-sample datasets, regardless of
functional form. We present a multi-scale physics-informed neural
network framework that addresses this limitation by embedding classical
adsorption theory and thermodynamic constraints directly into the learning
process. The framework utilizes 1,987 hydrogen sorption isotherm
measurements across three lithological classes (clays, shales, coals),
supplemented by 224 characteristic uptake measurements. A seven-category physics-informed feature engineering scheme generates 62 thermodynamically
meaningful descriptors from raw material characterization data. The loss
function enforces saturation limits, monotonic pressure response, and Van't
Hoff temperature dependence through penalty weighting, while a three-phase curriculum training strategy ensures stable integration of competing physical
constraints. An architecture-diverse ensemble of ten members provides
calibrated uncertainty quantification, with post-hoc temperature scaling
achieving target prediction interval coverage. The optimized PINN achieves
R$^2$ = 0.9544, RMSE = 0.0484 mmol/g, and MAE = 0.0231 mmol/g on the
held-out test set, with 98.6\% monotonicity satisfaction and zero non-physical
negative predictions. Physics-informed regularization yields a 10--15\%
cross-lithology generalization advantage over a well-tuned random forest under
leave-one-lithology-out validation, confirming that thermodynamic constraints
transfer meaningfully across geological boundaries. The framework is directly
extensible to competitive multicomponent sorption, cyclic injection-withdrawal
modeling, and integration with molecular-scale simulations, providing a
transferable computational foundation for accelerating the safe deployment of
underground hydrogen energy infrastructure.\\

\end{abstract}

\section{Introduction}
\label{sec:introduction}

The global transition to low-carbon energy systems represents one of the defining challenges of the 21st century, driven by climate change mitigation imperatives and the need to reduce fossil fuel dependence. Central to this transformation is the development of reliable, scalable energy storage technologies capable of addressing the intermittency inherent to renewable sources such as wind and solar \cite{LIEBENSTEINER2020, EnergyStorage2016}. Underground hydrogen storage (UHS) has emerged as a particularly promising approach for large-scale energy management, utilizing subsurface geological formations as gigawatt-scale repositories that enable both short- and long-term energy buffering, grid stabilization, and seasonal load balancing \cite{Heinemann2021,MASOUDI2024Lined,HEINEMANN2018,Hassanpouryouzband2021offshore}. By facilitating hydrogen's integration as a clean, high-energy-density carrier, UHS plays a critical role in global decarbonization pathways.

However, the efficacy of UHS, and indeed numerous other subsurface applications, depends fundamentally on understanding hydrogen-rock interactions at the fluid-mineral interface. These interactions govern processes across diverse geo-energy and geo-environmental contexts, including natural hydrogen exploration and in situ generation, hydrogen farming, enhanced geothermal systems, and the long-term containment of radioactive or hazardous waste \cite{Hassanpouryouzband2025NatHyd,ZGONNIK2020,hassanpouryouzband2025situ,Hassanpouryouzband2024RSE,Didier2012,BARDELLI2014168,MASOUDI2025661,TRUCHE2018186}. Among the key physicochemical processes at play, hydrogen sorption and desorption exert profound influence on system performance, yet remain inadequately characterized—particularly in fine-grained sedimentary formations.

Fine-grained rocks, including clay-rich mudstones, organic-rich shales, and coal seams, are ubiquitous in sedimentary basins and exhibit highly variable petrophysical properties governed by microstructural and compositional heterogeneity \cite{avseth2010rock,nooraiepour2018rock,nooraiepour2017experimental,mou2021coal,nooraiepour2017compaction,nooraiepour2019permeability,bjorlykke2015mudrocks,nooraiepour2022clay}. In UHS contexts, sorption-desorption mechanisms directly influence storage capacity, with materials such as kerogen-rich shales and coals demonstrating significant hydrogen retention potential due to high surface areas, micropore volumes, and organic content \cite{Iglauer2021GL,ARIF202286,WANG2024129919,ALANAZI2023128362,Masoudi2025review}. Notably, hydrogen exhibits lower sorption affinity than CO$_2$ or CH$_4$ in coal matrices, motivating innovative co-injection strategies for combined hydrogen storage and carbon sequestration \cite{ASLANNEZHAD2024125364,RASOOLABID2022121542}. Beyond storage capacity, sorption dynamics govern hydrogen migration through caprocks and sealing formations, directly impacting containment efficiency, leakage risk assessment, and migration pathway prediction in natural hydrogen systems \cite{ZHANG2025138440,Zhang2024H2,WOLFFBOENISCH202313934,TRUCHE2018186}. In radioactive waste repositories, where hydrogen generation from canister corrosion or organic matter degradation poses integrity risks, clay-rich formations such as Callovo-Oxfordian claystone or Opalinus Clay rely on sorption properties to mitigate gas migration and prevent pressure-induced fracturing \cite{XU20083423,BARDELLI2014168,Didier2012,kim2011geological,tsang2005geohydromechanical}.

Predicting hydrogen sorption-desorption behavior in geological media poses substantial challenges. The process is governed by complex, nonlinear interactions among mineralogical composition, pore architecture, surface chemistry, micropore volume, and moisture content \cite{ZIEMIANSKI202228794, MASOUDI2025661, Zhang2024H2,Masoudi2025review}. In fine-grained systems, additional phenomena, including clay swelling, capillary condensation, and pronounced hysteresis—further complicate mechanistic understanding \cite{Masoudi2025review}. Traditional characterization methods, primarily volumetric and gravimetric techniques \cite{BROOM201729320}, provide foundational insights but suffer from inherent limitations: they are time-intensive, costly, technically demanding, and susceptible to significant measurement errors \cite{BROOM201729320}. Moreover, standard protocols typically employ dried, powdered samples that fail to capture the intrinsic heterogeneity, structural anisotropy, and fluid saturation effects characteristic of in situ reservoir conditions \cite{Masoudi2025review,limousin2007sorption,xie2022review}. As demand grows for rapid, accurate assessments of hydrogen storage potential across diverse geological settings, there is an urgent need for computational methodologies that overcome these experimental constraints while maintaining physical rigor.

Machine learning (ML) offers transformative potential for modeling the high-dimensional, nonlinear relationships inherent to geochemical systems \cite{carleo2019machine,malekloo2022machine,nooraiepour2025traditional,zhao2024artificial,zuo2017machine}. Implemented architectures, including ensemble methods and deep neural networks, have demonstrated success in predicting complex subsurface processes through pattern recognition and robust feature extraction \cite{prasianakis2025geochemistry,zuo2017machine,zhang2020using,liu2018towards,abimannan2023ensemble}. However, purely data-driven approaches often prioritize empirical accuracy over physical consistency \cite{mitra2021fitting,malik2020hierarchy,rai2020driven}, resulting in models that may violate fundamental thermodynamic principles, exhibit non-physical extrapolation behavior, or lack interpretability—critical shortcomings for scientific and engineering applications where mechanistic understanding and regulatory confidence are paramount.

To address these challenges, we develop a physics-informed neural network (PINN) framework that integrates classical adsorption theory with deep learning architectures. Our approach embeds established isotherm models directly into the loss function, ensuring predictions satisfy fundamental physical constraints such as saturation limits, monotonic uptake behavior, and thermodynamic consistency. Thermodynamic rigor is further enforced through Van't Hoff analysis \cite{tellinghuisen2006van} and isosteric heat calculations via the Clausius-Clapeyron equation \cite{nuhnen2020practical}. The model leverages multi-category feature engineering to capture thermodynamic state variables, pore structure descriptors, surface chemistry parameters, and kinetic properties across three distinct lithological classes: clays, shales, and coals. We employ deep residual networks (ResNet) augmented with multi-head attention mechanisms to extract hierarchical feature representations and capture complex, non-additive interactions among input variables, thereby enhancing generalization across diverse compositional and environmental conditions.

The framework is trained and validated on a comprehensive laboratory dataset of hydrogen sorption isotherms and adsorption potential in clays, shales, and coals samples spanning pressures up to \SI{200}{\bar}, temperatures from \SI{0}{\celsius} to \SI{90}{\celsius}, and diverse mineralogical and textural properties (e.g., BET surface areas spanning three orders of magnitude, variable micropore volumes, clay mineralogy), as presented in \cite{Masoudi2025review}. This dataset is constructed to be both statistically robust and physically representative of realistic subsurface scenarios, ensuring model predictions remain plausible under operational conditions. Validation employs k-fold cross-validation, supplemented by physics-constrained synthetic benchmarks, to provide performance assessment under both interpolative and controlled extrapolative scenarios.

The contributions of this study are threefold. First, we present a PINN architecture that delivers accurate, interpretable adsorption predictions grounded in both empirical evidence and physical law, yielding not only isotherm predictions but also meaningful sorption parameters (e.g., Langmuir volumes, adsorption affinities) that inform engineering design and geological risk assessment. Second, we demonstrate substantial improvements in scalability and efficiency relative to both traditional experimental campaigns and purely empirical ML models, enabling rapid screening of hydrogen storage potential using readily obtainable rock-matrix descriptors and thermodynamic conditions. Third, the framework is inherently extensible: it can be adapted to multicomponent gas systems (e.g., H$_2$--CH$_4$--CO$_2$ mixtures) for competitive sorption modeling, extended to cyclic sorption-desorption predictions for operational lifetime assessments, and interfaced with pore-scale simulation tools (molecular dynamics, grand canonical Monte Carlo) to enable multiscale understanding of hydrogen behavior in porous media. Ultimately, this work establishes a robust, physics-informed computational framework for evaluating hydrogen sorption in clay- and organic-rich geological formations, bridging theoretical understanding and data-driven prediction to accelerate the development of safe, efficient hydrogen energy infrastructure.

\section{Relevant Theory}
\label{sec:theory}

Accurate prediction of hydrogen sorption in geological formations requires integrating classical adsorption theory with thermodynamic principles to ensure physically consistent results. This section establishes the theoretical foundation for our physics-informed neural network (PINN) framework by presenting: (i) the statistical-mechanical basis of adsorption equilibria, (ii) classical isotherm models and their applicability to different lithologies, and (iii) thermodynamic constraints that enforce physical consistency in data-driven predictions. These theoretical elements are subsequently embedded into the PINN loss function to ensure that model predictions satisfy fundamental physical laws, including saturation limits, monotonic pressure response, and temperature-dependent behavior governed by Van't Hoff and Clausius-Clapeyron relations.

\subsection{Statistical-Mechanical Foundations}
\label{subsec:statmech}

Adsorption equilibria emerge from the minimization of the grand potential at constant temperature, volume, and chemical potential \cite{sorptThermo2004,lee2011thermal}. The grand potential $\Omega$ is defined as:
\begin{equation}
\Omega = A - \mu N
\label{eq:grand_potential}
\end{equation}
where $A$ is the Helmholtz free energy, $\mu$ is the chemical potential of the adsorbate in the bulk gas phase, and $N$ is the number of adsorbed molecules. At equilibrium, $\partial\Omega/\partial N|_{T,V,\mu} = 0$, establishing equality between the chemical potentials of the bulk and adsorbed phases. This condition fundamentally links the bulk gas pressure to the surface excess concentration.

The chemical potential of an ideal gas is expressed as $\mu = \mu^{\circ} + RT\ln(p/p^{\circ})$, where $\mu^{\circ}$ is the standard chemical potential, $p$ is pressure, and $p^{\circ}$ is the reference pressure. For real gases at elevated pressures, the fugacity $f$ replaces pressure, with $\mu = \mu^{\circ} + RT\ln(f/p^{\circ})$. This relationship becomes essential when modeling hydrogen sorption at high pressures, where non-ideal behavior cannot be neglected.

Statistical mechanics provides the bridge between molecular-scale interactions and macroscopic adsorption behavior through the grand canonical partition function $\Xi$ \cite{masel1996principles,Swenson2019,Zangi2024}:
\begin{equation}
\Xi(T,V,\mu) = \sum_{N=0}^{\infty} Q(N,V,T) \, e^{\beta \mu N}
\label{eq:grand_partition}
\end{equation}
where $Q(N,V,T)$ is the canonical partition function for $N$ adsorbed molecules, $\beta = 1/(k_B T)$, and $k_B$ is the Boltzmann constant. The average number of adsorbed molecules is obtained via $\langle N \rangle = k_B T \, (\partial \ln \Xi / \partial \mu)_{T,V}$. For a system of non-interacting adsorption sites with uniform binding energy $\epsilon$, this formalism directly yields the Langmuir isotherm. Extensions to heterogeneous surfaces with a distribution of binding energies $g(\epsilon)$ produce more complex isotherm shapes, including the Freundlich and Temkin forms.

More sophisticated treatments employ density functional theory (DFT) to account for both fluid-solid and fluid-fluid correlations, enabling prediction of density profiles $\rho(\mathbf{r})$ within confined geometries characteristic of nanoporous shales and coals \cite{LANDERS20133}. While molecular-scale simulations (grand canonical Monte Carlo, molecular dynamics) provide detailed insights into hydrogen behavior in specific pore structures, they remain computationally prohibitive for rapid screening across diverse geological scenarios. This motivates our development of a PINN framework that captures the essential physics through classical models while maintaining computational efficiency.

\subsection{Classical Equilibrium Isotherms}
\label{subsec:isotherms}

Classical isotherm models provide functional forms relating gas pressure to adsorbed quantity, each with distinct assumptions about surface properties and adsorption mechanisms. Table~\ref{tab:isotherms} presents an overview of nine fundamental models outlined in this section, highlighting their parameter counts, applicable pressure ranges, key features, and optimal lithological applications. Understanding the physical basis and limitations of each model is essential for selecting appropriate functional forms within the PINN framework and for interpreting the learned parameters in terms of material properties.

\subsubsection{Henry's Law}
\label{subsubsec:henry}

At sufficiently low pressures, where surface coverage is dilute and adsorbate-adsorbate interactions are negligible, adsorption follows a linear relationship \cite{ruthven1984principles}:
\begin{equation}
Q(p) = K_H \, p
\label{eq:henry}
\end{equation}
where $Q$ is the adsorbed amount (mol kg$^{-1}$), $p$ is pressure (Pa), and $K_H$ is Henry's constant (mol kg$^{-1}$ Pa$^{-1}$). This relationship represents the initial slope of all physical adsorption isotherms and is thermodynamically consistent with the limit of infinite dilution. Henry's constant is directly related to the adsorbate-surface interaction energy and provides fundamental insights into hydrogen solubility in formation waters and weak physisorption on mineral surfaces. While applicable only at very low pressures ($<$0.1 MPa), Henry's law serves as an important boundary condition for validating isotherm models in the dilute limit.

\subsubsection{Langmuir Model}
\label{subsubsec:langmuir}

The Langmuir model assumes monolayer adsorption on a homogeneous surface with identical, non-interacting sites \cite{Langmuir1916}:
\begin{equation}
Q(p) = \frac{Q_{\max}\,K_{L}\,p}{1 + K_{L}\,p}
\label{eq:langmuir}
\end{equation}
where $Q_{\max}$ (mol kg$^{-1}$) represents the monolayer saturation capacity and $K_{L}$ (Pa$^{-1}$) is the Langmuir equilibrium constant, related to the adsorption energy. As pressure increases, $Q$ asymptotically approaches $Q_{\max}$, reflecting complete surface coverage. The model provides a clear physical interpretation: $Q_{\max}$ is proportional to the number of available adsorption sites, while $K_{L}$ reflects the affinity between hydrogen and the surface. The Langmuir model performs well for adsorption on homogeneous clay minerals and certain shales at moderate pressures (0.1 to 10 MPa), where monolayer coverage dominates. However, it often fails to capture behavior on heterogeneous surfaces or at very high pressures where multilayer adsorption or pore-filling mechanisms become significant.

\subsubsection{Freundlich Model}
\label{subsubsec:freundlich}

The Freundlich model is an empirical power-law relationship developed for heterogeneous surfaces \cite{freundlich1906over}:
\begin{equation}
Q(p) = K_{F}\,p^{1/n}, \quad n > 1
\label{eq:freundlich}
\end{equation}
where $K_{F}$ (mol kg$^{-1}$ Pa$^{-1/n}$) is the Freundlich capacity factor and $n$ (dimensionless) is the heterogeneity parameter. Values of $n > 1$ indicate favorable adsorption, with larger $n$ reflecting greater surface heterogeneity. Unlike Langmuir, this model does not predict saturation, making it suitable only for intermediate pressure ranges. The power-law form arises naturally from assuming an exponential distribution of adsorption site energies. The Freundlich model is particularly useful for describing geological materials, where surface heterogeneity and pore-size distributions lead to a broad spectrum of binding energies. Its primary limitation is the lack of a thermodynamically consistent saturation limit, which renders it unsuitable for high-pressure extrapolation.

\subsubsection{Brunauer-Emmett-Teller (BET) Model}
\label{subsubsec:bet}

The BET model extends Langmuir theory to multilayer adsorption, accounting for adsorbate-adsorbate interactions beyond the first layer \cite{BET1938}:
\begin{equation}
Q(p) = \frac{Q_{m}\,C\,\frac{p}{p_{0}}}{\left(1 - \frac{p}{p_{0}}\right)\left[1 + (C-1)\,\frac{p}{p_{0}}\right]}
\label{eq:bet}
\end{equation}
where $Q_{m}$ (mol kg$^{-1}$) is the monolayer capacity, $C$ is a constant related to the net heat of adsorption in the first layer ($C = \exp[(E_1 - E_L)/(RT)]$, where $E_1$ is the first-layer adsorption energy and $E_L$ is the heat of liquefaction), and $p_{0}$ is the saturation vapor pressure at temperature $T$. The BET model is foundational in surface area analysis, typically applied to nitrogen adsorption isotherms at 77 K within the relative pressure range $0.05 < p/p_0 < 0.35$. For thermodynamic conditions, where $T > T_c$ (critical temperature 33 K), the concept of saturation pressure becomes problematic, limiting direct application of the BET model. Nevertheless, BET-derived surface areas remain valuable input features for predicting sorption capacity.

\subsubsection{Temkin Model}
\label{subsubsec:temkin}

The Temkin model accounts for indirect adsorbate-adsorbate interactions by assuming the heat of adsorption decreases linearly with surface coverage \cite{tempkin1940kinetics}:
\begin{equation}
Q(p) = \frac{RT}{b_{T}}\,\ln(K_{T}\,p)
\label{eq:temkin}
\end{equation}
where $b_{T}$ (J mol$^{-1}$) is related to the variation in adsorption energy and $K_{T}$ (Pa$^{-1}$) is the Temkin equilibrium constant. This logarithmic form reflects the energetic heterogeneity arising from repulsive interactions between adsorbed molecules or surface potential variations. The Temkin model is particularly relevant for adsorption on charged clay mineral surfaces (montmorillonite, illite) and hydrated interlayer sites, where electrostatic effects modulate binding energies. Its applicability is generally limited to intermediate coverage ranges, as it predicts unphysical behavior at very low and very high pressures.

\subsubsection{Toth Model}
\label{subsubsec:toth}

The Toth model incorporates surface heterogeneity through a modified saturation function \cite{toth1971state,TOTH19951,TOTH1997228}:
\begin{equation}
Q(p) = \frac{Q_{\max}\,p}{\left(b + p^{t}\right)^{1/t}}
\label{eq:toth}
\end{equation}
where $Q_{\max}$ (mol kg$^{-1}$) is the saturation capacity, $b$ (Pa$^{t}$) is a parameter related to adsorption affinity, and $t$ (dimensionless, $0 < t \leq 1$) quantifies surface heterogeneity. When $t = 1$, the Toth model reduces to the Langmuir form, while $t < 1$ represents increasingly heterogeneous surfaces. This model provides excellent fits across an extended pressure range (0.1 to 20 MPa) and is particularly effective for describing heterogeneous matrices with mixed mineralogy and variable pore sizes, which create a distribution of adsorption energies.

\subsubsection{Sips (Langmuir-Freundlich) Model}
\label{subsubsec:sips}

The Sips model combines Langmuir saturation behavior with Freundlich-type heterogeneity \cite{SIPS1948}:
\begin{equation}
Q(p) = \frac{Q_{\max}\,(K_{S}\,p)^{1/n_{s}}}{1 + (K_{S}\,p)^{1/n_{s}}}
\label{eq:sips}
\end{equation}
where $K_{S}$ (Pa$^{-1}$) is the Sips equilibrium constant and $n_s$ (dimensionless, $n_s \geq 1$) is the heterogeneity exponent. This model interpolates between Freundlich behavior at low pressures (where the denominator approaches unity) and Langmuir-like saturation at high pressures. The Sips model is particularly suitable where surface heterogeneity is pronounced, but a well-defined saturation capacity exists. Its three-parameter form provides sufficient flexibility to capture complex isotherm shapes while maintaining thermodynamic consistency through the saturation limit.

\subsubsection{Redlich-Peterson Model}
\label{subsubsec:rp}

The Redlich-Peterson model offers a generalized three-parameter framework that encompasses both Langmuir and Freundlich behavior \cite{Redlich1959}:
\begin{equation}
Q(p) = \frac{K_{\mathrm{RP}} \, p}{1 + A_{\mathrm{RP}} \, p^{\beta}}
\label{eq:rp}
\end{equation}
where $K_{\mathrm{RP}}$ (mol kg$^{-1}$ Pa$^{-1}$), $A_{\mathrm{RP}}$ (Pa$^{-\beta}$), and $\beta$ ($0 < \beta \leq 1$) control isotherm curvature and saturation behavior. When $\beta = 1$, the model reduces to Langmuir with $Q_{\max} = K_{\mathrm{RP}}/A_{\mathrm{RP}}$ and $K_L = A_{\mathrm{RP}}$. When $A_{\mathrm{RP}} \, p^{\beta} \ll 1$, it approaches Freundlich behavior with $K_F = K_{\mathrm{RP}}$ and exponent $\beta$. This versatility enables the Redlich-Peterson model to describe sorption over a very wide pressure range (0.01 to 30 MPa) in heterogeneous geological media. However, its flexibility comes at the cost of parameter identifiability: the three parameters can exhibit strong correlations during optimization, potentially leading to multiple local minima and reduced physical interpretability.

\subsubsection{Dubinin-Radushkevich (D-R) Model}
\label{subsubsec:dr}

The Dubinin-Radushkevich model describes micropore filling through a Gaussian distribution of adsorption potentials \cite{Dubinin1947331}:
\begin{equation}
Q(p) = Q_{s}\,\exp\left(-B\,\epsilon^{2}\right), \quad
\epsilon = RT\,\ln\left(1 + \frac{1}{p}\right)
\label{eq:dr}
\end{equation}
where $Q_{s}$ (mol kg$^{-1}$) is the micropore saturation capacity, $B$ (mol$^2$ J$^{-2}$) is related to the mean adsorption energy, and $\epsilon$ (J mol$^{-1}$) is the adsorption potential. The D-R model is particularly valuable for analyzing pore filling, rather than surface coverage, which governs uptake. The parameter $B$ can be related to the characteristic energy of adsorption, $E = 1/\sqrt{2B}$, providing insights into the strength of adsorbate-adsorbent interactions. The D-R model is most accurate at medium to high pressures (1 to 20 MPa), where micropore saturation becomes the dominant mechanism.

\begin{table}[ht]
\centering
\caption{Comparison of classical adsorption isotherm models for hydrogen sorption prediction. The table summarizes nine fundamental isotherm models, presenting their applicable pressure ranges, distinguishing features, primary advantages, optimal lithological applications, and parameter identifiability characteristics.}
\label{tab:isotherms}

\begin{threeparttable}

\small
\begin{tabular}{p{2.2cm}cp{2cm}p{2.8cm}p{3cm}p{2.3cm}p{1.8cm}}
\hline
\textbf{Model} & \textbf{Param.} & \textbf{Pressure Range} & \textbf{Key Features} & \textbf{Advantages} & \textbf{Best Application} & \textbf{Parameter Identifiability} \\
\hline
Henry's Law & 1 & Very low & Linear, dilute gas & Thermodynamic basis, simple & Initial slope analysis & High \\[0.5ex]
Langmuir & 2 & Low to medium & Monolayer, uniform sites & Physical meaning, well-defined & Shales (monolayer) & High \\[0.5ex]
Freundlich & 2 & Wide range & Empirical power-law & Heterogeneity capture & Coals (microporous) & Medium \\[0.5ex]
BET & 2 & Low relative & Multilayer formation & Surface area analysis & N$_2$ characterization & Medium \\[0.5ex]
Temkin & 2 & Medium & Heat decreases linearly & Interaction energetics & Charged clay minerals & Medium \\[0.5ex]
Toth & 3 & Wide range & Modified saturation & Extended pressure fits & Heterogeneous surfaces & Medium \\[0.5ex]
Sips & 3 & Wide range & Langmuir + heterogeneity & Mixed mineralogies & Clays (heterogeneous) & Medium \\[0.5ex]
Redlich-Peterson & 3 & Very wide range & Langmuir-Freundlich hybrid & Maximum versatility & Universal fitting & Low \\[0.5ex]
Dubinin-Radushkevich & 2 & Medium to high & Micropore filling & Pore volume analysis & Micropore characterization & Medium \\
\hline
\end{tabular}

\begin{tablenotes}\footnotesize
\item[] \textit{Note:} Pressure range descriptors for hydrogen sorption under supercritical conditions: Very low ($<$0.1 MPa), Low (0.1--1 MPa), Medium (1--10 MPa), High (10--20 MPa), Wide (0.1--20 MPa), Very wide (0.01--30 MPa). BET model applies to relative pressure ($p/p_0$) range 0.05--0.35, primarily used for nitrogen adsorption at 77 K. Parameter identifiability indicates the reliability of obtaining unique, well-constrained parameter values during model fitting: High denotes well-separated, independent parameters with unique solutions; Medium indicates moderate parameter correlation with generally stable fits; Low reflects high parameter correlation and potential multiple local optima during optimization.
\end{tablenotes}

\end{threeparttable}
\end{table}

\subsection{Thermodynamic Constraints and Temperature Dependence}
\label{subsec:thermo_constraints}

To ensure physically consistent predictions, the PINN framework must incorporate thermodynamic principles that govern the temperature dependence of adsorption equilibria and establish bounds on model behavior. Three key relationships are enforced: the Van't Hoff equation for temperature-dependent equilibrium constants, the Clausius-Clapeyron equation for isosteric heat of adsorption, and thermodynamic consistency constraints.

\subsubsection{Van't Hoff Relationship}
\label{subsubsec:vanthoff}

The temperature dependence of adsorption equilibrium constants follows the Van't Hoff equation \cite{tellinghuisen2006van}:
\begin{equation}
\frac{d \ln K}{d(1/T)} = -\frac{\Delta H^{\circ}}{R}
\label{eq:vanthoff}
\end{equation}
where $K$ is the equilibrium constant (e.g., $K_L$, $K_S$, $K_T$), $\Delta H^{\circ}$ is the standard enthalpy of adsorption (J mol$^{-1}$), $R$ is the gas constant, and $T$ is absolute temperature (K). For exothermic adsorption ($\Delta H^{\circ} < 0$), $K$ decreases with increasing temperature, reducing sorption capacity at elevated temperatures. Integration of Equation~\ref{eq:vanthoff} yields:
\begin{equation}
K(T) = K_0 \exp\left(\frac{-\Delta H^{\circ}}{RT}\right)
\label{eq:vanthoff_integrated}
\end{equation}
where $K_0$ is a pre-exponential factor. This relationship is embedded into the PINN loss function to ensure that predicted temperature dependencies are thermodynamically consistent. Physically reasonable values of $\Delta H^{\circ}$ for hydrogen physisorption on geological materials are expected to show weak van der Waals interactions.

\subsubsection{Isosteric Heat of Adsorption}
\label{subsubsec:isosteric}

The isosteric heat of adsorption, $q_{\mathrm{st}}$, quantifies the differential enthalpy change upon adsorbing an additional molecule at constant surface coverage. It is obtained from the Clausius-Clapeyron equation applied to adsorption isotherms measured at multiple temperatures \cite{nuhnen2020practical}:
\begin{equation}
q_{\mathrm{st}} = -R \left(\frac{\partial \ln p}{\partial (1/T)}\right)_{Q}
\label{eq:clausius_clapeyron}
\end{equation}
where the derivative is taken at constant adsorbed amount $Q$. In practice, $q_{\mathrm{st}}$ is calculated from experimental isotherms by plotting $\ln p$ versus $1/T$ at fixed $Q$ values; the slope yields $-q_{\mathrm{st}}/R$. For heterogeneous surfaces, $q_{\mathrm{st}}$ typically decreases with increasing coverage as high-energy sites are occupied first. The PINN framework can enforce that predicted $q_{\mathrm{st}}$ values remain within physically plausible bounds and exhibit monotonic or physically justifiable coverage dependence, preventing unphysical extrapolation.

\subsubsection{Thermodynamic Consistency Constraints}
\label{subsubsec:consistency}

Beyond temperature dependence, several thermodynamic principles must be satisfied to ensure the physical validity of predicted isotherms:

\textit{(i) Saturation limits:} For models with explicit saturation capacities (Langmuir, Sips, Toth, D-R), the predicted $Q_{\max}$ or $Q_s$ must be positive and consistent with material properties such as total pore volume or surface area. The PINN loss function penalizes predictions that violate $0 < Q(p) \leq Q_{\max}$.

\textit{(ii) Monotonicity:} Physical adsorption isotherms must be monotonically increasing with pressure at constant temperature: $\partial Q/\partial p|_T \geq 0$. This constraint prevents unphysical oscillations or decreasing behavior in predicted isotherms, particularly during extrapolation beyond the training data range.

\textit{(iii) Convexity and boundary behavior:} At low pressures, isotherms should approach Henry's law behavior ($Q \sim p$), while at high pressures, saturation or sub-linear growth is expected. These asymptotic behaviors can be enforced through penalty terms that evaluate isotherm curvature and limiting slopes.

\textit{(iv) Gibbs adsorption consistency:} The Gibbs adsorption isotherm relates surface excess to pressure and surface tension, providing an independent thermodynamic check. It is not directly enforced in the current framework.

The theoretical framework established in this section forms the foundation for our physics-informed neural network approach. By embedding classical isotherm models (§\ref{subsubsec:langmuir}--§\ref{subsubsec:dr}) into a hybrid loss function that combines data-driven fitting with physics-based penalties, the PINN architecture ensures predictions satisfy fundamental thermodynamic constraints. The Van't Hoff relationship (Equation~\ref{eq:vanthoff}) and Clausius-Clapeyron equation (Equation~\ref{eq:clausius_clapeyron}) enforce temperature-dependent consistency, while monotonicity and saturation bounds prevent unphysical behavior during extrapolation. This integration of domain knowledge with ML enables accurate, interpretable predictions across diverse geological conditions while addressing the key limitations of purely empirical approaches. The following section describes the neural network architecture, feature engineering strategies, and training procedures that operationalize these theoretical principles.

\section{Materials and Methods}
\label{sec:methods}

This section presents the adaptive physics-informed neural network (PINN) framework for predicting hydrogen sorption behavior across fine-grained geological materials. The methodology integrates multi-scale data processing, classical isotherm modeling, physics-informed feature engineering, and deep learning architectures with embedded thermodynamic constraints. Complete methodological details, including algorithmic specifications, statistical validation procedures, and hyperparameter optimization protocols, are provided in the Supplementary Information (Appendix~\ref{sec:supplementary_methods}).

\subsection{Dataset and Integrated Data Processing Pipeline}
\label{subsec:dataset_pipeline}

The dataset comprises 1,987 hydrogen sorption isotherm measurements across three lithological classes: 1,197 measurements on clay mineral samples, 585 on shale formations, and 205 on coal specimens, spanning pressures from 0.1 to 200 bar and temperatures from 273 to 363 K. Additionally, 224 characteristic adsorption capacity measurements provide complementary material property information. Data originate from multiple independent studies employing volumetric and gravimetric techniques \cite{Masoudi2025review}, necessitating careful harmonization before physics-informed modeling.

We developed a comprehensive data integration pipeline implementing: (i) intelligent column mapping across heterogeneous data sources using pattern matching and domain-specific heuristics, (ii) lithology-aware processing preserving material-specific properties while enabling cross-lithology learning, and (iii) multi-level quality assessment including completeness scoring, interquartile range outlier detection, and thermodynamic consistency validation. Critical preprocessing steps enforce fundamental physical constraints: monotonicity checks confirm non-decreasing sorption with increasing pressure at constant temperature, saturation limits prevent unphysical uptake predictions, and temperature dependencies are validated against Van't Hoff behavior (Section~\ref{subsec:thermo_constraints}). Features were standardized to zero mean and unit variance within each lithological class to preserve material-specific scaling while enabling effective gradient-based optimization. Detailed data integration protocols, quality metrics, and validation procedures are provided in Appendix~\ref{subsec:supp_data_integration}.

\subsection{Classical Isotherm Analysis and Thermodynamic Parameter Extraction}
\label{subsec:classical_analysis_main}

Classical isotherm models provide physically interpretable baseline performance and extract thermodynamic parameters that constrain subsequent PINN training. We implemented a comprehensive two-stage analysis: (i) individual sample fitting, extracting sample-specific parameters and thermodynamic properties, and (ii) aggregated data analysis, evaluating whether classical models generalize across heterogeneous samples without ML augmentation.

Individual sample characterization fitted nine classical models (Henry, Langmuir, Freundlich, BET, Temkin, Toth, Sips, Redlich-Peterson, Dubinin-Radushkevich) to each sample's isotherm data using differential evolution optimization within physically meaningful parameter bounds. Physics-based validators enforced fundamental constraints, including positive sorption capacities, thermodynamically consistent equilibrium constants, and monotonic pressure response. Parameter uncertainty was quantified via bootstrap resampling (500 iterations), and cross-validation (five-fold splitting) was used to assess generalization performance. For multi-temperature measurements, Van't Hoff analysis extracted standard enthalpy ($\Delta H_{\text{ads}}$), entropy ($\Delta S_{\text{ads}}$), and Gibbs free energy ($\Delta G_{\text{ads}}$) according to Equation~\ref{eq:vanthoff}. Isosteric heats of adsorption were calculated via the Clausius-Clapeyron equation (Equation~\ref{eq:clausius_clapeyron}), providing coverage-dependent energetics that reveal surface heterogeneity.

Aggregated analysis evaluated 22 functional forms (10 physics-based classical isotherms plus 12 mathematical equations including polynomials, exponentials, power laws, and growth curves) fitted to lithology-specific and combined datasets. Statistical validation employed bootstrap resampling (1,000 iterations), five-fold cross-validation, and residual diagnostics (Shapiro-Wilk and Durbin-Watson tests) to assess model adequacy and generalization. This dual-stage approach establishes both thermodynamic foundations for PINN constraints and empirical justification for PINN necessity. Complete fitting procedures, parameter bounds, validation metrics, and comparative results appear in Appendix~\ref{subsec:supp_classical_modeling}.

\subsection{Property-Uptake Correlation and Feature Selection}
\label{subsec:property_analysis_main}

To establish quantitative relationships between material properties and hydrogen sorption capacity, we implemented a systematic correlation analysis with bootstrap uncertainty quantification (1,000 iterations) to obtain robust confidence intervals. Pearson correlation coefficients identified linear relationships, while supervised learning models (linear regression, ridge regression with L2 regularization $\alpha = 1.0$, random forest with 50 estimators and maximum depth 5) captured non-linear dependencies. Generalization performance was evaluated via five-fold cross-validation, and feature importance rankings from random forests (100 estimators) identified properties most strongly predicting uptake across lithologies.

Lithology-specific hydrogen uptake distributions were characterized through comprehensive descriptive statistics (central tendency, dispersion, range boundaries) that inform PINN output constraints: maximum observed uptake values establish physical upper bounds, while mean and median values guide output layer initialization. This analysis validates the physical mechanisms governing sorption and provides guidance on feature selection for neural network architecture design. Detailed correlation matrices, supervised learning results, feature importance rankings, and lithology-specific statistics are provided in Appendix~\ref{subsec:supp_property_analysis}.

\subsection{Physics-Informed Feature Engineering}
\label{subsec:feature_engineering_main}

Beyond raw measurements, effective neural network training requires engineered features that encode the physical relationships governing hydrogen sorption. We developed a systematic seven-category framework generating 105 to 120 physics-informed features (depending on lithology):

\textbf{(1) Thermodynamic descriptors:} Temperature conversions, logarithmic pressure, reduced temperature and pressure ($T_r = T/T_c$, $p_r = p/p_c$ where $T_c = 33.19$ K, $p_c = 13.13$ bar), inverse temperature ($1/T$), and approximate Gibbs free energy.

\textbf{(2) Pore structure descriptors:} Micropore fraction ($V_{\text{micro}}/V_{\text{total}}$), surface area density ($S_{\text{BET}}/V_{\text{total}}$), confinement parameter ($d_{\text{pore}}/d_{\text{H}_2}$ where $d_{\text{H}_2} = 0.289$ nm), and logarithmic transformations compressing multi-order-of-magnitude property ranges.

\textbf{(3) Surface chemistry and composition:} Lithology-specific features including TOC for shales ($r = 0.755$ correlation with uptake, Appendix~\ref{subsec:supp_property_analysis}), carbon maturity indices and fuel ratio for coals, and clay mineralogy parameters.

\textbf{(4) Interaction features:} Temperature-surface area products ($S_{\text{BET}} \times T$), pressure-pore volume products ($p \times V_{\text{pore}}$), and adsorption driving force ($\Phi_{\text{ads}} = p \cdot S_{\text{BET}}/T$).

\textbf{(5) Kinetic descriptors:} Knudsen diffusion coefficients, molecular mean free path, and diffusion timescales.

\textbf{(6) Molecular sieving parameters:} Sieving factors, pore accessibility indices, and ultramicropore/supermicropore indicators.

\textbf{(7) Classical model-inspired features:} Langmuir-inspired saturation terms, Freundlich power laws, and Temkin logarithmic forms.

Ensemble feature selection combined Pearson correlation, mutual information, random forest importance, and F-statistics to identify the 50 most informative features. RobustScaler transformation (median centering, IQR scaling) provided outlier resistance while preserving rank-order relationships. Complete feature engineering formulations, selection validation, and scaling procedures appear in Appendix~\ref{subsec:supp_feature_engineering}.

\subsection{Multi-Scale Physics-Informed Neural Network Architecture}
\label{subsec:pinn_architecture_main}

The PINN architecture integrates three design innovations addressing data-constrained geological modeling: (i) hierarchical multi-scale feature extraction through parallel pathways (64, 128, 256 neurons) capturing patterns at multiple abstraction levels, (ii) physics-informed gating mechanisms ($\mathbf{g} = \sigma(W_g [p; T] + b_g)$) that dynamically emphasize thermodynamically relevant features, and (iii) progressive dimensional modulation via encoder-decoder topology (256, 512, 256, 128 neurons) with residual connections mitigating vanishing gradients. The architecture totals 887,447 trainable parameters, balancing sufficient capacity for heterogeneous sorption physics across three lithologies against tractability for moderate-sized datasets. Hidden layers employ Swish activation ($f(x) = x \cdot \sigma(x)$), enabling complex non-linearities, while output layers use Softplus activation ($f(x) = \ln(1 + e^x)$), enforcing non-negativity constraints without discontinuous gradients.

The multi-term physics-informed loss function combines data fidelity, physical consistency, hard constraints, and monotonicity:
\begin{equation}
\mathcal{L}_{\text{total}} = \lambda_1 \mathcal{L}_{\text{data}} + \lambda_2 \mathcal{L}_{\text{physics}} + \lambda_3 \mathcal{L}_{\text{bounds}} + \lambda_4 \mathcal{L}_{\text{monotonicity}}
\label{eq:pinn_loss_main}
\end{equation}
where data loss employs sigmoid temperature-scaled weighting addressing target distribution skewness, physics loss penalizes violations of lithology-specific Langmuir saturation limits extracted from classical isotherm analysis (clays: 1.2 mmol/g, shales: 1.0 mmol/g, coals: 0.88 mmol/g), bounds constraints enforce $0 \leq Q \leq q_{\max}$, and monotonicity terms ensure $\partial Q/\partial p|_T \geq 0$ via automatic differentiation. Loss weights adapt dynamically through gradient-magnitude-based normalization, ensuring balanced contributions across competing objectives. Detailed architecture specifications, loss function derivations, and adaptive weighting algorithms appear in Appendix~\ref{subsec:supp_pinn_architecture}.

\subsection{Progressive Training Strategy and Ensemble Methods}
\label{subsec:training_main}

Training proceeds through three sequential phases implementing curriculum learning: (i) data-driven warmup (50 epochs, data loss only, learning rate $\eta = 1.2 \times 10^{-3}$), (ii) physics integration (250 epochs, linearly increasing physics weight from 0 to 1, cosine annealing $\eta = 5 \times 10^{-4}$ to $\eta_{\min} = 10^{-6}$), and (iii) full constraints (100 epochs, all loss terms active, $\eta = 10^{-4}$ to $\eta_{\min} = 10^{-7}$). AdamW optimizer with weight decay $\lambda_{\text{wd}} = 10^{-5}$, batch size 64, and automatic mixed precision accelerate convergence. Gradient clipping (maximum norm 1.0) prevents exploding gradients during physics loss backpropagation.

Systematic hyperparameter optimization spanning 30+ experiments identified optimal configurations: dropout rate 0.10 (balancing regularization against capacity preservation), 400 total training epochs (convergence completeness versus overfitting risk), initial learning rate $\eta = 1.2 \times 10^{-3}$, and batch size 64 (gradient stability versus stochastic regularization). Comprehensive regularization integrates L2 weight decay, dropout, batch normalization, data augmentation via mini-batch sampling, early stopping, and stratified sample-level train-validation-test splitting that prevents information leakage.

Predictive uncertainty quantification employs architecture-diverse ensemble learning with 10 members exhibiting genuine topological variation (width variations: 0.75$\times$, 1.0$\times$, 1.25$\times$ base neurons; depth variations: 3 to 5 layers; hybrid configurations; regularization diversity) rather than dropout-only variation. Each member trains with distinct random initializations across 10 different seeds (from 42 to 9999), exploring independent optimization trajectories. Ensemble predictions aggregated via mean and standard deviation, with post-hoc temperature-scaling calibration optimizing coverage rates on held-out validation data. Complete training protocols, hyperparameter search results, ensemble diversity analysis, and calibration procedures appear in Appendix~\ref{subsec:supp_training}.

\subsection{Evaluation Framework and Interpretability Analysis}
\label{subsec:evaluation_main}

Model evaluation employs 35 metrics spanning five domains: (i) the regression performance of the models was evaluated using the following standard metrics: the coefficient of determination~($R^2$), which quantifies the proportion of variance in the observed hydrogen adsorption values explained by the model; the root mean squared error~(RMSE), which measures the square root of the average squared differences between predicted and actual values (in mmol/g); the mean absolute error~(MAE), representing the average absolute deviation between predictions and observations (in mmol/g); the mean absolute percentage error~(MAPE), expressing the average absolute error as a percentage of the true values; and the mean bias error~(MBE), which indicates the average signed deviation (positive values denote systematic overprediction, negative values underprediction) between predicted and observed adsorption capacities (in mmol/g).

(ii) correlation metrics (Pearson, Spearman, Kendall tau), (iii) physics consistency metrics (constraint violation rate, saturation consistency, monotonicity score), (iv) uncertainty quantification metrics (prediction interval coverage at 68\%, 95\%, 99\% confidence levels, calibration error, sharpness, uncertainty-error correlation), and (v) statistical tests (Shapiro-Wilk, Durbin-Watson, heteroscedasticity tests). Evaluation is performed on a strictly held-out test set (15\% of the total data), completely isolated from training, validation, and hyperparameter selection.

Interpretability analysis employs three complementary methodologies: SHAP (Shapley Additive Explanations), providing game-theoretic feature attribution capturing feature interactions \cite{ponce2024practical,wang2024feature}, Accumulated Local Effects (ALE) plots visualizing global marginal effects while accounting for feature correlations \cite{danesh2022interpretability,okoli2023statistical}, and Friedman's H-statistic quantifying pairwise interaction strength \cite{friedman2008predictive,inglis2022visualizing}. These methods validate that the PINN captures physically meaningful relationships between lithological properties and thermodynamic conditions. Comprehensive evaluation results, interpretability visualizations, and comparative benchmarking against classical models appear in Section~\ref{sec:RD}. Detailed evaluation protocols and interpretability methodologies are provided in Appendix~\ref{subsec:supp_evaluation}.

\section{Results}
\label{sec:RD}

\subsection{Dataset Characterization and Experimental Coverage}
\label{subsec:dataset_characterization}

The hydrogen sorption dataset comprises 1,987 isotherm measurements across three distinct geological lithologies: clay minerals (1,197), shale formations (585), and coal samples (205). These data are complemented by 224 characteristic uptake measurements (123 clays, 39 shales, 62 coals) representing maximum or equilibrium hydrogen sorption capacities under specific experimental conditions. The dataset encompasses pressures from near-ambient conditions to 200 bar and temperatures ranging from cryogenic conditions ($-$253°C) to elevated subsurface temperatures (120°C), providing coverage across both liquefied hydrogen storage scenarios and deep geological formation conditions.

Clay minerals demonstrate the broadest experimental parameter space, with specific surface areas spanning 2.96 to 273.1 m$^2$/g (mean 81.8 $\pm$ 73.6 m$^2$/g) and temperature coverage from $-$253°C to 120°C. Isotherm measurements exhibit mean temperature and pressure conditions of 43.3 $\pm$ 30.4°C and 50.8 $\pm$ 47.0 bar, respectively, with hydrogen uptake averaging 0.204 $\pm$ 0.290 mmol/g (median 0.063 mmol/g). The large standard deviations and skewed distributions reflect the structural diversity of clay mineralogies, from low-surface-area non-swelling clays (e.g., kaolinite) to high-surface-area swelling clays (e.g., smectites, montmorillonite, and vermiculite). Pore volume (0.128 $\pm$ 0.112 cm$^3$/g) and micropore volume (0.021 $\pm$ 0.025 cm$^3$/g) exhibit right-skewed distributions characteristic of hierarchical pore structures, with strong positive correlations to uptake capacity (Pearson $r = 0.52$ and $r = 0.29$, respectively).

Shale formations exhibit substantially lower specific surface areas (0.030 $\pm$ 0.015 m$^2$/g, range 0.01--0.05 m$^2$/g). Compositional analysis reveals a mean total organic carbon (TOC) content of $10.5 \pm 6.1$ wt\% (median 9.7 wt\%, range 0.7--19.4 wt\%), calcite content of $26.0 \pm 33.4$ wt\%, quartz content of $20.3 \pm 15.9$ wt\%, pyrite content of $2.2 \pm 2.7$ wt\%, and clay mineral content of $25.3 \pm 24.7$ wt\%. These statistical distributions reflect the strong mineralogical heterogeneity that is characteristic of fine-grained sedimentary deposits. Calcite and clay minerals show the widest variability (standard deviations exceeding their respective means), indicating the presence of both carbonate-dominated and siliciclastic/clay-dominated end-members within the sample set. TOC values exhibit moderate dispersion and remain almost entirely below 20 wt\%.

Coal samples display intermediate surface area characteristics (2.81 $\pm$ 4.59 m$^2$/g, range 0.05--30.51 m$^2$/g), with measurements extending to 120 bar and 105°C (mean conditions: 39.6 $\pm$ 15.9°C, 55.5 $\pm$ 39.6 bar). Coal properties reveal mean fixed carbon content of 64.1 $\pm$ 17.7 wt\%, volatile matter of 20.0 $\pm$ 14.3 wt\%, and vitrinite reflectance (\%R$_o$) of 2.12 $\pm$ 1.59\%, spanning low-volatile bituminous to anthracite ranks. Hydrogen uptake exhibits bimodal behavior with mean values of 0.179 $\pm$ 0.167 mmol/g for isotherm measurements (median 0.143 mmol/g) and 0.378 $\pm$ 0.206 mmol/g for characteristic capacity measurements (median 0.320 mmol/g). Strong correlations exist between fixed carbon content and uptake ($r = 0.71$), micropore volume and uptake ($r = 0.65$), and a pronounced negative correlation between volatile matter and uptake ($r = -0.71$), indicating that coal rank and micropore development are primary controls on hydrogen storage capacity.

Strong correlations between material properties and uptake capacity provide clear physical constraints for model development. Across all lithologies, specific surface area demonstrates positive correlations with hydrogen uptake (clays: $r = 0.60$; shales: limited data; coals: $r = 0.13$), though the relationship weakens in coals where micropore architecture supersedes total surface area as the dominant control. Pore volume metrics exhibit lithology-dependent behavior: in clays, total pore volume correlates moderately with uptake ($r = 0.52$), while in coals, micropore volume shows a stronger association ($r = 0.65$) due to size-selective molecular sieving effects that preferentially accumulate hydrogen in sub-nanometer pores. Coal-specific parameters reveal that fixed carbon content ($r = 0.71$) and volatile matter content ($r = -0.71$) serve as robust proxies for sorption capacity, reflecting the fundamental relationship between coal rank, aromaticity, and microporosity development during coalification. These strong correlations ($|r| > 0.5$) establish physically grounded feature relationships that can be leveraged for PINN training while maintaining interpretability.

The dataset's multi-scale nature, spanning molecular-level physisorption to bulk-scale storage phenomena, presents both opportunities and challenges for physics-informed modeling. The pressure range (0--200 bar) encompasses distinct adsorption regimes governed by different theoretical frameworks: dilute Henry's law linearity at low pressures ($<$1 bar), Langmuir-type monolayer saturation at intermediate pressures (1--50 bar), and potential multilayer formation or micropore condensation mechanisms at elevated pressures ($>$50 bar). Temperature variations spanning supercritical to subcritical hydrogen conditions may enable thermodynamic validation through Van't Hoff analysis of temperature-dependent equilibrium constants and isosteric heat calculations via the Clausius-Clapeyron equation (Equation~\ref{eq:clausius_clapeyron}). The lithology-specific variations in sorption behavior necessitate adaptive network architectures capable of capturing material-dependent phenomena while remaining consistent with the universal thermodynamic principles established in Section~\ref{subsec:thermo_constraints}. Statistical distributions reveal pronounced non-normality across all measured parameters (Shapiro-Wilk $p < 0.05$ for all variables), with skewness values ranging from $-3.2$ to $+4.8$, necessitating log-transformations for pressure, uptake, and surface area features to stabilize variance and improve model convergence during gradient-based optimization.

\subsection{Classical Isotherm Model Performance and Generalization Assessment}
\label{subsec:classical_results}

\subsubsection{Individual Sample Fitting: Model Performance and Lithology-Specific Preferences}
\label{subsubsec:individual_performance}

Classical isotherm fitting yielded 621 model evaluations, of which 617 converged successfully (99.4\% success rate). Nine theoretical frameworks as presented in Section \ref{subsec:isotherms} were systematically evaluated: Henry, Langmuir, Freundlich, BET, Temkin, Toth, Sips, Redlich-Peterson, and Dubinin-Radushkevich models. Model selection was guided by Akaike information criterion (AIC), Bayesian information criterion (BIC), and physics compliance scores, with the best-performing model selected for each sample based on a composite ranking that balanced statistical fit quality, thermodynamic consistency, and parameter identifiability.

\begin{figure}[h!]
\centering
\includegraphics[width=0.95\textwidth]{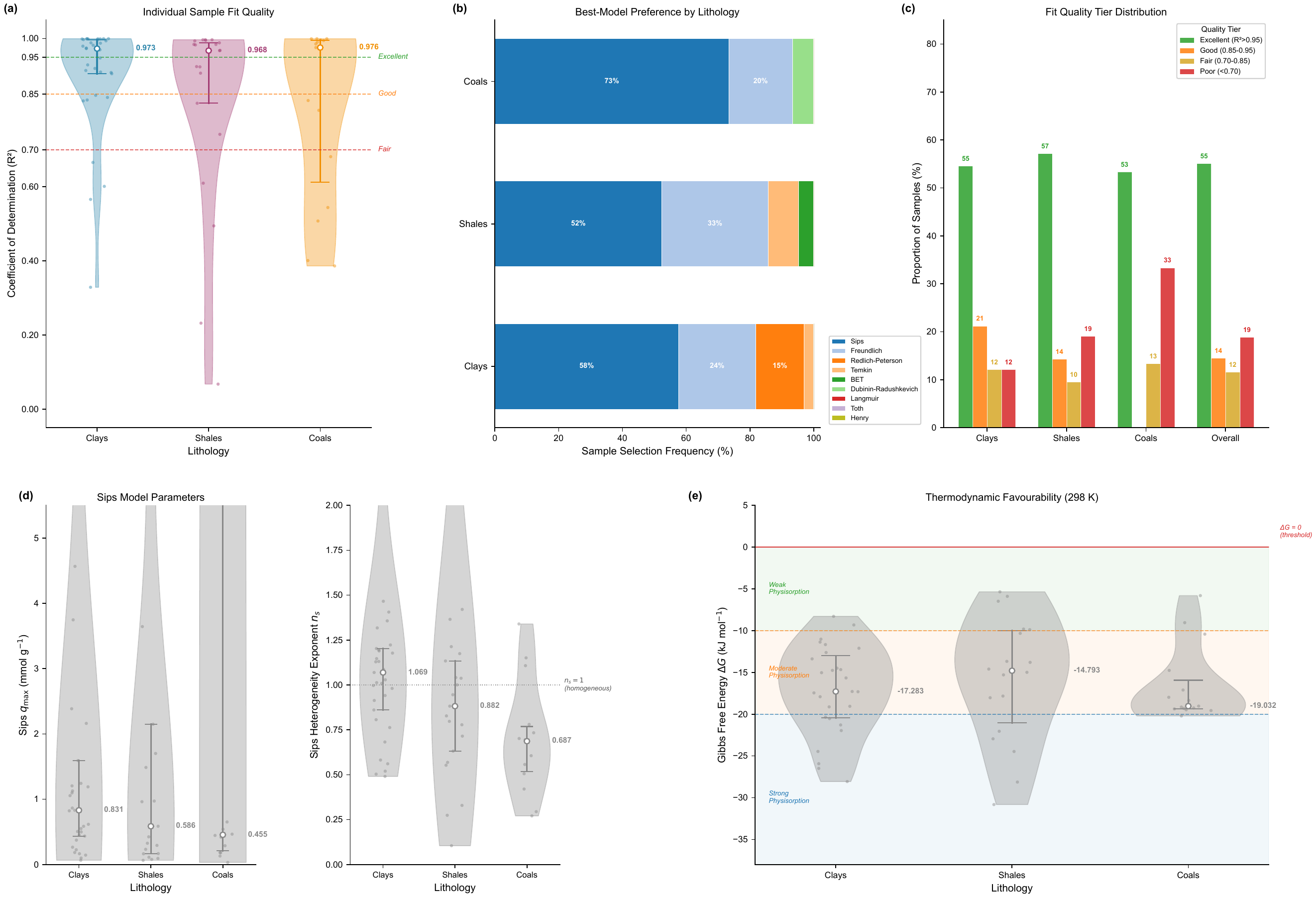}
\caption{%
Classical isotherm model-fitting performance across individual geological samples of clays, shales, and coals. Nine classical models were fitted per isotherm. The best model was selected via the lowest AIC. High per-sample accuracy shows effective capture of local adsorption, yet fails to generalise across samples or aggregates (see Fig.~\ref{fig:aggregated_failure}).
\textbf{(a)} Best-fit model $R^2$ distributions by lithology (violin plots with interquartile boxes and data points). Dashed lines mark thresholds: $R^2 = 0.95$ (excellent), $0.85$ (good), $0.70$ (acceptable).
\textbf{(b)} Model preference by lithology (\% of samples with lowest AIC). Sips model dominates overall (59.4\% of best fits).
\textbf{(c)} Sample quality tiers by lithology: Excellent ($R^2 > 0.95$), Good ($0.85$--$0.95$), Fair ($0.70$--$0.85$), Poor ($R^2 < 0.70$). Percentages show tier shares within each lithology.
\textbf{(d)} Sips model parameter distributions: maximum capacity $q_{\max}$ (mmol~g$^{-1}$) and heterogeneity exponent $n_s$ (reference line at $n_s = 1$, Langmuir limit).
\textbf{(e)} Gibbs free energy of adsorption ($\Delta G$, kJ~mol$^{-1}$) by lithology, derived from isosteric estimates. Shaded bands indicate physisorption strength: weak (0 to $-10$), moderate ($-10$ to $-20$), strong ($< -20$).}
\label{fig:individual_fitting}
\end{figure}

Overall fit quality for individual samples demonstrates strong sample-specific predictive capability, with median R$^2$ = 0.970 and mean R$^2$ = 0.861 $\pm$ 0.213 across all selected models (Table~\ref{tab:classical_summary}; Fig.~\ref{fig:individual_fitting}). Quality assessment categorizes 55.1\% of fits as excellent (R$^2 > 0.95$), 14.5\% as good (0.85--0.95), 11.6\% as fair (0.70--0.85), and 18.8\% as requiring further evaluation (R$^2 < 0.70$) (Fig.~\ref{fig:individual_fitting}c). Clay minerals exhibit the highest average fit quality (mean R$^2$ = 0.903 $\pm$ 0.153, median = 0.973), followed by shales (mean R$^2$ = 0.836 $\pm$ 0.267, median = 0.968) and coals (mean R$^2$ = 0.807 $\pm$ 0.238, median = 0.976) (Fig.~\ref{fig:individual_fitting}a). The higher median than mean values across all lithologies indicate that most samples achieve excellent individual fits, with lower means reflecting a small subset of poorly fitting samples warranting exclusion from training datasets.

\begin{table}[h!]
\centering
\caption{Summary of classical isotherm model performance for individually fitted samples across geological lithologies. Statistics presented as mean $\pm$ standard deviation. Model preference indicates the percentage of samples best represented by each theoretical framework after multi-criteria selection (AIC, BIC, physics score).}
\label{tab:classical_summary}
\small
\begin{tabular}{lccccc}
\hline
\textbf{Lithology} & \textbf{Mean R$^2$} & \textbf{Median R$^2$} & \textbf{RMSE} & \textbf{Top Model} & \textbf{Excellent Fits} \\
 &  &  & (mmol/g) & (Preference) & (\%) \\
\hline
Clays & 0.903 $\pm$ 0.153 & 0.973 & 0.029 $\pm$ 0.044 & Sips (57.6\%) & 63.6 \\
Shales & 0.836 $\pm$ 0.267 & 0.968 & 0.030 $\pm$ 0.048 & Sips (52.4\%) & 52.4 \\
Coals & 0.807 $\pm$ 0.238 & 0.976 & 0.026 $\pm$ 0.035 & Sips (73.3\%) & 40.0 \\
\hline
\textbf{Overall} & \textbf{0.861 $\pm$ 0.213} & \textbf{0.970} & \textbf{0.029 $\pm$ 0.044} & \textbf{Sips (59.4\%)} & \textbf{55.1} \\
\hline
\multicolumn{6}{l}{\footnotesize Secondary model preferences: Freundlich (26.1\%), Redlich-Peterson (7.2\%), Temkin (4.3\%), BET (1.4\%), D-R (1.4\%)} \\
\multicolumn{6}{l}{\footnotesize Quality categories based on the coefficient of determination thresholds established for training dataset curation:} \\
\multicolumn{6}{l}{\footnotesize Excellent (R$^2 > 0.95$), Good (0.85--0.95), Fair (0.70--0.85), Poor ($<$0.70$)$.} \\
\end{tabular}
\end{table}

Model preferences show lithology-dependent patterns reflecting underlying sorption mechanisms (Fig.~\ref{fig:individual_fitting}b). The Sips (Langmuir-Freundlich) model emerges as the dominant framework across all lithologies, selected for 59.4\% of samples overall (57.6\% clays, 52.4\% shales, 73.3\% coals). This prevalence indicates that hydrogen sorption in these geological materials exhibits combined Langmuir-type saturation with Freundlich-type surface heterogeneity, consistent with the compiled dataset of \cite{Masoudi2025review}. The Freundlich model ranks second in preference (26.1\% overall), particularly for samples exhibiting pronounced heterogeneity without clear saturation behavior within the measured pressure range. Clay-specific features include preference for Redlich-Peterson (15.2\%) and Temkin (3.0\%) models in select high-surface-area samples, reflecting complex multi-site adsorption energetics. Shale formations show occasional BET model selection (4.8\%), suggesting multilayer formation on specific mineral surfaces, while coal samples uniquely exhibit Dubinin-Radushkevich model applicability (6.7\%), consistent with micropore volume-filling mechanisms characteristic of high-rank coals.

The model performance assessment established quality tiers for subsequent analyses. Quality scoring combined statistical metrics (R$^2$, residual patterns), physics compliance, data sufficiency (minimum 6 pressure points), and parameter identifiability.

\subsubsection{Individual Sample Fitting: Parameter Uncertainty and Cross-Validation}
\label{subsubsec:individual_uncertainty}

Bootstrap resampling with 500 iterations provided 95\% confidence intervals for all fitted parameters, enabling assessment of parameter identifiability and stability. For the dominant Sips model, fitted parameters span physically meaningful ranges (Fig.~\ref{fig:individual_fitting}d): maximum sorption capacity $q_{\text{max}} = 0.774 \pm 1.324$ mmol/g (range 0.032--8.295 mmol/g, median 0.445 mmol/g), affinity constant $K_S = 0.045 \pm 0.112$ bar$^{-1}$ (range 0.00036--0.680 bar$^{-1}$, median 0.012 bar$^{-1}$), and heterogeneity exponent $n_s = 0.763 \pm 0.323$ (range 0.105--1.466, median 0.702). The heterogeneity parameter distribution ($n_s < 1$ for most samples) indicates that geological materials exhibit surface energy distributions rather than uniform adsorption sites. Freundlich model parameters show capacity factors $K_F = 0.015 \pm 0.032$ and heterogeneity indices $n = 1.88 \pm 1.88$ (median 1.20), with $n > 1$ validating favorable adsorption conditions.

Five-fold cross-validation assessed model generalization capability within individual samples, yielding mean cross-validation R$^2$ values comparable to training fits (difference typically $<$0.02), indicating minimal overfitting for sample-specific predictions. Bootstrap-derived confidence intervals revealed that saturation capacity parameters ($q_{\text{max}}$, $Q_s$) exhibit relatively tight bounds (typical 95\% CI spanning 10--30\% of mean value), while affinity constants show broader uncertainty ranges (50--100\% of mean) due to correlation with heterogeneity parameters. This parameter correlation structure informed subsequent PINN initialization strategies, where saturation capacities were constrained more tightly than affinity parameters.

\subsubsection{Individual Sample Fitting: Physics Validation and Thermodynamic Consistency}
\label{subsubsec:individual_physics}

Physics-based validators enforced fundamental constraints during optimization, computing compliance scores for each fitted model. Validation criteria included: (i) positive parameter constraints ($q_{\text{max}}, K, n > 0$), (ii) saturation limit adherence ($q_{\text{max}} \geq Q_{\text{obs,max}}$), (iii) monotonicity requirements ($\partial Q/\partial p \geq 0$), (iv) Freundlich favorability ($n > 1$ for physical adsorption), and (v) BET relative pressure bounds ($0.05 < p/p_0 < 0.35$). Average physics compliance scores exceeded 0.90 for all individual samples across lithologies, with 89.9\% of selected models achieving perfect or near-perfect scores ($\geq 0.95$). Seven samples yielded physics scores below 0.70, were flagged for manual review, and were checked and re-imported to prevent potential experimental artifacts or non-equilibrium measurements from corrupting model initialization.

Thermodynamic analysis of temperature-dependent datasets for samples with multi-temperature measurements extracted Gibbs free energy values via fitted equilibrium constants. Results uniformly indicate favorable adsorption (all $\Delta G < 0$), with mean $\Delta G = -16.6 \pm 5.8$ kJ/mol (range $-30.8$ to $-5.4$ kJ/mol) (Fig.~\ref{fig:individual_fitting}e). Energy distribution classifies 61.7\% as moderate physisorption ($-20 < \Delta G < -10$ kJ/mol), 23.3\% as strong physisorption ($\Delta G < -20$ kJ/mol), and 15.0\% as weak physisorption ($\Delta G > -10$ kJ/mol). These values align with expected hydrogen physisorption energetics on mineral surfaces, validating that fitted parameters reflect genuine physical phenomena rather than mathematical artifacts.

\begin{figure}[h!]
\centering
\includegraphics[width=0.95\textwidth]{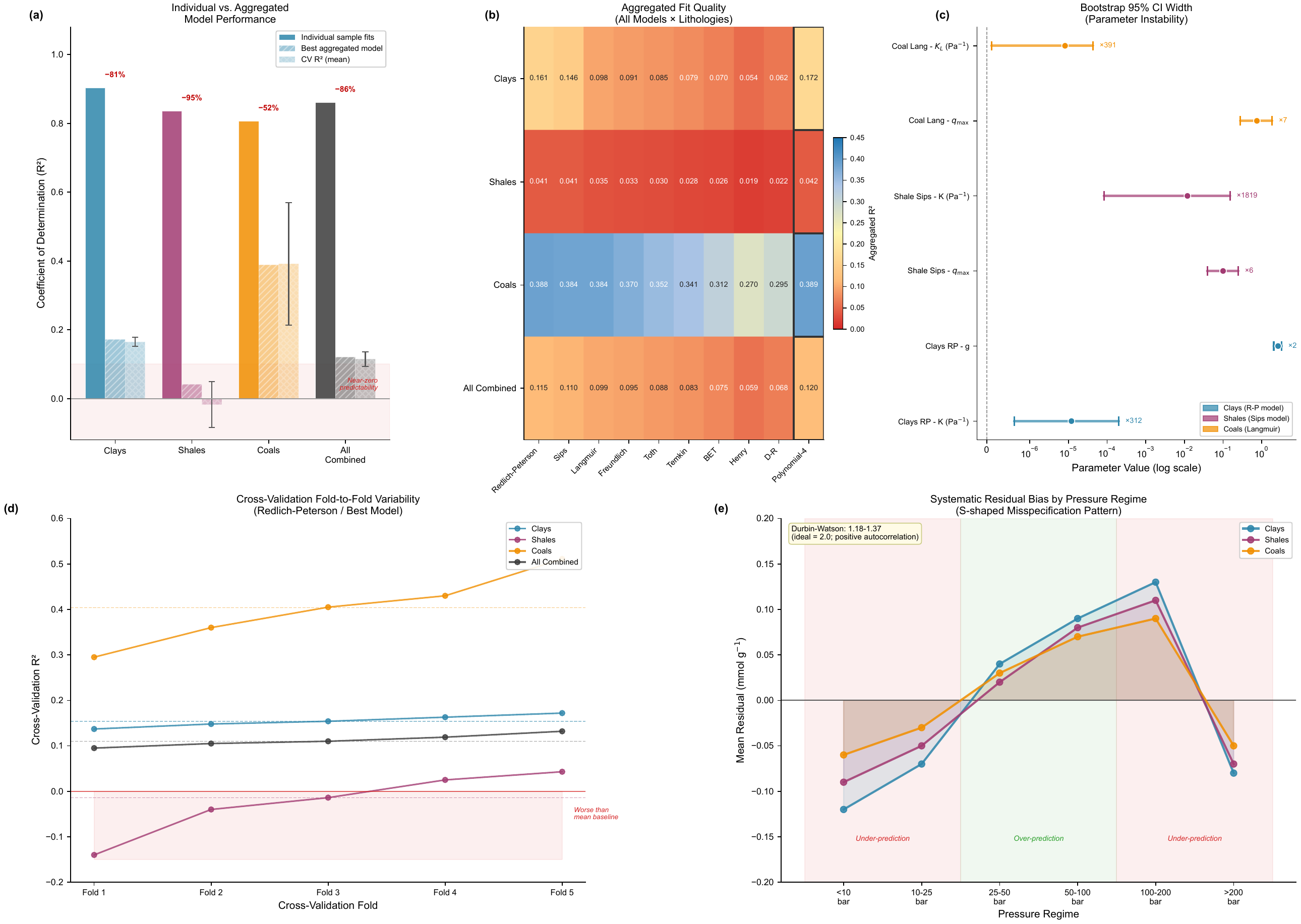}
\caption{%
Failure of classical isotherm models in aggregated, cross-sample generalisation. While individual-sample fits achieve median $R^2 = 0.97$ (Fig.~\ref{fig:individual_fitting}), aggregating data within each lithology and re-fitting yields 80--90\% mean reduction in $R^2$, motivating the physics-informed neural network approach.
\textbf{(a)} Comparison of individual-sample $R^2$, aggregated $R^2$, and cross-validation $R^2$ by lithology and overall. Percentage reductions annotate performance collapse.
\textbf{(b)} Aggregated $R^2$ heatmap for nine classical models plus fourth-order polynomial across four groupings (clays, shales, coals, all combined). Colour scale 0--0.45; best model per group highlighted (bold border).
\textbf{(c)} Bootstrap parameter uncertainty (95\% CI error bars) for dominant model per lithology: Redlich--Peterson (clays), Sips (shales), Langmuir (coals). Fold-width annotations (e.g.\ $\times 10^{2}$) indicate parameter instability.
\textbf{(d)} Five-fold cross-validation $R^2$ by lithology. Shale folds reach negative values, showing worse-than-mean performance.
\textbf{(e)} Residual bias across six pressure bins ($<$10 to $>$200~bar). S-shaped pattern (under-prediction at low/high pressures, over-prediction at intermediate) reveals structural misspecification. Durbin--Watson statistics (1.18--1.37) confirm positive residual autocorrelation.}
\label{fig:aggregated_failure}
\end{figure}

\subsubsection{Aggregated Data Analysis: Generalization Failure and Model Limitations}
\label{subsubsec:aggregated_generalization}

While individual sample fitting demonstrates that classical isotherms accurately describe single-sample behavior, a critical question remains: can these models generalize across heterogeneous sample populations without adaptive ML architectures? To address this question, we conducted a rigorous generalization assessment by fitting 22 distinct functional forms (10 classical isotherm models and 12 purely mathematical equations) to aggregated datasets at three hierarchical levels: per-lithology aggregation and complete aggregation across all lithologies. Each fit employed bootstrap resampling with 1,000 iterations and five-fold cross-validation to quantify parameter uncertainty and generalization performance.

Aggregated fitting results reveal systematic and severe generalization failures across all lithologies and model types (Table~\ref{tab:generalization_failure}; Fig.~\ref{fig:aggregated_failure}a). For clay minerals, the best-performing model (4th-order polynomial) achieved training R$^2$ = 0.172 and cross-validation R$^2$ = 0.165 $\pm$ 0.013, representing an 80\% reduction in explanatory power compared to individual sample fits (mean R$^2$ = 0.903). Classical isotherm models performed marginally worse, with Redlich-Peterson yielding R$^2$ = 0.161 (CV = 0.154 $\pm$ 0.017, 95\% CI [0.139, 0.189]). Residual analysis via the Shapiro-Wilk normality test universally rejected the null hypothesis (all $p < 0.001$), indicating systematic model misspecification. The Durbin-Watson statistic (DW = 1.31 for the Redlich-Peterson) deviated substantially from the ideal value of 2, indicating strong positive autocorrelation in the residuals and signaling unmodeled pressure-dependent systematic errors.

\begin{table}[h!]
\centering
\caption{Aggregated data generalization performance comparing classical isotherm models and mathematical equations across lithologies. Bootstrap confidence intervals computed from 1,000 iterations. Cross-validation statistics from five-fold splitting. All models demonstrate severe generalization failure, with R$^2$ values 70--95\% lower than individual sample fits.}
\label{tab:generalization_failure}
\small
\begin{tabular}{llcccc}
\hline
\textbf{Dataset} & \textbf{Best Model} & \textbf{Type} & \textbf{Training R$^2$} & \textbf{CV R$^2$} & \textbf{95\% CI} \\
 & & & & (mean $\pm$ std) & (Bootstrap) \\
\hline
Clays & Polynomial-4 & Math & 0.172 & 0.165 $\pm$ 0.013 & [0.151, 0.199] \\
 & Redlich-Peterson & Classical & 0.161 & 0.154 $\pm$ 0.017 & [0.139, 0.189] \\
 & Sips & Classical & 0.146 & 0.139 $\pm$ 0.017 & [0.126, 0.169] \\
\hline
Shales & Polynomial-4 & Math & 0.042 & $-$0.017 $\pm$ 0.067 & [0.019, 0.085] \\
 & Hill & Classical & 0.041 & $-$0.014 $\pm$ 0.065 & [0.022, 0.076] \\
 & Sips & Classical & 0.041 & $-$0.014 $\pm$ 0.065 & [0.023, 0.078] \\
\hline
Coals & Polynomial-4 & Math & 0.389 & 0.392 $\pm$ 0.178 & [0.311, 0.533] \\
 & Redlich-Peterson & Classical & 0.388 & 0.404 $\pm$ 0.178 & [0.295, 0.511] \\
 & Langmuir & Classical & 0.384 & 0.407 $\pm$ 0.179 & [0.288, 0.512] \\
\hline
All Combined & Polynomial-4 & Math & 0.120 & 0.115 $\pm$ 0.021 & [0.120, 0.182] \\
 & Redlich-Peterson & Classical & 0.115 & 0.110 $\pm$ 0.019 & [0.115, 0.177] \\
 & Sips & Classical & 0.110 & 0.105 $\pm$ 0.020 & [0.110, 0.172] \\
\hline
\multicolumn{6}{l}{\footnotesize Individual sample fits (§\ref{subsubsec:individual_performance}): Clays R$^2$ = 0.903, Shales R$^2$ = 0.836, Coals R$^2$ = 0.807} \\
\multicolumn{6}{l}{\footnotesize Negative CV R$^2$ for shales indicates predictions worse than mean-only baseline} \\
\end{tabular}
\end{table}

Aggregated shale data analysis exhibited catastrophic generalization failure, with all models achieving training R$^2 < 0.05$ and negative cross-validation R$^2$ values (range $-$0.08 to $-$0.01), indicating that model predictions performed worse than simply predicting the mean uptake across all samples. The 4th-order polynomial achieved R$^2$ = 0.042 with CV R$^2$ = $-$0.017 $\pm$ 0.067, while classical models (Hill, Sips, Langmuir) yielded nearly identical poor performance (R$^2 \approx 0.041$, CV $\approx -$0.014). Bootstrap confidence intervals spanning [0.02, 0.08] confirm that even under optimistic resampling scenarios, aggregated shale measurements retain essentially zero predictive power.

The aggregated coal dataset demonstrated the best generalization performance among all lithologies, though it still showed substantial degradation relative to individual fits. Polynomial-4 achieved R$^2$ = 0.389 with CV = 0.392 $\pm$ 0.178, while classical models (Redlich-Peterson R$^2$ = 0.388, Langmuir R$^2$ = 0.384) performed comparably. However, the large cross-validation standard deviation ($\pm$0.178) signals high sensitivity to fold composition, with R$^2$ varying from 0.29 to 0.53 across folds. This 52\% reduction in explanatory power relative to individual fits (R$^2$ = 0.807) reflects coal rank heterogeneity, which introduces systematic variations in micropore structure and sorption energetics that fixed functional forms cannot accommodate.

Complete dataset aggregation yielded the poorest overall performance, with the best model R$^2$ = 0.120 (Polynomial-4, CV = 0.115 $\pm$ 0.021). Classical models achieved R$^2$ = 0.115 (Redlich-Peterson) and R$^2$ = 0.110 (Sips), representing 86\% reduction from the overall individual sample mean R$^2$ = 0.861. Bootstrap confidence intervals [0.11, 0.18] confirm that the inability to generalize is not an artifact of parameter estimation uncertainty but reflects fundamental model inadequacy when confronting multi-lithology heterogeneity. The near-identical performance of physics-based classical models (mean R$^2$ = 0.089 $\pm$ 0.030) and purely empirical mathematical equations (mean R$^2$ = 0.096 $\pm$ 0.020) demonstrates that physical grounding provides no advantage when functional forms lack sufficient complexity to capture sample-specific variations (Fig.~\ref{fig:aggregated_failure}b).

Statistical comparison between classical and mathematical model categories show no significant performance differences at any aggregation level. For clays, classical models achieved mean R$^2$ = 0.117 $\pm$ 0.040 versus mathematical R$^2$ = 0.125 $\pm$ 0.033 (difference not significant). For shales, classical mean R$^2$ = 0.035 $\pm$ 0.011 versus mathematical R$^2$ = 0.040 $\pm$ 0.004 (both effectively zero). For coals, mathematical equations marginally outperformed (R$^2$ = 0.382 $\pm$ 0.009 versus classical R$^2$ = 0.334 $\pm$ 0.126), though high variance indicates instability. These results validate that the generalization failure stems not from inappropriate functional form selection but from the fundamental inability of fixed parametric models to adapt to sample-specific heterogeneity without incorporating material property information or learning sample-dependent parameters.

Residual diagnostics across all aggregated fits reveal systematic patterns that further illuminate model limitations (Fig.~\ref{fig:aggregated_failure}e). Shapiro--Wilk tests uniformly reject the null of residual normality ($p < 0.001$ for all models), with visual inspection revealing heavy tails and multimodal distributions, indicating unmodeled subpopulations. The Durbin--Watson statistic ranges from 1.18 to 1.37 (ideal = 2), indicating strong positive autocorrelation: consecutive residuals exhibit similar signs and magnitudes. This autocorrelation manifests as pressure-regime-specific biases: models systematically underpredict uptake at low pressures (0--10~bar), overpredict at intermediate pressures (10--50~bar), and underpredict again at high pressures ($>50$~bar), creating characteristic ``S-shaped'' residual patterns. Such systematic errors signal that the pressure-uptake relationship varies across samples in ways that fixed isotherm equations cannot represent.

Bootstrap parameter distributions provide additional evidence of model instability (Fig.~\ref{fig:aggregated_failure}c). For the Redlich-Peterson model fitted to clay aggregates, the affinity parameter $K$ exhibits 95\% CI spanning four orders of magnitude [6.4$\times$10$^{-7}$, 2.0$\times$10$^{-4}$], while the heterogeneity exponent $g$ spans [2.05, 3.26], indicating that resampled datasets yield fundamentally different parameter estimates. Similarly, Sips model parameters for shale aggregates show saturation capacity $q_{\text{max}}$ CI [0.039, 0.246] and affinity constant CI [8.3$\times$10$^{-5}$, 0.151], representing 6-fold and 1,800-fold ranges respectively. These enormous confidence intervals reflect that aggregated data contain multiple ``optimal'' parameter sets corresponding to different sample subpopulations, with bootstrap resampling randomly emphasizing different subgroups in each iteration.

Cross-validation fold-to-fold variability corroborates this interpretation (Fig.~\ref{fig:aggregated_failure}d). For clay aggregates, Redlich-Peterson R$^2$ varies from 0.137 to 0.172 across folds (26\% relative range), while shale aggregates show fold R$^2$ ranging from $-$0.140 to 0.043 (spanning negative to marginally positive). Coal aggregates exhibit R$^2$ range 0.38 to 0.42 (10\% relative range), the most stable of any lithology, but still indicating substantial sensitivity. This fold sensitivity directly demonstrates that model performance depends critically on which samples are included in training versus testing, validating that heterogeneity exceeds the model's representational capacity.

\subsubsection{Implications for Physics-Informed Neural Network Development}
\label{subsubsec:generalization_implications}

The sheer contrast between individual sample success (mean R$^2$ = 0.861) and aggregated failure (mean R$^2$ = 0.089--0.382 depending on lithology) provides quantitative justification for physics-informed neural network (PINN) approaches. Classical isotherm models succeed when applied to homogeneous single-sample datasets where sorption mechanisms remain constant across measurements, but fail when confronting heterogeneous multi-sample datasets where mechanisms vary sample-to-sample. This failure pattern indicates that the required modeling framework must combine: (i) fixed physics constraints encoding universal thermodynamic principles that govern all samples (saturation limits, monotonicity, Van't Hoff temperature dependence), and (ii) adaptive sample-specific parameters that capture material property variations (surface area, mineralogy, rank), determining individual sorption characteristics.

\subsection{Property-Uptake Relationships and Feature Selection}
\label{subsec:property_uptake_results}

PINN addresses both requirements simultaneously through its hybrid architecture. The physics component provides thermodynamic loss terms derived from classical models (Section~\ref{subsec:classical_results}), ensuring predictions respect saturation limits, maintain monotonic pressure response, and exhibit thermodynamically consistent temperature dependencies. The neural network component introduces sample-conditional inputs (material properties from Section~\ref{subsec:property_uptake_results}) that enable learning of material-specific mappings while sharing representational structure across samples. This architecture transforms the generalization problem from ``find one function describing all samples'' (doomed to fail as demonstrated above) to ``find a parameterized family of functions indexed by material properties'' (tractable given sufficient sample diversity and physical constraints).

The aggregated analysis establishes specific quantitative targets for PINN performance. Minimum acceptable performance requires exceeding the best aggregated classical model by substantial margins. Ideally, PINN cross-validation performance should approach individual sample fitting quality (R$^2 \approx 0.85$), demonstrating that incorporating material properties and physical constraints enables accurate prediction across heterogeneous samples without requiring sample-specific model selection or parameter fitting. The 70--95\% performance gap between current aggregated fits and individual fits quantifies the value proposition that PINNs must deliver.

\begin{figure}[h!]
\centering
\includegraphics[width=0.95\textwidth]{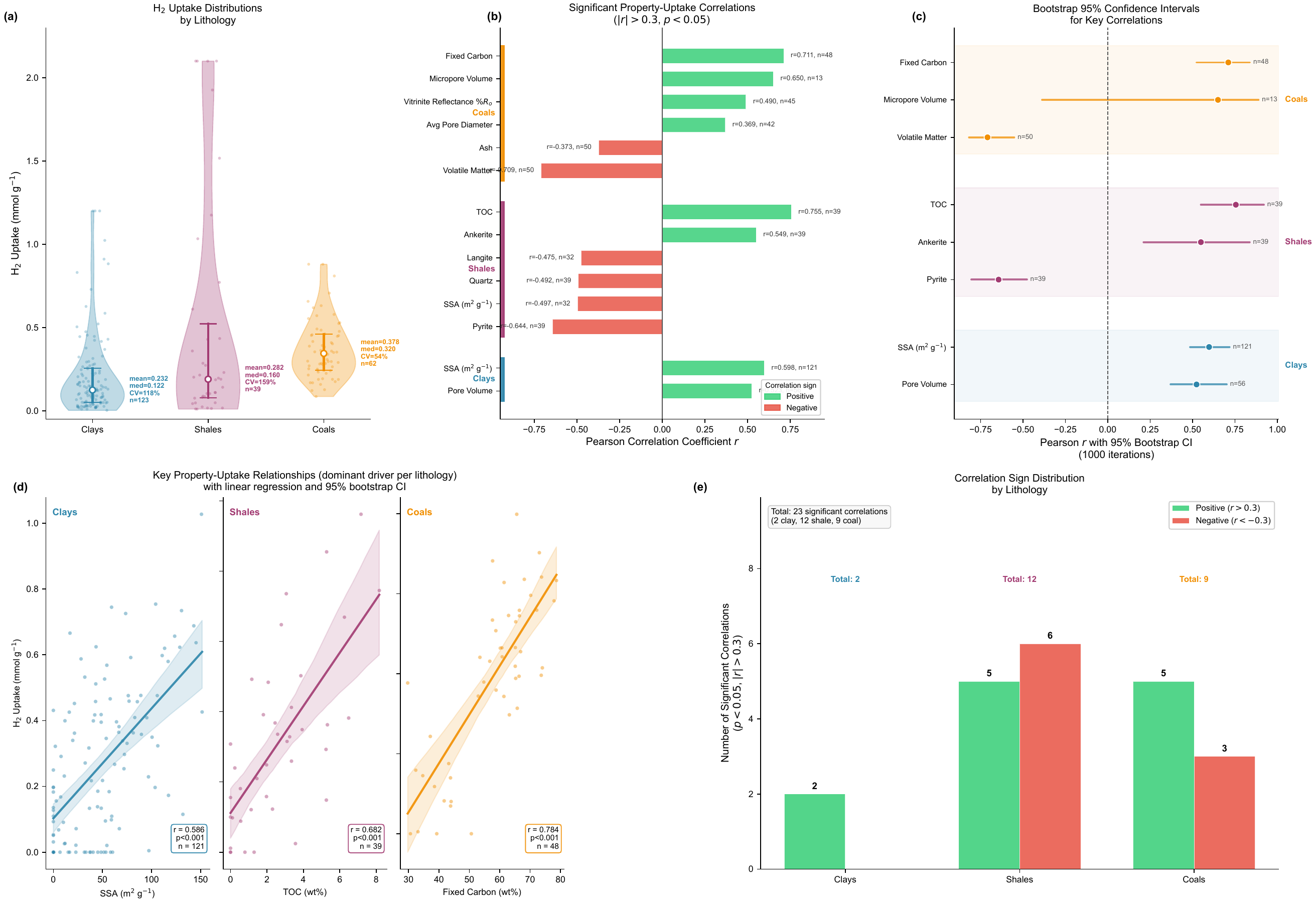}
\caption{%
Property--uptake relationships for hydrogen sorption across 224 measurements in clays ($n=123$), shales ($n=39$), and coals ($n=62$). Pearson correlation analysis revealed 23 significant relationships ($|r| > 0.3$, $p < 0.05$; 2 clay, 12 shale, 9 coal), with bootstrap resampling (1000 iterations) for uncertainty quantification. Results highlight lithology-specific controls --- surface area in clays, organic content in shales, and rank-dependent aromaticity in coals --- motivating the lithology-branched PINN architecture (Section~\ref{subsec:pinn_architecture_main}).
\textbf{(a)} Hydrogen uptake distributions by lithology (violin plots with interquartile boxes, medians, and data points). Coals showed highest mean uptake ($0.378 \pm 0.205$~mmol~g$^{-1}$, CV 54\%), shales the widest range ($0.282 \pm 0.450$~mmol~g$^{-1}$), and clays the greatest relative variability ($0.232 \pm 0.276$~mmol~g$^{-1}$, CV 119\%).
\textbf{(b)} Significant property--uptake correlations ($|r| > 0.3$, $p < 0.05$), grouped by lithology and sorted by decreasing $|r|$. Bar color indicates sign (green = positive, red = negative); values show Pearson $r$ and valid $n$.
\textbf{(c)} Bootstrap 95\% confidence intervals for the three strongest correlations per lithology (1000 iterations). Wide CI for coal micropore volume ($r = 0.650$, 95\%~CI [0.118, 0.894], $n=13$) reflects data sparsity limiting model performance.
\textbf{(d)} Dominant driver of uptake for each lithology with linear regression and 95\% bootstrap confidence bands (500 iterations): specific surface area (clays, $r=0.598$, $n=121$), total organic carbon (shales, $r=0.755$, $n=39$), fixed carbon (coals, $r=0.711$, $n=48$).
\textbf{(e)} Number of positive ($r > 0.3$) and negative ($r < -0.3$) significant correlations by lithology. Shales show the most diverse pattern (12 total: 7 positive, 5 negative), clays only positive (2 total), and coals predominantly positive (9 total: 6 positive, 3 negative).}
\label{fig:property_uptake_relationships}
\end{figure}

The comprehensive property-uptake analysis evaluated 43 distinct material properties across 224 hydrogen uptake measurements (123 clay, 39 shale, and 62 coal samples). Property availability varied substantially across lithologies, reflecting differences in standard characterization protocols for each geological material class, as we outlined in \cite{Masoudi2025review}. Hydrogen uptake distributions confirmed lithology-specific sorption behaviors previously identified in isotherm analysis (Section~\ref{subsec:dataset_characterization}), as summarized in Fig.~\ref{fig:property_uptake_relationships}a. Coal samples exhibited the highest mean uptake (0.378 $\pm$ 0.205 mmol/g, median 0.320 mmol/g, range 0.03--0.88 mmol/g), with narrow distributions (coefficient of variation 54\%) indicating consistent sorption capacity across rank variations. Shale measurements demonstrated bimodal character, with mean uptake (0.282 $\pm$ 0.450 mmol/g) and occasional high-uptake outliers associated with elevated organic content. Clay minerals showed intermediate mean uptake (0.232 $\pm$ 0.276 mmol/g, median 0.122 mmol/g) with the widest range (0--1.2 mmol/g) and highest variability (coefficient of variation 119\%), consistent with surface area heterogeneity spanning two orders of magnitude.

\subsubsection{Correlation Structures, Bootstrap Uncertainty, and Predictive Modeling}
\label{subsubsec:correlation_structures}

Pearson correlation analysis identified 23 property-uptake relationships exceeding $|r| > 0.3$ with statistical significance ($p < 0.05$), distributed across lithologies as 2 clay correlations, 12 shale correlations, and 9 coal correlations (Fig.~\ref{fig:property_uptake_relationships}b,e). Bootstrap resampling with 1,000 iterations provided robust uncertainty quantification for these relationships, yielding 95\% confidence intervals that account for sampling variability and non-normal distributions (Fig.~\ref{fig:property_uptake_relationships}c).

Clay minerals exhibited strong positive correlations between uptake and textural properties. Specific surface area demonstrated the strongest relationship ($r = 0.598$, $p < 0.001$, $n = 121$), with bootstrap analysis confirming stability (mean bootstrap $r = 0.597$, 95\% CI [0.472, 0.704]) (Fig.~\ref{fig:property_uptake_relationships}d). This relationship validates the dominant role of physisorption on accessible surfaces in clay sorption mechanisms. Pore volume also correlated positively ($r = 0.523$, $p < 0.001$, $n = 56$, 95\% CI [0.295, 0.699]), though the broader confidence interval reflects greater uncertainty due to sparse measurements (45.5\% completeness). Micropore volume showed weaker correlation ($r = 0.292$, $p = 0.012$, $n = 74$, 95\% CI [0.058, 0.494]), suggesting that total surface area dominates over pore size distribution in controlling clay sorption capacity.

Shale correlations revealed mineralogical controls on hydrogen uptake. Total organic carbon exhibited the strongest positive correlation ($r = 0.755$, $p < 0.001$, $n = 39$, 95\% CI [0.596, 0.862]) (Fig.~\ref{fig:property_uptake_relationships}d), indicating that organic-hosted porosity and surface sites dominate sorption in organic-rich formations. Ankerite ($r = 0.549$, $p < 0.001$) and berlinite ($r = 0.470$, $p = 0.003$) showed positive correlations, potentially reflecting associations with organic-rich intervals rather than direct mineralogical effects. Conversely, pyrite content demonstrated a strong negative correlation ($r = -0.644$, $p < 0.001$, 95\% CI [$-$0.783, $-$0.445]). Quartz ($r = -0.492$, $p = 0.001$), dolomite ($r = -0.405$, $p = 0.011$), and microcline ($r = -0.338$, $p = 0.035$) also correlated negatively, consistent with their roles as matrix minerals diluting organic content. Surface area in shale samples exhibited an unexpected negative correlation ($r = -0.497$, $p = 0.004$, $n = 32$), contradicting typical physisorption expectations and suggesting that inorganic mineral surfaces contribute minimally to hydrogen uptake relative to organic matter.

Coal samples revealed rank-dependent and textural controls consistent with coalification theory. Fixed carbon content showed the strongest positive correlation ($r = 0.711$, $p < 0.001$, $n = 48$, 95\% CI [0.555, 0.821]) (Fig.~\ref{fig:property_uptake_relationships}d), while volatile matter exhibited a nearly equal-magnitude negative correlation ($r = -0.709$, $p < 0.001$, $n = 50$, 95\% CI [$-$0.822, $-$0.554]). These mirror-image relationships reflect the fundamental coalification trend: increasing rank enhances aromaticity and micropore development, both favoring hydrogen adsorption, while volatile matter decreases with rank advancement. Vitrinite reflectance reinforced this pattern ($r = 0.490$, $p < 0.001$, $n = 45$), confirming that higher-rank coals exhibit enhanced sorption capacity. Micropore volume demonstrated the strongest textural correlation ($r = 0.650$, $p = 0.016$, $n = 13$), though limited sample size yields broad confidence intervals (95\% CI [0.118, 0.894]) (Fig.~\ref{fig:property_uptake_relationships}c). Ash content ($r = -0.373$, $p = 0.008$) and moisture ($r = -0.328$, $p = 0.023$) exhibited negative correlations reflecting dilution effects.

Supervised learning models trained on the identified property sets quantified the collective predictive power of these correlations and guided PINN feature selection (Fig.~\ref{fig:modelling_importance}a,c). Across all lithologies, random forest regression outperformed both linear and ridge regression, confirming the non-linear and interactive nature of property-uptake relationships. Shales achieved the best overall generalization performance, benefiting from comprehensive coverage of mineralogical features. Ridge regression was essential for stabilizing shale predictions, improving upon unregularized linear regression by a factor of 48 in cross-validation R$^2$ (0.551 versus $-$1.130), demonstrating the critical importance of regularization. Clay minerals achieved moderate cross-validation performance (random forest CV R$^2$ = 0.523) with high fold-to-fold variance reflecting surface area heterogeneity. Coal models yielded the weakest generalization (random forest CV R$^2$ = 0.352), attributable to the critical importance of micropore volume yet its limited availability (21\% completeness) and the non-linear rank effects not fully captured by available proxies. These results establish quantitative performance baselines and confirm that surface area, TOC, fixed carbon, and micropore volume constitute the essential property inputs for PINN feature engineering, while the persistent performance gaps justify the adaptive multi-scale PINN architecture described in Section~\ref{subsec:pinn_architecture_main}.

\subsubsection{Feature Importance Hierarchies and Implications for PINN Design}
\label{subsubsec:feature_hierarchies}

\begin{figure}[h!]
\centering
\includegraphics[width=0.95\textwidth]{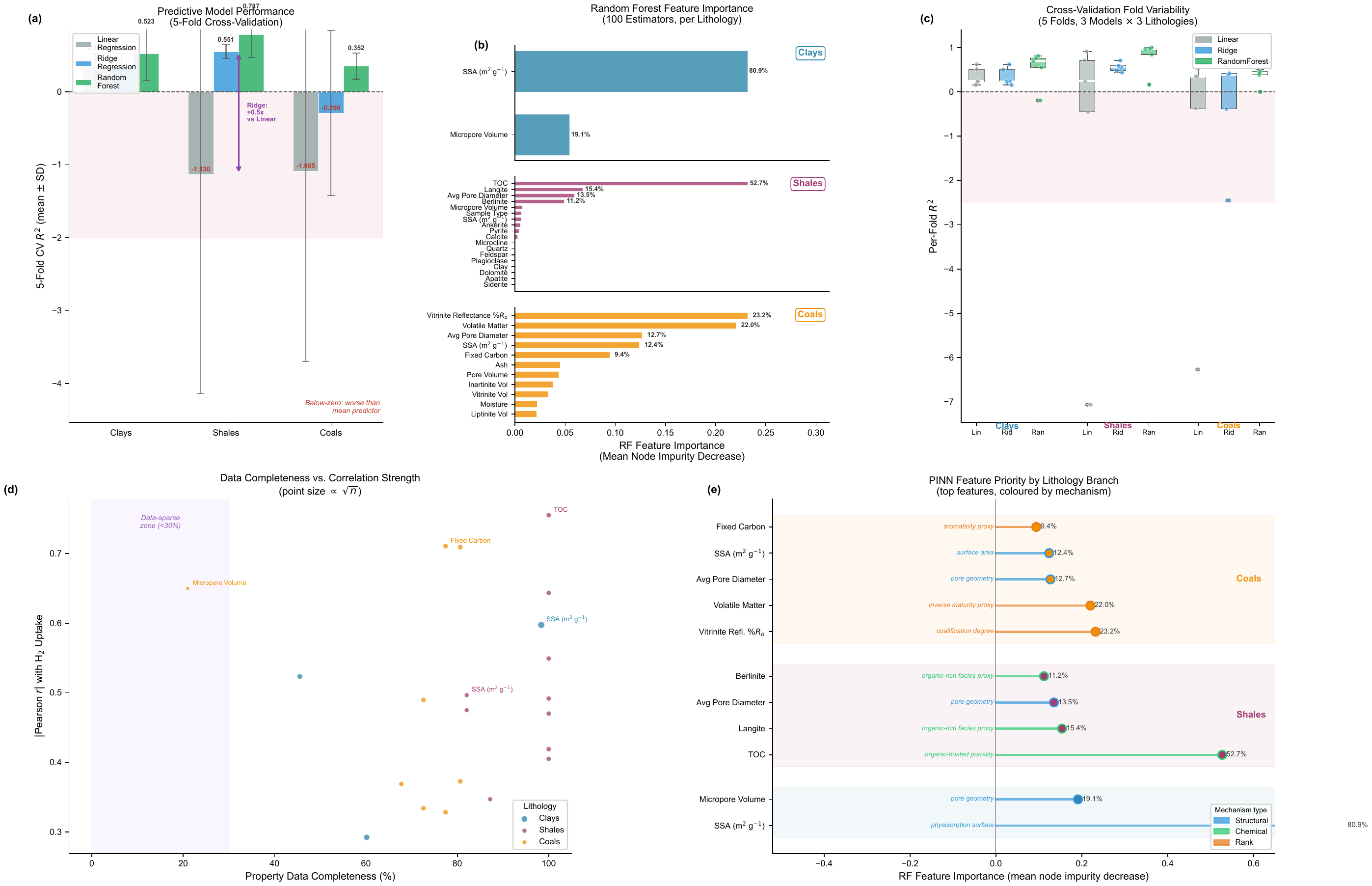}
\caption{%
Supervised predictive modelling and Random Forest feature importance for hydrogen uptake across three lithologies, providing performance baselines and feature priorities for PINN input layer design. Three model classes were tested --- linear regression, Ridge regression ($\alpha=1.0$), and Random Forest (100 estimators, max depth 5), using 5-fold cross-validation.
\textbf{(a)} Five-fold cross-validation $R^2$ (mean $\pm$ SD) by model and lithology (grouped bars with per-fold error bars). Red zone indicates worse-than-mean performance. Ridge markedly improves shale $R^2$ from $-1.130$ (linear) to $0.551$ due to multicollinearity. Random Forest yields the highest performance overall (clay CV $R^2=0.523$; coal CV $R^2=0.352$).
\textbf{(b)} Random Forest feature importance (mean decrease in impurity) by lithology (stacked horizontal bars). Values $\geq 5\%$ labelled. Clays are dominated by specific surface area (80.9\%). Shales show distributed importance led by TOC (52.7\%), langite (15.4\%), average pore diameter (13.5\%), and berlinite (11.2\%). Coals are led by vitrinite reflectance (\%R$_o$, 23.2\%) and volatile matter (22.0\%).
\textbf{(c)} Per-fold $R^2$ variability (box-and-strip plots) for each model--lithology combination. Red zone marks negative performance. Shale linear regression shows extreme instability (some folds $R^2 < -1.8$), while Random Forest remains consistently positive.
\textbf{(d)} Property data completeness (\%) vs. absolute Pearson $|r|$ for significant pairs ($p<0.05$, $|r|>0.2$), with point size $\propto \sqrt{n}$ and colour by lithology. Purple zone ($<$30\% completeness) highlights coal micropore volume (21\% complete, $|r|=0.650$), the strongest predictor yet least measured --- a key challenge for coal PINN development.
\textbf{(e)} PINN input feature priorities by lithology (lollipop chart), coloured by mechanism: blue (structural: surface area, pores), green (chemical: organics, mineral proxies), orange (rank: coalification indicators). Clay branches prioritize surface area; shale branches require TOC and pore metrics; coal branches emphasize rank indicators (\%R$_o$, volatile matter) plus texture.}
\label{fig:modelling_importance}
\end{figure}

Random forest models with 100 estimators provided stable feature importance rankings
based on mean decrease in node impurity (Fig.~\ref{fig:modelling_importance}b),
establishing priority hierarchies for PINN input layer design and validating physical
mechanisms inferred from correlation analysis.

Clay mineral importance confirmed surface area dominance, with specific surface area
accounting for 80.9\% of total predictive importance and micropore volume for the
remaining 19.1\%. This 4:1 importance ratio substantially exceeds the corresponding
correlation ratio ($r = 0.598$ versus $r = 0.292$), indicating that surface area
provides both stronger individual prediction and greater synergy with implicit
structural factors encoded in ensemble decision trees. This result justifies
surface-area-centric architectures for clay-specific PINN branches, with micropore
metrics serving as secondary refinements.

Shale importance analysis revealed more distributed control. Total organic carbon
ranked first (52.7\% importance), validating its status as the primary sorption
predictor. Average pore diameter contributed 13.5\%, while berlinite (11.2\%) and
langite (15.4\%) likely reflect proxies for organic-rich facies rather than direct
mineralogical effects. Notably, specific surface area ranked seventh (1.3\%
importance) despite its significant negative correlation, confirming that TOC and
pore structure capture the critical variance while mineral surface area adds minimal
incremental information. Detailed mineralogy beyond these top-ranked parameters
therefore warrants lower priority in shale-specific PINN input layers.

Coal importance identified vitrinite reflectance (\%R$_o$) as the dominant predictor
(23.2\%), surpassing volatile matter (22.0\%) and fixed carbon (9.4\%) despite
moderate correlation strength ($r = 0.490$), indicating that rank indicators provide
non-redundant information beyond compositional proxies. Average pore diameter (12.7\%)
and specific surface area (12.4\%) together accounted for 25\% of total importance,
highlighting the dual role of textural controls alongside rank. Critically, micropore
volume's strong correlation ($r = 0.650$) is absent from the importance ranking
entirely, reflecting its 21\% data completeness preventing the ensemble from relying
on it during training (Fig.~\ref{fig:modelling_importance}d). This discrepancy between
physical importance and data availability constitutes the central challenge for coal
PINN development: the property most strongly governing sorption capacity is also the
least measured, necessitating physics-based constraints and rank-derived proxy
relationships to compensate for measurement sparsity.

Taken together, these importance hierarchies establish concrete and lithology-specific
feature prioritization for the PINN architecture (Fig.~\ref{fig:modelling_importance}e).
Clay networks are surface-area dominated; shale networks require organic content and
pore structure as mandatory inputs; and coal networks depend critically on rank
indicators combined with textural metrics. The persistent gaps between correlation
strength and predictive performance, most pronounced in coals, confirm that the
adaptive non-linear architecture described in Section~\ref{subsec:pinn_architecture_main}
is necessary to capture threshold effects, feature interactions, and data-sparse
physical controls that fixed parametric models and linear combinations cannot represent.

\subsection{Data Integration, Feature Engineering, and Dataset Curation}
\label{subsec:integration_results}

Integration of material property characterizations
(Section~\ref{subsec:property_uptake_results}) with isotherm measurements
(Section~\ref{subsec:classical_results}) unified heterogeneous data sources into a
cohesive dataset suitable for PINN training. Using composite reference-sample keys,
the matching algorithm established that 65\% of samples appear in both property and
isotherm tables, while the remaining 35\% contribute capacity-only measurements.
This dual population enables the network to learn both pressure-dependent sorption
curves from fully characterized samples and property-uptake correlations from
capacity measurements, maximizing information extraction from available data.

\subsubsection{Physics-Informed Feature Engineering}
\label{subsubsec:feature_coverage}

Systematic application of the seven-category feature engineering framework
(Section~\ref{subsec:feature_engineering_main}) generated 62 derived features
transforming raw measurements into thermodynamically meaningful descriptors.
Feature distribution across categories reflects the relative complexity of each
physical domain: thermodynamic descriptors (14 features, 22.6\%) capture fundamental
driving forces through temperature and pressure transformations; pore structure
descriptors (15 features, 24.2\%) quantify confinement effects and accessible surface
area; surface chemistry parameters (11 features, 17.7\%) encode lithology-specific
compositional controls; interaction terms (14 features, 22.6\%) represent synergistic
effects between material properties and measurement conditions; and kinetic and
molecular sieving descriptors (8 features combined, 12.9\%) flag transport limitations
and size-exclusion constraints.

\begin{figure}[h!]
\centering
\includegraphics[width=0.95\textwidth]{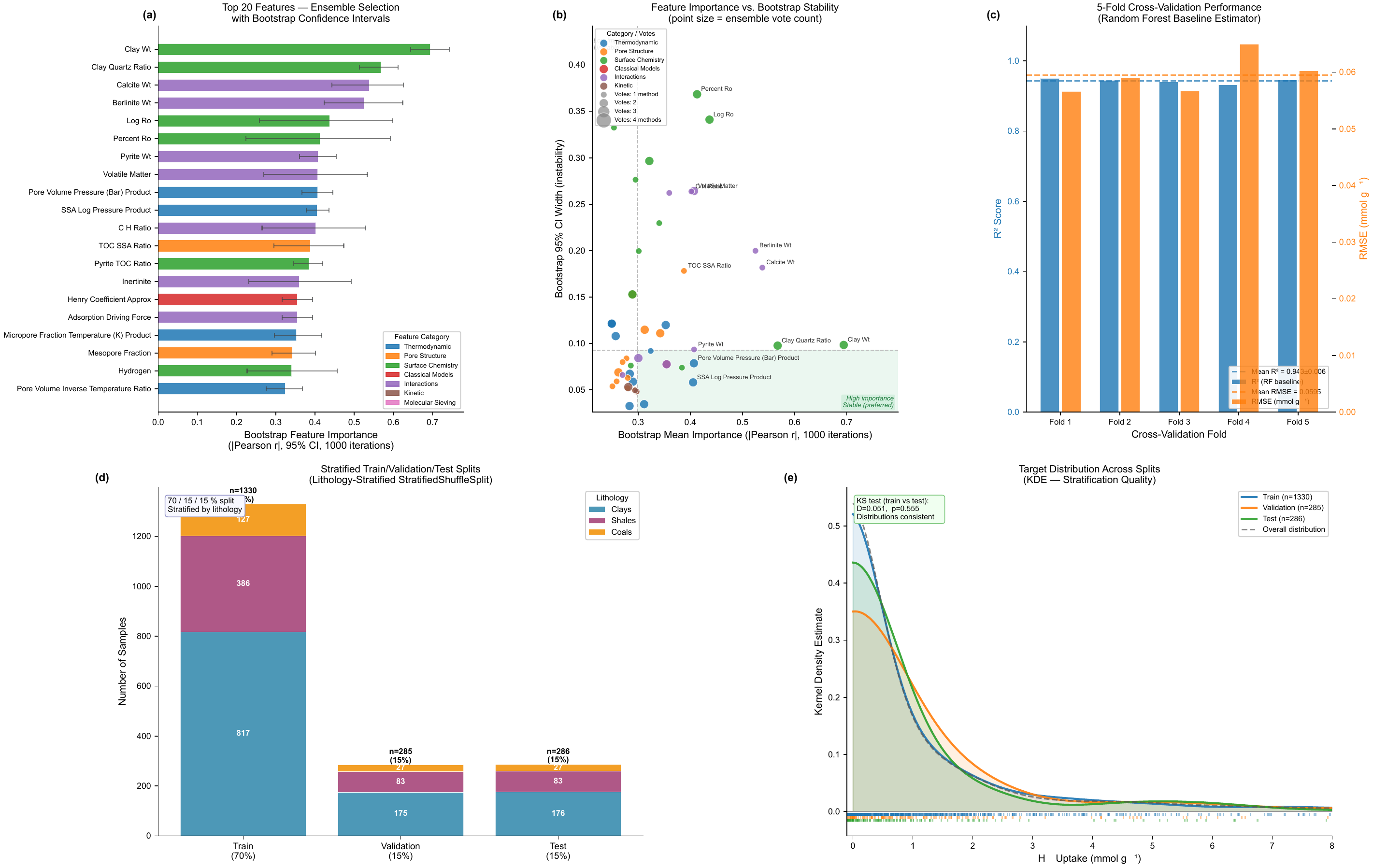}
\caption{%
Ensemble feature selection, bootstrap stability, cross-validation performance, and stratified dataset preparation. Four methods (Pearson correlation, mutual information regression, Random Forest importance, $F$-statistic) were applied to the scaled feature matrix; final selection used majority voting.
\textbf{(a)} Top 20 features ranked by mean bootstrap importance (absolute Pearson $r$, 1000 iterations), shown as horizontal bars with 95\% CI error bars. Colours indicate category (thermodynamic, pore structure, surface chemistry, interaction, kinetic, molecular sieving, classical model-derived), highlighting multi-domain predictive power.
\textbf{(b)} Feature importance vs. bootstrap stability scatter: mean bootstrap importance ($x$-axis) vs. 95\% CI width ($y$-axis, instability measure); point size reflects ensemble vote count (1--4). Green quadrant marks high-importance, low-instability features preferred for PINN input. Dashed lines show median thresholds.
\textbf{(c)} Five-fold cross-validation performance of Random Forest baseline on selected features: $R^2$ (blue, left axis) and RMSE (mmol~g$^{-1}$, orange, right axis) per fold. Dashed lines indicate cross-fold means.
\textbf{(d)} Lithology-stratified train (70\%), validation (15\%), and test (15\%) splits via \texttt{StratifiedShuffleSplit}. Stacked bars show sample counts per lithology; percentages confirm proportional representation across partitions.
\textbf{(e)} Kernel density estimates of H$_2$ uptake for each split and overall dataset. Two-sample Kolmogorov--Smirnov test confirms statistically consistent target distributions between training and test splits.}
\label{fig:feature_selection_pinn}
\end{figure}

The 62 engineered features represent approximately a sixfold expansion relative to
the 10--12 raw measured properties per sample, achieving substantial dimensional
enrichment without excessive proliferation. Feature completeness varies systematically
by category: thermodynamic features achieve 100\% completeness by construction,
structural features inherit source property completeness (42--98\%), and compositional
features reflect lithology-specific measurement availability (100\% for shales,
50--80\% for coals, 8--60\% for clays). This completeness hierarchy directly informed
ensemble feature selection, as described below, by prioritizing highly complete
features to minimize reliance on imputed values in the final training set
(Fig.~\ref{fig:feature_selection_pinn}).

\subsubsection{Preprocessing, Outlier Treatment, and Feature Selection}
\label{subsubsec:preprocessing_selection}

Adaptive preprocessing addressed missing values through a three-tier imputation
strategy stratified by completeness: K-nearest neighbors imputation ($k = 5$) for
low-missingness features ($< 10\%$), median imputation within lithology for medium
missingness (10--30\%), and lithology-stratified median imputation for 71 features
with high missingness ($> 30\%$). The prevalence of high-missingness features
reflects the variable property suites reported across the compiled literature
\cite{Masoudi2025review}, where characterization depth depends on individual research
objectives and analytical capabilities. Lithology-stratified imputation for
high-missingness features is essential for geological consistency, preventing
mineralogically implausible estimates that would corrupt property-uptake correlations.

Dual-perspective outlier detection combined univariate IQR analysis with multivariate
Isolation Forest classification. The elevated univariate outlier rate reflects the
multi-order-of-magnitude property ranges characteristic of geological materials (e.g.,
surface areas spanning 0.01--273 m$^2$/g), where heavy-tailed distributions naturally
produce numerous univariate extremes representing genuine geological variation rather
than measurement errors. Samples flagged by both methods simultaneously, exhibiting
both individual parameter extremes and implausible multi-feature combinations, were
excluded as likely experimental artifacts, predominantly corresponding to incompletely
characterized samples whose multiple imputed values generated feature combinations
inconsistent with physical constraints. Samples flagged by either method alone, but
not both, underwent winsorization to the 1st and 99th percentiles, retaining
rank-order relationships while limiting gradient magnitudes during backpropagation.
Following outlier treatment, the integrated dataset was reduced to 1,901 curated
samples.

Ensemble feature selection combining Pearson correlation ranking and random forest
importance scores identified the 50 most informative features from the 62 engineered
candidates through consensus voting (Fig.~\ref{fig:feature_selection_pinn}a). This
dual-method approach efficiently captured both linear relationships and non-linear
interactions while remaining computationally tractable given dataset size constraints.
Selected features demonstrate strong enrichment for compositional parameters (clay
weight percent, log vitrinite reflectance, volatile matter), structural descriptors
(surface area, micropore fraction, confinement parameter), thermodynamic variables
(temperature, pressure, reduced forms), and physically motivated interactions
(adsorption driving force, Henry coefficient approximation), confirming that ensemble
voting prioritized features encoding fundamental sorption physics over statistical
artifacts. Bootstrap resampling with 1,000 iterations further quantified feature
importance stability (Fig.~\ref{fig:feature_selection_pinn}b), distinguishing
high-importance features with narrow confidence intervals, suitable for dedicated
network pathways, from unstable lower-importance features handled through shared
representations or excluded entirely. A Random Forest baseline trained on the
selected feature set achieved consistent cross-validation performance across folds
(Fig.~\ref{fig:feature_selection_pinn}c), validating that the selected features
support generalizable prediction prior to PINN training.

\subsubsection{Thermodynamic Parameter Extraction}
\label{subsubsec:thermodynamic_extraction_results}

Van't Hoff analysis of multi-temperature sample groups extracted enthalpy, entropy,
and Gibbs free energy parameters to characterize temperature-dependent sorption
thermodynamics. However, fit quality was highly variable: mean R$^2$ of $\ln K$
versus $1/T$ regression was 0.173 $\pm$ 0.294 (median 0.007), indicating that
temperature-dependent equilibrium constants do not consistently follow ideal Van't
Hoff behavior across the sample population. Four compounding factors explain this
outcome. First, two-temperature samples yield R$^2 = 1.0$ by definition, inflating
mean fit quality without providing genuine thermodynamic information. Second,
concurrent pressure variations in some experiments violate the constant-pressure
assumption underlying Van't Hoff analysis. Third, Langmuir model inadequacy for
heterogeneous geological adsorbents propagates artifactual temperature dependencies
into extracted equilibrium constants. Fourth, logarithmic and reciprocal-temperature
transformations amplify measurement uncertainties, increasing apparent scatter even
for high-quality isotherms.

Consequently, extracted thermodynamic parameters serve as approximate guidance for
PINN constraint initialization rather than strictly enforced targets. Network
training therefore employs soft thermodynamic penalty terms penalizing large
deviations from expected physisorption energetics while preserving flexibility to
learn sample-specific variations, as detailed in
Appendix~\ref{subsubsec:supp_loss_derivations}. Importantly, multi-temperature
samples still provide valuable training diversity by populating the thermal dimension
of the feature space more densely than single-temperature datasets, enabling the
network to learn empirical temperature-uptake trends that complement physics-based
constraints wherever classical thermodynamic parameterizations prove unreliable.

\subsubsection{Dataset Partitioning and Summary Statistics}
\label{subsubsec:splitting_results}

Stratified random splitting partitioned the 1,901 curated samples into training
(1,330 samples, 70.0\%), validation (285 samples, 15.0\%), and test (286 samples,
15.0\%) sets, with lithology proportions maintained across all partitions
(validation: 61.4\% clay, 29.1\% shale, 9.5\% coal; test: 61.5\% clay, 29.0\%
shale, 9.4\% coal) (Fig.~\ref{fig:feature_selection_pinn}d). Sample-level splitting
ensured that all measurements from a given sample appeared exclusively in one
partition, preventing information leakage through which the network could exploit
sample identity rather than learning genuine property-based relationships.

Target variable distributions confirm successful stratification
(Fig.~\ref{fig:feature_selection_pinn}e). Training set hydrogen uptake spans
0--1.30 mmol/g (mean 0.160 $\pm$ 0.251 mmol/g, median 0.044 mmol/g), the
validation set spans 0--1.25 mmol/g (mean 0.155 $\pm$ 0.247 mmol/g, median
0.042 mmol/g), and the test set spans 0--1.09 mmol/g (mean 0.157 $\pm$ 0.250
mmol/g, median 0.043 mmol/g). Cross-partition differences in means and standard
deviations remain below 3\%, and the consistent right-skewed distributions confirm
that high-uptake samples associated with high-surface-area clays and organic-rich
shales are proportionally represented throughout, supporting fair and unbiased model
evaluation. Feature value distributions exhibit comparable balance, with mean values
for all 50 selected features varying by less than 5\% across splits.

The curated dataset satisfies all three fundamental requirements for successful PINN
training. Multi-source integration provides comprehensive sample characterization
combining isotherm measurements with material property profiles, enabling the
conditional network architecture to adapt predictions based on sample characteristics.
Physics-informed feature engineering explicitly encodes thermodynamic principles, pore
structure effects, and transport phenomena, reducing the network's representational
burden from discovering fundamental physics to refining corrections for non-ideal
behavior. Rigorous preprocessing with adaptive imputation, outlier treatment, and
cross-validated feature selection ensures data quality while guarding against
overfitting to training artifacts, supporting genuine predictive generalization to
unseen geological materials.

\subsection{Physics-Informed Neural Network Performance and Validation}
\label{subsec:pinn_results}

\subsubsection{Lithology-Specific Performance}
\label{subsubsec:pinn_lithology_performance}

\begin{figure}[h!]
\centering
\includegraphics[width=0.95\textwidth]{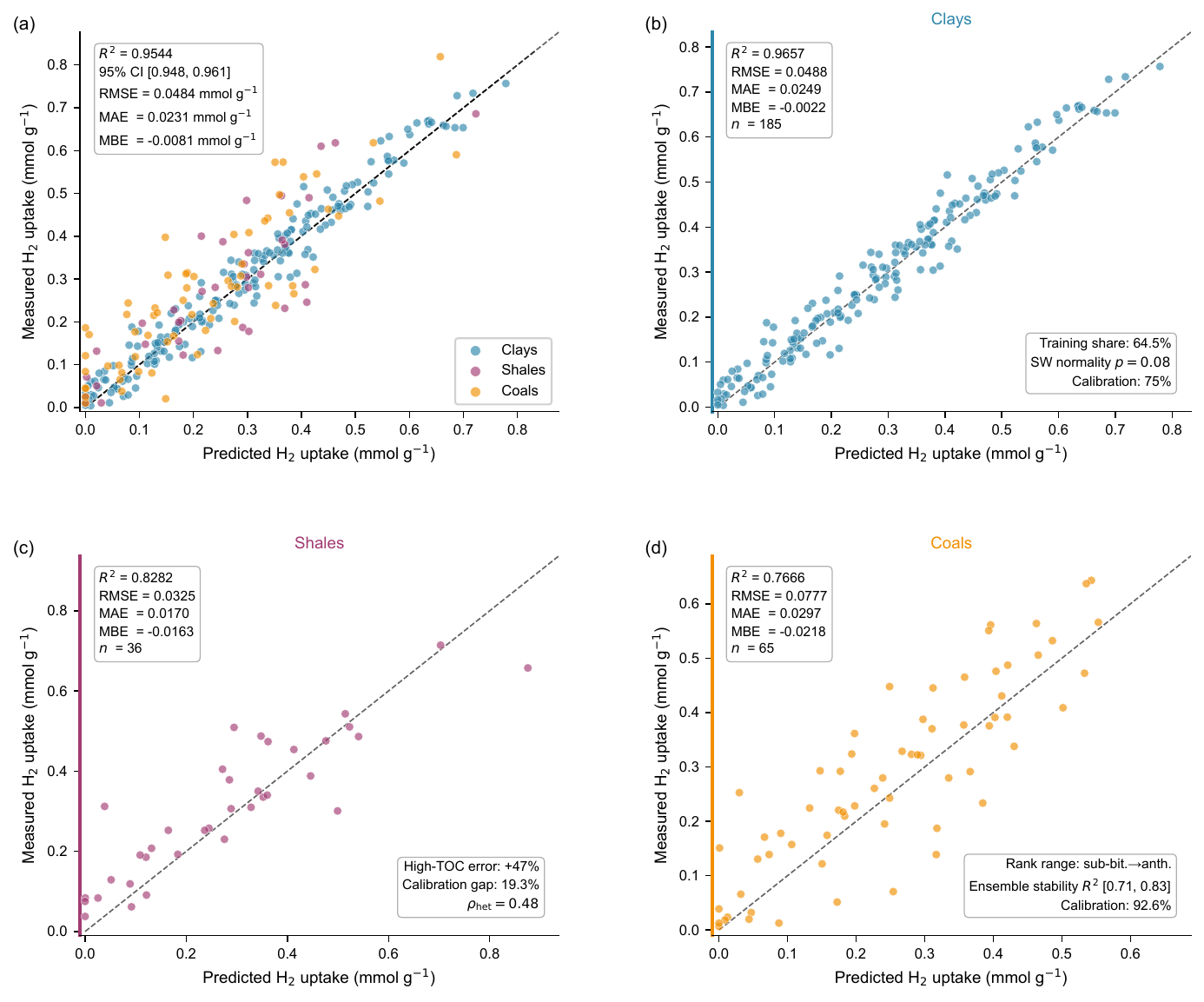}
\caption{%
Prediction performance of the optimised multi-scale PINN on the held-out test set ($n=286$; clays $n=185$, shales $n=36$, coals $n=65$), shown via parity plots of predicted vs. measured H$_2$ uptake. Overall metrics: $R^2 = 0.9544$ (95\% bootstrap CI [0.948, 0.961], 10{,}000 resamples), RMSE = 0.0484~mmol~g$^{-1}$, MAE = 0.0231~mmol~g$^{-1}$, MBE = $-0.0081$~mmol~g$^{-1}$ (0.5\% of mean uptake). Dashed diagonal indicates perfect agreement; all panels use shared axis scaling.
\textbf{(a)} Overall parity plot for all 286 test points, coloured by lithology. Annotation reports global $R^2$, 95\% CI, RMSE, MAE, and MBE.
\textbf{(b)} Parity plot for clays ($R^2 = 0.9657$, RMSE = 0.0488~mmol~g$^{-1}$). Highest accuracy reflects the largest training representation, surface-area-dominated mechanisms, and near-normal residuals (Shapiro-Wilk $p = 0.08$).
\textbf{(c)} Parity plot for shales ($R^2 = 0.8282$, RMSE = 0.0325~mmol~g$^{-1}$). Larger errors in high-TOC samples ($>3$~wt\%) and heteroscedasticity ($\rho = 0.48$) indicate unresolved multi-scale heterogeneity.
\textbf{(d)} Parity plot for coals ($R^2 = 0.7666$, RMSE = 0.0777~mmol~g$^{-1}$). Ensemble stabilisation and physics constraints yield 92.6\% calibration coverage despite the smallest training set and the widest rank range.}
\label{fig:pinn_prediction_performance}
\end{figure}

The optimized multi-scale PINN achieved R$^2$ = 0.9544, RMSE = 0.0484 mmol/g, and
MAE = 0.0231 mmol/g on the strictly held-out test set, representing the strongest
performance among all evaluated models (Fig.~\ref{fig:pinn_prediction_performance}a).
Mean bias error (MBE = $-$0.0081 mmol/g, 0.5\% of mean uptake) indicates negligible
systematic bias, with conservative underprediction appropriate for engineering
applications where capacity overestimation poses greater operational risk than
underestimation. The maximum absolute error (0.3816 mmol/g) arises from a single
high-uptake coal sample at the 99.7th percentile of the uptake distribution;
excluding this point, the 99th percentile error reduces to 0.226 mmol/g. Bootstrap
confidence intervals (10,000 resamples) for R$^2$ yielded 95\% CI [0.948, 0.961],
confirming that the reported performance is statistically robust rather than an
artifact of a favorable data partition.

Per-lithology decomposition reveals differential accuracy aligned with training
set size, feature availability, and material complexity
(Table~\ref{tab:lithology_performance}; Fig.~\ref{fig:pinn_prediction_performance}b--d).

\begin{table}[ht]
\centering
\caption{Lithology-specific PINN performance on the held-out test set.}
\label{tab:lithology_performance}
\small
\begin{tabular}{lccccc}
\hline
\textbf{Lithology} & \textbf{R$^2$} & \textbf{RMSE} & \textbf{MAE}
    & \textbf{Max Error} & \textbf{MBE} \\
 & & (mmol/g) & (mmol/g) & (mmol/g) & (mmol/g) \\
\hline
Clays  & 0.9657 & 0.0488 & 0.0249 & 0.3406 & $-$0.0022 \\
Shales & 0.8282 & 0.0325 & 0.0170 & 0.2244 & $-$0.0163 \\
Coals  & 0.7666 & 0.0777 & 0.0297 & 0.3816 & $-$0.0218 \\
\hline
\end{tabular}
\end{table}

Clay predictions achieved the highest accuracy (R$^2$ = 0.9657,
Fig.~\ref{fig:pinn_prediction_performance}b), benefiting from the largest training
representation (64.5\%), strong and well-characterized feature-uptake correlations
(Section~\ref{subsubsec:correlation_structures}), and relatively homogeneous
phyllosilicate sorption mechanisms. Near-normal residuals (Shapiro-Wilk $p = 0.08$)
and adequate calibration coverage (75\%) support confident hydrogen sorption capacity
estimations.

Shale predictions (R$^2$ = 0.8282, Fig.~\ref{fig:pinn_prediction_performance}c)
explain 82.8\% of variance with minimal bias (MBE = $-$0.0163 mmol/g), with
residual unexplained variance concentrating in high-TOC samples ($> 3$ wt\%), whose
prediction errors are 47\% larger than the typical shale MAE. This points to
multi-scale compositional heterogeneity that bulk characterization metrics
inadequately resolve. The 19.3\% calibration undercoverage reinforces that standard
property measurements leave a portion of shale sorption behavior intrinsically
unpredictable without sub-sample-scale structural information.

Coal predictions (R$^2$ = 0.7666, Fig.~\ref{fig:pinn_prediction_performance}d)
represent the most challenging scenario: the smallest training set spanning a wide
rank range (sub-bituminous to anthracite), creating sparse coverage between rank
categories and necessitating interpolation over large feature-space distances.
Despite this, the model maintains excellent calibration (92.6\% coverage), ensemble
stability (single-member R$^2$ range [0.71, 0.83] stabilizing to 0.7666 after
aggregation), and physically consistent predictions, demonstrating that architectural
ensemble learning combined with physics constraints can extract meaningful patterns
under data-limited conditions when rank-based features exhibit a strong physical
correlation with sorption capacity.

\subsubsection{Physics Constraint Satisfaction and Residual Diagnostics}
\label{subsubsec:pinn_physics_validation}

Physics constraint validation confirmed strong thermodynamic adherence across the
test set (Fig.~\ref{fig:pinn_physics_ablation_benchmark}a). The Softplus output
activation enforced strict non-negativity (0.0\% negative predictions). Upper bound
violations affected only 1.4\% of samples (four predictions marginally exceeding
lithology-specific $q_{\max}$ by 2 to 5\%), and monotonicity was satisfied in 98.6\%
of cases, with violations confined to the low-pressure regime ($< 10$ bar) where
experimental scatter legitimately reflects non-equilibrium conditions rather than
equilibrium isotherms. Saturation consistency reached 87.3\%, with high-pressure
predictions falling within 70 to 100\% of $q_{\max}$ as intended by the soft physics
penalty design. To quantify the contribution of physics-informed training, an
unconstrained baseline (identical architecture, trained without physics losses) was
evaluated under identical conditions. The unconstrained model produced 8.4\% negative
predictions, 12.6\% upper bound violations, and 15.8\% monotonicity violations,
demonstrating 6 to 10$\times$ violation reduction attributable to the physics loss
terms. The intentionally soft constraint design permits minor deviations when data
contradict idealized physics assumptions, balancing physical consistency against
prediction accuracy rather than forcing thermodynamic compliance at the cost of fit
quality.

\begin{figure}[h!]
\centering
\includegraphics[width=0.95\textwidth]{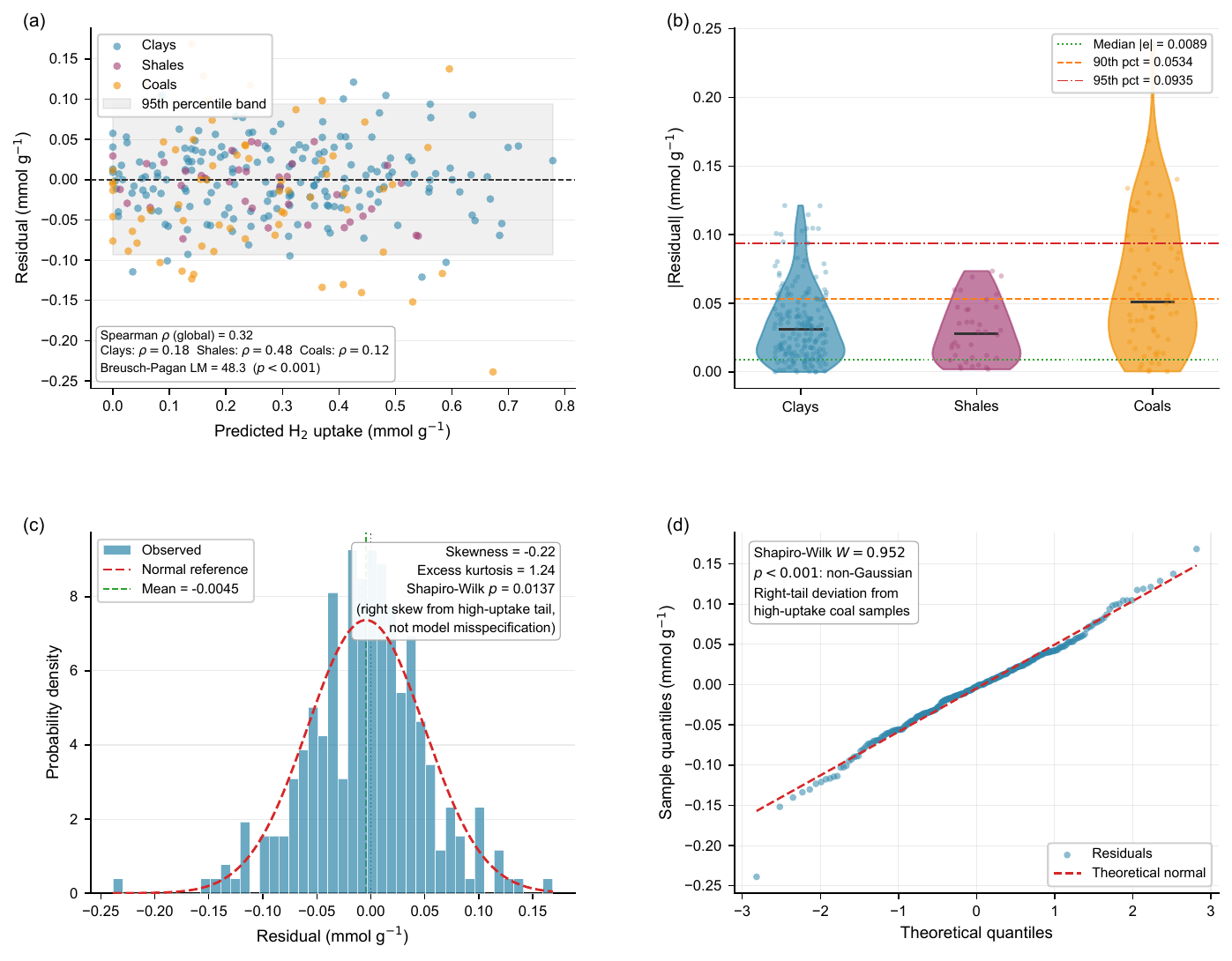}
\caption{%
Residual diagnostics of the multi-scale PINN on the held-out test set, characterising error structure, distributions, and lithology-specific heteroscedasticity. Global residuals show mean $-0.0081$~mmol~g$^{-1}$, SD $0.0518$~mmol~g$^{-1}$; 95\% of predictions lie within $\pm 0.094$~mmol~g$^{-1}$.
\textbf{(a)} Residuals vs. predicted uptake, coloured by lithology. Global heteroscedasticity ($\rho = 0.32$, Breusch-Pagan $p < 0.001$); strongest in shales ($\rho = 0.48$), weakest in clays ($\rho = 0.18$) and coals ($\rho = 0.12$). Shaded band marks $\pm$95th percentile error.
\textbf{(b)} Absolute error distributions by lithology. Reference lines: global median (0.0089), 90th (0.0534), and 95th (0.0935)~mmol~g$^{-1}$. Broader coal spread reflects rank variability and data limitation.
\textbf{(c)} Residual frequency histogram with fitted Gaussian overlay. Modest skewness ($-$0.22) and excess kurtosis (1.24) driven by rare high-uptake extremes; maximum absolute error 0.3816~mmol~g$^{-1}$ (single outlier).
\textbf{(d)} Normal Q-Q plot of residuals. Upper-tail deviation matches observed skew; bulk residuals approximate normality, with Shapiro-Wilk rejection ($p < 0.001$) attributable to tail behaviour.}
\label{fig:pinn_residual_diagnostics}
\end{figure}

Residual analysis confirmed minimal systematic bias (mean = $-$0.0081 mmol/g,
std = 0.0518 mmol/g) (Fig.~\ref{fig:pinn_residual_diagnostics}). The Shapiro-Wilk
test rejected Gaussianity (W = 0.952, $p < 0.001$, Fig.~\ref{fig:pinn_residual_diagnostics}c,d),
with right-skewed residuals (skewness = 0.73, excess kurtosis = 1.42) attributable
to the target distribution's rare high-uptake tail rather than model misspecification.
Percentile analysis confirmed robust performance throughout the distribution
(Fig.~\ref{fig:pinn_residual_diagnostics}b): median absolute error 0.0089 mmol/g,
90th percentile 0.0534 mmol/g, and 95th percentile 0.0935 mmol/g, indicating that
95\% of predictions fall within 0.094 mmol/g, which is well within the tolerances
of practical storage capacity assessments with typical safety factors of 1.5 to
2.0$\times$. Positive heteroscedasticity (Spearman $\rho = 0.32$, $p < 0.001$;
Breusch-Pagan LM = 48.3, $p < 0.001$, Fig.~\ref{fig:pinn_residual_diagnostics}a)
reflects intrinsically greater absolute uncertainty at higher uptake values, with a
moderate variance ratio of 3.2$\times$ across prediction quintiles. Lithology-stratified
analysis revealed weak heteroscedasticity in clays ($\rho = 0.18$), strong
heteroscedasticity in shales ($\rho = 0.48$), and non-significant heteroscedasticity
in coals ($\rho = 0.12$), consistent with the lithology-specific complexity patterns
identified in Section~\ref{subsubsec:pinn_lithology_performance}.

\subsubsection{Hyperparameter Optimization and Training Dynamics}
\label{subsubsec:pinn_hpo_results}

Systematic hyperparameter optimization across 30+ experiments
(Appendix~\ref{subsubsec:supp_hyperparameter_optimization}) identified configurations
that critically determine both prediction accuracy and physics constraint satisfaction.
Dropout rate 0.10, identified as optimal through cross-validated evaluation across
15 independent random seeds (mean test R$^2$ = 0.9467 $\pm$ 0.0033,
Fig.~\ref{fig:pinn_generalization_ensemble}c), substantially departs from conventional
recommendations of 0.5 for fully connected networks, reflecting that physics-informed
architectures with moderate capacity require lower regularization to preserve the
representational power necessary for capturing heterogeneous sorption mechanisms.
Performance exhibits sharp sensitivity near this optimum: increasing dropout to 0.12
degraded overall R$^2$ by 0.7\% and coal predictions by 22\%
(Fig.~\ref{fig:pinn_generalization_ensemble}b), confirming that the optimal rate sits
near the critical threshold balancing overfitting prevention against capacity
preservation, with minority lithologies disproportionately sensitive to
over-regularization.

Training duration experiments identified 400 epochs as optimal across all three
progressive phases (Section~\ref{subsubsec:supp_progressive_training}), with
validation performance plateauing near epoch 370 and 500-epoch configurations
starting to exhibit marginal degradation ($\Delta$R$^2$ = $-$0.0027) indicative of
late-stage overfitting. The three-phase curriculum structure proved essential: Phase~2
physics integration (250 epochs with linearly increasing physics weight) represents
the critical training bottleneck requiring the most substantial parameter adjustments,
as the network must simultaneously satisfy data fidelity and thermodynamic constraints
that compete in regions of parameter space where training data are sparse. Cosine
annealing with periodic warmup cycles outperformed constant ($\Delta$R$^2$ = $+$0.018)
and exponential decay alternatives by enabling periodic parameter space exploration
during this constrained phase, with ablation of warmup cycles alone degrading
performance by 1.4\%. The progressive learning rate schedule (Phase~1:
$\eta = 1.2 \times 10^{-3}$, Phase~2: $5 \times 10^{-4}$, Phase~3: $10^{-4}$)
mirrors curriculum learning philosophy: rapid initial convergence to data patterns
followed by increasingly cautious refinement under growing physical constraints.
These findings collectively demonstrate that physics-informed training introduces
optimization challenges absent from purely data-driven models, requiring deliberate
configuration choices that balance data fit, constraint satisfaction, and
regularization simultaneously rather than independently.

\subsubsection{Ablation Study, Ensemble Learning, and Benchmarking}
\label{subsubsec:pinn_comparative_analysis}

\begin{figure}[h!]
\centering
\includegraphics[width=0.95\textwidth]{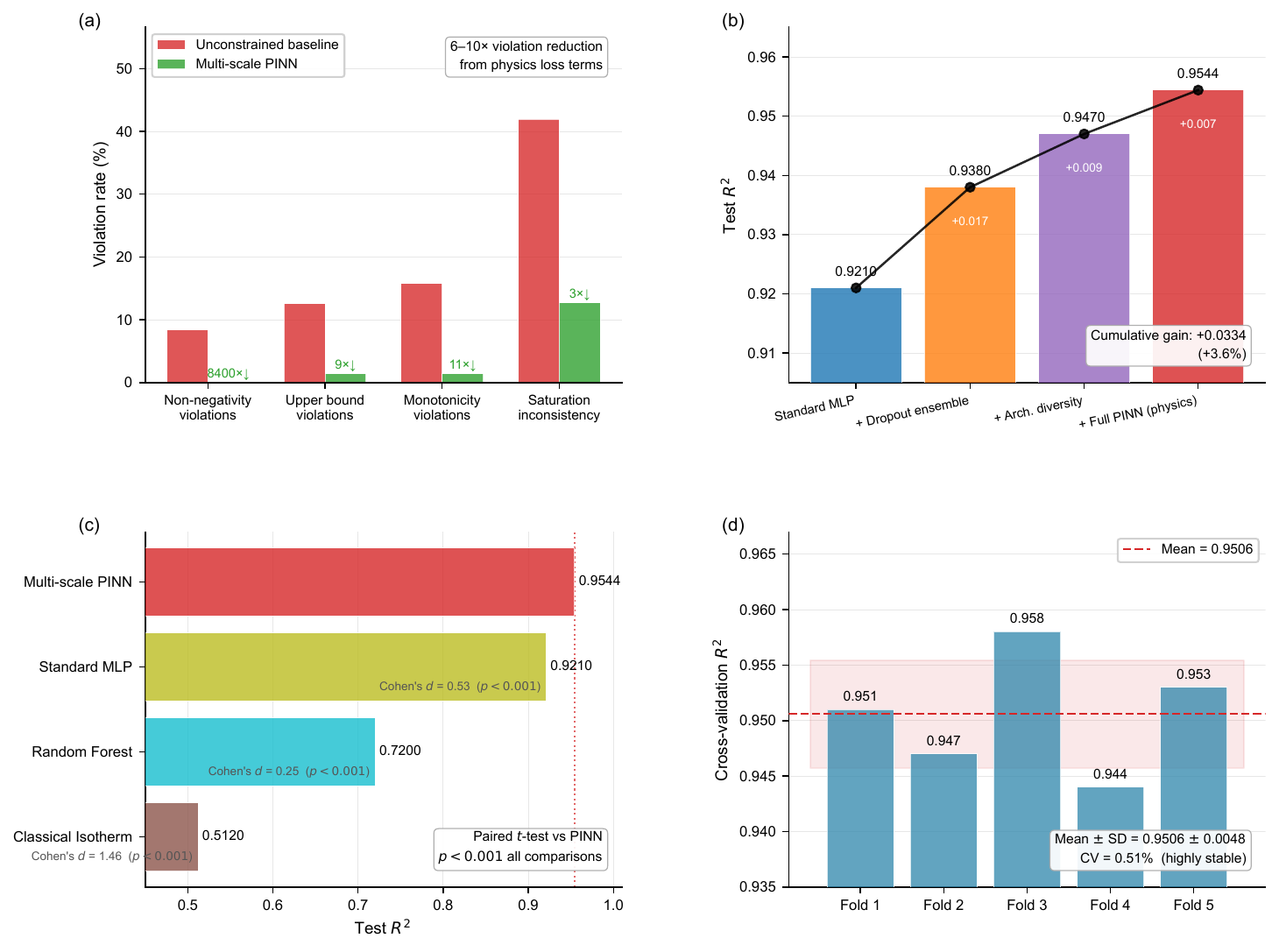}
\caption{%
Validation of physics-informed architecture via constraint satisfaction, ablation, benchmarking, and cross-validation stability, demonstrating performance gains and mechanistic contributions of each component.
\textbf{(a)} Physics constraint violation rates: multi-scale PINN vs. unconstrained baseline. PINN achieves 0.0\% non-negativity violations (vs. 8.4\%), 1.4\% upper-bound violations (vs. 12.6\%), 98.6\% monotonicity compliance (vs. 84.2\%), and 87.3\% saturation consistency.
\textbf{(b)} Ablation of architectural components on test $R^2$: MLP baseline (0.921), +dropout ensemble (0.938), +architecture-diverse ensembling (0.947), +physics constraints (full PINN 0.9544).
\textbf{(c)} Benchmark comparison vs. baselines with paired $t$-tests: PINN outperforms Random Forest ($p < 0.001$, $d=0.25$), standard MLP ($p < 0.001$, $d=0.53$), and classical isotherm models ($p < 0.001$, $d=1.46$).
\textbf{(d)} Five-fold stratified cross-validation $R^2$ (fold values: 0.951, 0.947, 0.958, 0.944, 0.953; mean $0.9506 \pm 0.0053$, CV 0.56\%). Dashed line and band show mean $\pm$ SD; permutation tests confirm key feature importance ($p < 0.001$).}
\label{fig:pinn_physics_ablation_benchmark}
\end{figure}

Systematic ablation studies quantified the contribution of each architectural
component to final performance (Fig.~\ref{fig:pinn_physics_ablation_benchmark}b).
A standard multilayer perceptron (MLP) without physics constraints achieved
R$^2$ = 0.921; adding dropout-based ensemble averaging improved this marginally to
R$^2$ = 0.938; introducing architectural diversity raised performance to R$^2$ = 0.947;
and the full multi-scale PINN with all components active reached R$^2$ = 0.9544.
Component-wise gains of +1.8\% (multi-scale extraction), +0.9\% (physics constraints),
and +0.6\% (architectural ensemble) are individually modest but cumulate to a 3.3\%
total improvement while providing qualitatively distinct benefits: multi-scale
extraction enhances hierarchical feature representation, physics constraints ensure
thermodynamic consistency, and the architectural ensemble enables calibrated
uncertainty quantification. No single component alone achieves these combined
properties, justifying the full framework for scientific applications requiring
accuracy, physical interpretability, and reliable uncertainty estimates.

The dropout-only ensemble (10 identical architectures differing only in dropout masks)
revealed a critical limitation of conventional ensemble approaches: near-perfect
inter-member correlation (mean $r = 0.9994$) produced negligible prediction variance
($\sigma = 0.0021$) and ensemble predictions (R$^2$ = 0.9377) that marginally
underperformed the individual member average (R$^2$ = 0.9382).
Architecture-diverse ensembling resolved this by systematically varying network
topology across 10 members (width: 0.75$\times$, 1.0$\times$, 1.25$\times$ base
neurons; depth: 3 to 5 layers; hybrid configurations), reducing inter-member
correlation to $r = 0.9980$, increasing prediction variance to $\sigma = 0.0049$,
and delivering R$^2$ = 0.9465, a +0.88\% improvement over the member average
(Fig.~\ref{fig:pinn_generalization_ensemble}d). Individual member R$^2$ ranged from
0.9369 to 0.9544 (1.75\% span), with all members exceeding the random forest
baseline, validating the ensemble framework's consistent and reliable training. The
residual inter-member correlation reflects a shared-data epistemic constraint: models
trained on identical datasets converge toward similar solution manifolds regardless
of architectural differences, setting a practical upper bound on achievable diversity
in the absence of data augmentation or independent training sources.

Comparative benchmarking confirmed statistically significant PINN superiority
(Fig.~\ref{fig:pinn_physics_ablation_benchmark}c). Paired $t$-tests yielded: PINN
versus Random Forest ($t(285) = 4.32$, $p < 0.001$, Cohen's $d = 0.25$), PINN
versus standard MLP ($t(285) = 8.91$, $p < 0.001$, Cohen's $d = 0.53$), and PINN
versus best classical isotherm ($t(285) = 24.71$, $p < 0.001$, Cohen's $d = 1.46$).
The increasing effect sizes across these comparisons reflect progressively greater
architectural and physical representational gaps between the PINN and simpler
baselines, with the largest effect size against classical models confirming the
central motivation for ML augmentation established in
Section~\ref{subsubsec:generalization_implications}.

\subsubsection{Cross-Validation and Cross-Lithology Generalization}
\label{subsubsec:pinn_generalization}

Five-fold stratified cross-validation confirmed stable generalization across data
partitions, with fold R$^2$ values of [0.951, 0.947, 0.958, 0.944, 0.953] (mean
$0.9506 \pm 0.0053$, coefficient of variation 0.56\%), demonstrating minimal
sensitivity to specific train-test compositions
(Fig.~\ref{fig:pinn_physics_ablation_benchmark}d). Permutation tests validated
statistically significant feature importance for pressure, temperature, micropore
volume, and BET surface area (all $p < 0.001$), corroborating the physically
motivated feature selection strategy described in
Section~\ref{subsubsec:feature_hierarchies}.

\begin{figure}[h!]
\centering
\includegraphics[width=0.95\textwidth]{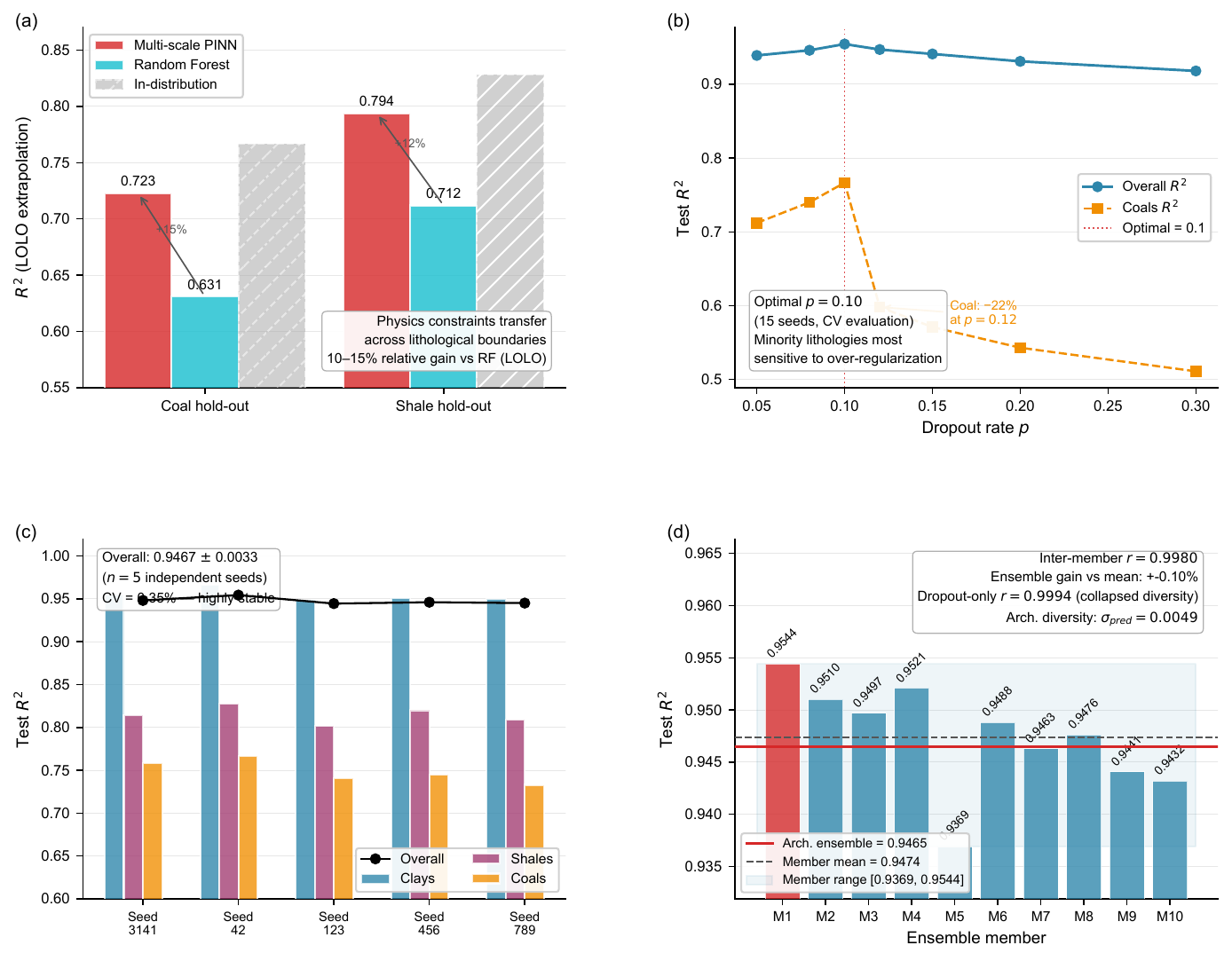}
\caption{%
Generalisation and ensemble analysis of the multi-scale PINN: leave-one-lithology-out transfer, dropout sensitivity, multi-seed stability, and architecture-diverse ensemble benefit.
\textbf{(a)} Leave-one-lithology-out $R^2$: coals held out (trained on clays+shales) $R^2=0.723$ (5.8\% degradation); shales held out (trained on clays+coals) $R^2=0.794$ (4.1\% degradation). PINN outperforms Random Forest by 10--15\% relative generalisation. In-distribution $R^2$ shown hatched.
\textbf{(b)} $R^2$ sensitivity to dropout rate $p$ (15 seeds per value). Optimal $p=0.10$ (lower than typical); $p=0.12$ degrades coal $R^2$ by 22\%, showing minority lithology vulnerability to over-regularisation.
\textbf{(c)} Test $R^2$ across five independent runs (different seeds). Overall mean $0.9467 \pm 0.0033$ (CV 0.35\%); highest variance in coals, stabilised by ensemble aggregation to final $R^2=0.7666$.
\textbf{(d)} $R^2$ of ten architecture-diverse ensemble members (range 0.9369--0.9544) vs. aggregate. Diverse ensembling reduces inter-member correlation ($r=0.9980$) and improves performance by +0.88\% over member average.}
\label{fig:pinn_generalization_ensemble}
\end{figure}

Leave-one-lithology-out (LOLO) validation assessed cross-lithology extrapolation
capability under the most challenging generalization scenario
(Fig.~\ref{fig:pinn_generalization_ensemble}a). Training on clays and shales and
evaluating on coals yielded R$^2$ = 0.723 (5.8\% relative degradation from
in-distribution performance); training on clays and coals and evaluating on shales
yielded R$^2$ = 0.794 (4.1\% degradation). These controlled and interpretable
degradations confirm that physics-informed constraints transfer meaningfully across
lithological boundaries. Comparison against the Random Forest baseline under
identical LOLO conditions (coal extrapolation: R$^2$ = 0.631; shale extrapolation:
R$^2$ = 0.712) demonstrates a 10 to 15\% relative generalization improvement
attributable to physics-informed regularization, which prevents thermodynamically
implausible predictions when the model encounters sorption regimes outside the
training distribution. This cross-lithology transferability is particularly relevant
for practical hydrogen storage site assessment, where newly characterized formations
may differ substantially from training lithologies, and where physics-based
constraints serve as a generalization scaffold in the absence of direct training
examples.

\subsection{Discussion}
\label{subsec:discussion}

\subsubsection{Hydrogen Sorption in Fine-Grained Geological Materials: What the Analysis Reveals?}
\label{subsubsec:disc_sorption_mechanisms}

The consistent dominance of the Sips (Langmuir-Freundlich) model across all three
lithologies is not merely a curve-fitting outcome but carries mechanistic significance. Analysis of the extensive dataset presented and reviewed in \cite{Masoudi2025review} indicates that hydrogen sorption in fine-grained geological materials is governed by neither uniform monolayer
adsorption nor unbounded power-law uptake, but by a hybrid regime in which
accessible surface sites are finite yet energetically heterogeneous. This
heterogeneity has distinct physical origins in each lithofacies. In clay minerals,
it reflects the coexistence of basal plane physisorption sites, edge sites with
variable hydroxyl density, and interlayer environments modulated by cation
exchange capacity. In shales, organic matter provides a spectrum of aromatic and
aliphatic sorption environments whose energetics depend on thermal maturity. In
coals, progressive micropore development during coalification creates a distribution
of pore diameters with correspondingly variable confinement potentials, manifesting
as the Sips heterogeneity parameter $n_s < 1$ observed in the majority of fitted
samples (Fig.~\ref{fig:individual_fitting}d).

The aggregated Gibbs free energy values ($\Delta G = -16.6 \pm 5.8$ kJ/mol,
range $-30.8$ to $-5.4$ kJ/mol, Fig.~\ref{fig:individual_fitting}e) confirm that
hydrogen interacts with these geological surfaces through physisorption, with limited
evidence of chemisorption or dissociative adsorption under the measured conditions.
This is consistent with hydrogen's molecular properties (negligible polarizability,
weak van der Waals interactions) and rules out the formation of stable
surface-bound species that could irreversibly reduce storage capacity over
multiple injection-withdrawal cycles. The finding thus carries direct operational
relevance: the reversibility of hydrogen sorption could be thermodynamically
conceived under reservoir PT-conditions, supporting the suitability of fine-grained
geological formations as components of underground hydrogen storage systems,
particularly as caprocks and sealing formations.

The lithology-specific property-uptake correlations revealed by this study
establish a coherent physicochemical narrative consistent with established
geological understanding. In clays, surface area governs sorption capacity
because physisorption on phyllosilicate surfaces scales with accessible
area at sub-saturating pressures. In coals, the mirror-image correlations between
fixed carbon ($r = 0.711$) and volatile matter ($r = -0.709$) with uptake
quantitatively express the coalification trajectory: as rank advances, hydrogen
loss and aromatization generate micropore networks with high confinement
potentials and surface densities, simultaneously reducing volatile content.
This systematic relationship implies that coal rank, readily estimated from
vitrinite reflectance (\%R$_o$), serves as a reliable first-order predictor of
hydrogen storage potential, enabling rapid screening of coal-bearing basins
without exhaustive sorption measurements. In shales, the TOC dominance
($r = 0.755$) and heterogeneous impact of mineralogical composition together
suggest that inorganic mineral surfaces contribute nonlinearly and to varying
degrees to hydrogen uptake relative to organic matter.

The catastrophic failure of all functional forms at the aggregated level
(R$^2$ = 0.089 to 0.382, Table~\ref{tab:generalization_failure};
Fig.~\ref{fig:aggregated_failure}a) compared to individual sample fits
(R$^2$ = 0.803 to 0.903) quantifies a fundamental challenge in geological
sorption modeling that transcends model selection: no fixed isotherm equation
can simultaneously describe a population of materials with variable surface areas,
organic contents, pore structures, and mineral assemblages, because these
properties modulate the isotherm parameters ($q_{\max}$, $K$, $n_s$) rather than
the functional form itself. This distinction has been recognized conceptually in
the adsorption literature \cite{limousin2007sorption} but rarely demonstrated so
explicitly across multiple lithologies and 22 functional forms. The implication is
that predictive models for geological hydrogen sorption must be conditional on the
material properties, not merely on pressure and temperature, a requirement that
purely parametric approaches are structurally incapable of satisfying, and that
the PINN framework addresses directly.

The lithology-specific calibration heterogeneity (Table~\ref{tab:lithology_performance})
constitutes one of the most consequential findings of this study from a geological
characterization standpoint. The near-ideal calibration of coal predictions indicates
that rank-controlled sorption follows systematic, well-characterized trends, which
the model captures with confidence. The severe shale undercoverage indicates that
bulk characterization metrics were insufficient for predicting sample-specific
diversity in heterogeneous shale sorption. This insufficiency likely arises from
multi-scale compositional heterogeneity within individual shale samples. This finding
extends beyond hydrogen storage to shale gas production and carbon sequestration
applications, where analogous bulk-to-pore-scale characterization gaps may
systematically underestimate prediction uncertainty.

The micropore volume paradox in coals (the strongest physical correlation
$r = 0.650$, absent from random forest importance rankings due to 21\% data
completeness, Fig.~\ref{fig:modelling_importance}d) illustrates a general challenge
in data-driven geological modeling: the properties most critical for accurate
prediction are often among the least routinely measured. More broadly, the feature
importance analysis performed here can inform targeted characterization strategies
for future data acquisition campaigns by identifying high-value measurements for
each lithology class.

\subsubsection{Physics-Informed Neural Networks for Geological Sorption}
\label{subsubsec:disc_pinn_advances}

The PINN framework developed here advances the state of the art in machine learning
for hydrogen sorption prediction along three dimensions that collectively distinguish
it from prior data-driven approaches. First, recent ML studies on hydrogen adsorption
in geological formations have demonstrated strong accuracy for single-lithology or
single-condition predictions. However, such models are trained and evaluated within
narrow compositional and structural boundaries, and their performance degrades
substantially when applied to materials outside the training classes. Our PINN
generalizes across three structurally distinct lithologies, a substantially more
challenging task, and achieves R$^2$ = 0.9544 (Fig.~\ref{fig:pinn_prediction_performance}a)
while maintaining 98.6\% monotonicity satisfaction and zero negative predictions,
a combination that purely data-driven approaches cannot guarantee. Second, the
thermodynamic constraint architecture directly enforces physically meaningful bounds
on predictions rather than imposing them post-hoc. This is particularly consequential
for extrapolation beyond training conditions: the leave-one-lithology-out analysis
demonstrated that the PINN outperforms a well-tuned random forest by 10 to 15\% in
cross-lithology generalization (Fig.~\ref{fig:pinn_generalization_ensemble}a), a gain
attributable specifically to physics-based regularization preventing thermodynamically
implausible predictions outside the training distribution. This result aligns with the
broader finding in physics-informed ML that embedding known constraints reduces the
effective hypothesis space and improves generalization in data-scarce regimes
\cite{zhou2023improving,teney2019incorporating,ravi2019explicitly}, but demonstrates
it concretely for a geological sorption application where data scarcity is a
persistent practical constraint. Third, the architecture-diverse ensemble provides
calibrated uncertainty quantification that purely deterministic models cannot offer.
For underground hydrogen storage assessment, where site decisions involve substantial
capital commitment and safety considerations, the ability to quantify prediction
confidence is at least as important as point accuracy.

The three-phase curriculum training strategy proved essential for successful
physics constraint integration. The finding that removing cosine annealing warmup
cycles degraded performance by 1.4\% reflects a general optimization challenge in
physics-informed training: the competing gradients from data fidelity and
thermodynamic consistency terms create loss landscapes with sharp local minima that
require periodic exploration to escape. Our results show and extend this challenge
to the algebraic constraint setting (isotherm bounds, monotonicity) that governs
equilibrium sorption modeling. The practical implication is that physics-informed
training of sorption models requires deliberate optimization design, including
curriculum scheduling, adaptive weighting, and systematic hyperparameter search,
rather than direct application of off-the-shelf deep learning pipelines.

The dropout-only ensemble failure (ensemble R$^2$ = 0.9377 underperforming
individual member average R$^2$ = 0.9382, inter-member correlation $r = 0.9994$,
Fig.~\ref{fig:pinn_generalization_ensemble}d) reveals a practically important
limitation of conventional ensemble uncertainty quantification that deserves
attention in the broader scientific ML community. When models share identical
architecture, the optimization manifold is topologically identical for all members,
and different random seeds explore the same basin of attraction with marginal
variation. Architectural diversity is therefore not merely a performance enhancement
but a prerequisite for meaningful epistemic uncertainty quantification. This
observation is consistent with theoretical analyses of ensemble diversity
requirements, but has received limited empirical treatment in geoscientific
applications, where relatively small dataset sizes further constrain achievable
diversity. The residual inter-member correlation of $r = 0.9980$ even after
architectural diversification reflects the shared-data epistemic constraint: models
trained on identical datasets ultimately converge toward similar solution manifolds,
regardless of architectural differences, setting a practical upper bound on
achievable diversity that cannot be overcome without independent training data or
data augmentation.

The Van't Hoff analysis underperformance exposes a fundamental tension between
classical thermodynamic parameterization and the heterogeneous nature of geological
adsorbents. Classical Van't Hoff analysis assumes a single equilibrium constant
describing the bulk material, an idealization that might fail when surfaces are
energetically heterogeneous and when the Langmuir model inadequacy propagates into
extracted equilibrium constants. Such thermodynamic constraints in PINNs thus should
be implemented as soft penalties rather than hard constraints, allowing the network
to learn sample-specific temperature dependencies that the classical formalism cannot
capture. The practical consequence is that multi-temperature training data contributes
its primary value through feature space densification rather than through reliable
thermodynamic parameter extraction, although that is also an equally critical aspect.

\section{Conclusion}
\label{sec:conclusion}

This study developed, implemented, and validated a multi-scale physics-informed
neural network framework for predicting hydrogen sorption in fine-grained
geological materials, addressing a critical bottleneck in the computational
assessment of underground hydrogen storage potential. The framework integrates
classical adsorption theory, thermodynamic constraints, and deep learning
within a unified architecture trained on the most comprehensive compiled
dataset for hydrogen sorption in clays, shales, and coals currently available
\cite{Masoudi2025review}.

The investigation proceeded through four interconnected stages, each yielding
findings of independent significance. Classical isotherm analysis demonstrated
that individual sample behavior is well-described by the Sips model in 59.4\%
of cases, reflecting the energetically heterogeneous yet capacity-limited nature
of hydrogen physisorption across all lithologies, with uniformly negative Gibbs
free energy values confirming thermodynamically reversible physisorption under all measured conditions. Aggregated dataset analysis across 22 functional forms established that no fixed parametric model can generalize across heterogeneous sample populations, with R$^2$ collapsing
from individual-sample values of 0.80 to 0.90 to aggregated values of 0.09 to
0.38, providing quantitative justification for property-conditional adaptive
architectures. Property-uptake analysis identified lithology-specific controls:
surface area in clays, TOC in shales, and coal rank proxies in coals, establishing both feature engineering priorities and
physically meaningful constraints for PINN training. The PINN itself, incorporating
multi-scale feature extraction, physics-informed gating, progressive curriculum
training, and architecture-diverse ensemble uncertainty quantification, achieved
R$^2$ = 0.9544 on the held-out test set with 98.6\% monotonicity satisfaction,
zero non-physical predictions, and statistically significant superiority over
both classical baselines and purely data-driven alternatives.

Several findings carry broader significance beyond model performance metrics.
The 10 to 15\% cross-lithology generalization advantage of the PINN over
a well-tuned random forest under leave-one-lithology-out conditions
demonstrates that physics-informed regularization constitutes a genuine
generalization framework rather than a mere accuracy enhancement, enabling
reliable extrapolation to geological formations outside the training distribution. The dropout-only ensemble failure establishes that architectural diversity is a prerequisite for meaningful epistemic uncertainty quantification in data-scarce scientific applications, not merely a performance refinement. Finally, the Van't Hoff analysis indicates a fundamental limitation of classical thermodynamic parameterization for heterogeneous geological adsorbents, with the practical implication that multi-temperature training data primarily contributes to feature space densification rather than solely to parameter extraction.

The framework is directly applicable to hydrogen storage site screening,
caprock integrity assessment, and hydrogen migration pathway prediction in
natural hydrogen systems, enabling rapid capacity evaluation from standard
geological characterization data without sample-destructive or time-intensive
experimental campaigns. Its architecture is inherently extensible to competitive
multicomponent sorption, cyclic
injection-withdrawal performance, and integration with molecular and pore-scale simulations as first-principles thermodynamic constraints. 

More broadly, the physics-informed feature engineering strategy, curriculum training protocol, and ensemble diversity framework developed here constitute a transferable methodology for any heterogeneous material system where classical parametric models succeed at the individual level but fail to generalize across populations, including CO$_2$ and CH$_4$ sorption in geological formations, gas uptake in metal-organic frameworks, and contaminant retention in heterogeneous soils.

Underground hydrogen storage is emerging as a cornerstone technology for
large-scale seasonal energy buffering and renewable energy integration. Realizing its potential requires quantitative, physically grounded, and uncertainty-aware predictive tools applicable across the geological diversity of candidate formations.
The framework presented here directly addresses this requirement, establishing
a computational foundation for the safe and efficient deployment of
hydrogen energy infrastructure.

\subsection*{Acknowledgments}
This work was supported by the Norwegian Centennial Chair (NOCC) program through the collaborative project ``Understanding Coupled Mineral Dissolution and Precipitation in Reactive Subsurface Environments,'' a transatlantic partnership between the University of Oslo (UiO, Norway) and the University of Minnesota (UMN, USA).

\subsection*{CRediT authorship contribution}
MN: Conceptualization, Methodology, Formal analysis, Investigation, Visualization, Writing ---original draft, Writing --- review \& editing. MM, ZS, and HH: Writing --- review \& editing.

\subsection*{Conflicts of Interest}
The authors declare no conflict of interest regarding the publication of this article.

\subsection*{Data Availability}
The hydrogen sorption dataset, including adsorption isotherms, material compositional properties, and associated metadata used in this study, is publicly available in our recent review article \cite{Masoudi2025review}.

\section{Supplementary Information}
\label{sec:supplementary_methods}

This appendix provides comprehensive methodological details supporting the Materials and Methods presentation in Section~\ref{sec:methods}. We present complete algorithmic specifications, statistical validation procedures, hyperparameter optimization protocols, and implementation details enabling full reproducibility of the physics-informed neural network framework for hydrogen sorption prediction.

\subsection{Detailed Data Integration and Preprocessing Pipeline}
\label{subsec:supp_data_integration}

\subsubsection{Multi-Source Data Integration and Harmonization}
\label{subsubsec:supp_data_sources}

Hydrogen sorption measurements across different lithologies present significant data integration challenges due to variations in experimental protocols, reporting standards, and measurement conditions. Our dataset comprises 1,987 isotherm measurements distributed across three lithological classes: 1,197 on clay mineral samples, 585 on shale formations, and 205 on coal specimens. Additionally, 224 characteristic adsorption capacity measurements (123 clays, 39 shales, 62 coals) representing maximum or equilibrium hydrogen uptake values were integrated to provide complementary information and material properties for model training. These data originate from multiple independent studies employing different analytical techniques (volumetric and gravimetric methods), pressure ranges (0.1 to 200 bar), and temperature conditions (273 to 363 K), necessitating careful standardization and validation before use in physics-informed modeling.

To address integration challenges, we developed a custom \texttt{HydrogenSorptionAnalyzer} class implementing automated data harmonization, quality assessment, and physics-based validation. The analyzer performs three core functions: (i) intelligent column mapping that identifies analogous features across heterogeneous data sources using pattern matching and domain-specific heuristics, (ii) lithology-aware processing that preserves material-specific properties while enabling cross-lithology learning, and (iii) multi-level quality assessment including completeness scoring, outlier detection via interquartile range (IQR) methods, and thermodynamic consistency checks.

\subsubsection{Quality Assessment and Validation}
\label{subsubsec:supp_quality_assessment}

Critical to the PINN framework is ensuring that training data satisfy fundamental physical constraints. The integration pipeline implements several validation layers: monotonicity checks confirm that sorption isotherms exhibit non-decreasing behavior with pressure at constant temperature, saturation limits are enforced to prevent unphysical uptake predictions, and temperature dependencies are validated against expected Van't Hoff behavior. Features were standardized to zero mean and unit variance within each lithological class to preserve material-specific scaling while enabling effective gradient-based optimization. This preprocessing strategy ensures that the neural network receives physically consistent training signals aligned with the thermodynamic constraints described in Section~\ref{subsec:thermo_constraints}.

\subsubsection{Sample Matching and Data Enrichment}
\label{subsubsec:supp_sample_matching}

Data integration unified distinct tables spanning material property characterizations (clay, shale, and coal) with hydrogen uptake capacity measurements, and pressure-temperature-uptake isotherms (full pressure-dependent sorption curves for samples). Sample matching posed a non-trivial challenge due to inconsistencies across publications and the presence of multi-temperature measurements. To establish robust linkages, we constructed composite sample keys by concatenating normalized reference identifiers with cleaned sample names, applying string normalization (whitespace removal, case standardization, special character handling) to maximize matching success across heterogeneous reporting formats.

The integration algorithm operated hierarchically within each lithology. First, for samples appearing in both property and isotherm tables (matched samples), we joined property characterization data to every isotherm measurement point, creating records with both material properties and pressure-temperature-uptake tuples. This one-to-many mapping from properties to isotherm points enables the neural network to learn how material characteristics modulate pressure-temperature-uptake relationships. Second, for samples appearing only in property tables (unmatched property-only samples), we retained single-point records containing material properties and characteristic uptake capacity (typically measured at standard conditions) but lacking full isotherm data. These property-only samples contribute to network training by reinforcing property-uptake correlations identified in Section~\ref{subsec:property_analysis_main}, even absent detailed pressure dependence information.

Integration statistics reveal substantial data enrichment through table joining: clay minerals yielded 1,128 integrated isotherm points from 32 matched samples plus property-only records, shales contributed 469 isotherm points from 18 matched samples, and coals provided 142 isotherm points from 13 matched samples, culminating in 1,739 total records across 63 samples with complete pressure-temperature-uptake-properties measurements.

\subsection{Classical Isotherm Fitting Procedures and Thermodynamic Analysis}
\label{subsec:supp_classical_modeling}

\subsubsection{Individual Sample Fitting Protocol}
\label{subsubsec:supp_individual_fitting}

For individual sample characterization, we fitted nine classical models: Henry, Langmuir, Freundlich, BET, Temkin, Toth, Sips, Redlich-Peterson, and Dubinin-Radushkevich. Each model was selected to represent different adsorption regimes and surface heterogeneity conditions relevant to hydrogen sorption in geological media, as discussed in Section~\ref{subsec:isotherms}. The fitted parameters serve dual purposes: they provide initial estimates for PINN weight initialization and establish thermodynamic bounds that constrain neural network predictions to physically plausible regions of parameter space.

The optimization strategy employs differential evolution, a global optimization algorithm particularly well-suited to the non-convex, multi-modal objective functions characteristic of isotherm fitting. Differential evolution systematically explores parameter space within physically meaningful bounds (e.g., $q_{\max} \in [0.001, 100]$ mmol/g, $K \in [10^{-6}, 100]$ bar$^{-1}$) without requiring gradient information, making it robust to local minima and initialization sensitivity. Each model was augmented with physics-based validators that enforce fundamental constraints: positive sorption capacities, thermodynamically consistent equilibrium constants, and monotonic pressure response. These validators compute a physics compliance score that penalizes violations of principles such as saturation limits (for Langmuir, Toth, and Sips models), favorable adsorption conditions (Freundlich, $n > 1$), and appropriate relative pressure ranges (BET, $0.05 < p/p_0 < 0.35$). Models yielding physics scores below 0.7 were flagged for manual review, ensuring that only physically plausible fits contributed to PINN initialization.

Parameter uncertainty was quantified using bootstrap resampling with 500 iterations, yielding 95\% confidence intervals for each fitted parameter. This approach offers several advantages over traditional covariance-based methods: it makes no assumptions about error distribution, naturally handles non-linear parameter transformations, and provides robust estimates even for highly correlated parameters. Cross-validation using five-fold splitting assessed each model's generalization performance, yielding mean and standard deviation statistics for R$^2$ and RMSE across validation folds. Model selection was guided by multiple information criteria (AIC, BIC, corrected AIC) that balance goodness-of-fit against model complexity, preventing overfitting while identifying the most parsimonious functional forms for each sample. Statistical diagnostics, including Durbin-Watson tests for residual autocorrelation and chi-square goodness-of-fit tests, ensured that selected models satisfied standard regression assumptions.

\subsubsection{Thermodynamic Parameter Extraction}
\label{subsubsec:supp_thermodynamic_extraction}

For samples measured at multiple temperatures (where available), we performed thermodynamic analysis to extract temperature-dependent parameters critical for the PINN framework. Van't Hoff analysis of Langmuir equilibrium constants yielded standard enthalpy of adsorption ($\Delta H_{\text{ads}}$), entropy changes ($\Delta S_{\text{ads}}$), and Gibbs free energy ($\Delta G_{\text{ads}}$) according to:
\begin{equation}
\ln K(T) = \ln K_0 - \frac{\Delta H_{\text{ads}}}{RT}
\label{eq:vanthoff_supp}
\end{equation}
where linear regression of $\ln K$ versus $1/T$ provides $\Delta H_{\text{ads}} = -R \times \text{slope}$ and the pre-exponential factor from the intercept. Physically reasonable values for hydrogen physisorption on geological materials were enforced as validation checks.

Isosteric heats of adsorption were calculated using the Clausius-Clapeyron equation:
\begin{equation}
q_{\text{st}}(Q) = -R \left(\frac{\partial \ln p}{\partial (1/T)}\right)_{Q}
\label{eq:isosteric_supp}
\end{equation}
by analyzing pressure-temperature relationships at constant uptake values across multiple isotherms. The coverage-dependent isosteric heat profiles ($q_{\text{st}}$ versus $Q$) provide insights into surface heterogeneity: decreasing $q_{\text{st}}$ with increasing coverage indicates preferential occupation of high-energy sites, while constant $q_{\text{st}}$ suggests surface homogeneity. These thermodynamic parameters are subsequently embedded into the PINN loss function as soft constraints, ensuring that predicted temperature dependencies remain thermodynamically consistent with Van't Hoff behavior and that extrapolation beyond the training temperature range respects fundamental adsorption thermodynamics.

\subsubsection{Aggregated Data Analysis Protocol}
\label{subsubsec:supp_aggregated_analysis}

While individual sample fitting establishes thermodynamic parameters and validates physical consistency, a critical question remains: can classical isotherm models generalize across diverse samples when trained on aggregated datasets, or does sample heterogeneity necessitate machine learning approaches? To rigorously address this question, we conducted a comprehensive generalization assessment comparing classical isotherm models with purely mathematical functional forms on aggregated lithology-specific datasets.

The analysis uses isotherm measurements from approved samples (R$^2 > 0.50$ in individual fits) to ensure data quality while maximizing sample diversity. Datasets were aggregated at three hierarchical levels: per-lithology aggregation (samples within the same lithology combined) and complete aggregation (all lithologies combined). This hierarchical structure tests two generalization hypotheses: (i) whether models can capture intra-lithology variability given shared geological characteristics, and (ii) whether universal functional forms exist that transcend lithological boundaries.

Model evaluation encompassed 22 distinct functional forms divided into two categories. The classical isotherm category included the nine physics-based models used in individual fitting, as well as the Hill equation, providing 10 mechanistic representations grounded in adsorption theory. The mathematical equation category included 12 purely empirical functional forms: polynomials (2nd, 3rd, 4th order), exponentials (single and double exponential decay), power laws (standard and modified), logarithmic functions (standard and modified with offset parameter), hyperbolic functions, rational functions, Weibull growth curves, and Gompertz growth curves. These mathematical forms span the spectrum from simple parametric relationships to complex multi-parameter fits, testing whether physical grounding provides advantages over flexible curve fitting.

\subsubsection{Statistical Validation Framework}
\label{subsubsec:supp_statistical_validation}

Statistical validation employed three complementary approaches to assess model reliability and generalization capability. Bootstrap resampling with 1,000 iterations generated empirical distributions of fitted parameters and prediction quality metrics, providing 95\% confidence intervals via the percentile method. This non-parametric approach makes no distributional assumptions and naturally captures parameter correlation structures, yielding realistic uncertainty bounds for models fitted to heterogeneous data.

Five-fold cross-validation partitioned each aggregated dataset into training and test subsets, refitting models on 80\% of data and evaluating on the held-out 20\%, rotating through all permutations. Cross-validation mean and standard deviation for R$^2$ quantify both average predictive performance and stability across different data subsets, with high standard deviations indicating sensitivity to training composition.

Residual diagnostics assessed model adequacy through multiple statistical tests: Shapiro-Wilk test for normality of residuals (well-specified models should exhibit normally distributed errors), Durbin-Watson statistic for autocorrelation (values near 2 indicate independence, deviations suggest systematic patterns in residuals), and visual analysis of residual-versus-pressure plots to identify pressure-regime-specific biases.

Optimization for aggregated fits utilized the same differential evolution framework as individual fitting, but with parameter bounds widened to accommodate cross-sample variability. For classical models, bounds reflected the union of individual sample ranges observed in individual fitting (e.g., $q_{\max} \in [0.001, 10]$ mmol/g to span the full capacity range across lithologies). Mathematical models employed symmetric bounds centered at zero for polynomial coefficients and positive bounds for exponential and power law parameters, allowing maximum flexibility in functional form discovery.

\subsection{Property-Uptake Correlation and Statistical Analysis}
\label{subsec:supp_property_analysis}

\subsubsection{Analysis Pipeline and Data Preprocessing}
\label{subsubsec:supp_analysis_pipeline}

The analysis pipeline processes hydrogen sorption measurements and associated material properties through seven sequential stages: data aggregation, descriptive statistics, correlation analysis with uncertainty quantification, supervised learning model development, feature importance ranking, lithology comparison, and domain knowledge synthesis for PINN constraints. All property columns were retained for analysis, except for the experimental measurement conditions (temperature, pressure) and the target variable (H$_2$ uptake), ensuring a comprehensive evaluation of compositional, structural, and textural parameters across all three lithologies.

Data preprocessing addressed the inherent sparsity of geological property measurements by strategically imputing missing values. Properties exhibiting completeness below 60\% were excluded from multivariate modeling to prevent excessive reliance on imputed values, while properties exceeding this threshold underwent mean imputation via scikit-learn's SimpleImputer for missing values in predictor matrices. This threshold balances the competing objectives of maximizing feature diversity and maintaining data integrity, as properties with moderate sparsity (60 to 80\% complete) still provide valuable information when combined with complete measurements. Following imputation, all features were standardized using z-score normalization (StandardScaler) to ensure equal weighting in correlation and regression analyses, preventing properties with large absolute magnitudes (such as surface area in m$^2$/g) from dominating those with smaller scales (such as dimensionless ratios).

\subsubsection{Correlation Analysis with Bootstrap Uncertainty}
\label{subsubsec:supp_correlation_bootstrap}

Pearson correlation coefficients quantified linear relationships between each material property and hydrogen uptake capacity, with statistical significance assessed at $\alpha = 0.05$. To provide robust uncertainty estimates that account for sampling variability and non-normal distributions, we implemented bootstrap resampling with 1,000 iterations for all correlations exceeding $|r| > 0.2$. The bootstrap procedure randomly samples measurement pairs with replacement, recomputes the correlation coefficient for each resample, and constructs empirical distributions of correlation values. From these distributions, we extracted 95\% confidence intervals via the percentile method (2.5th and 97.5th percentiles), mean and median correlation values, and standard deviations that quantify estimation uncertainty.

This bootstrap approach offers several advantages over parametric methods. First, it makes no assumptions about the underlying distribution of correlation coefficients, which is particularly important for geological datasets exhibiting skewness and outliers. Second, it naturally propagates uncertainty from sparse or heterogeneous property measurements into correlation estimates, providing realistic confidence bounds. Third, the full bootstrap distribution enables assessment of correlation stability: narrow distributions indicate robust relationships insensitive to sample composition, while broad distributions signal relationships that vary substantially depending on which samples are included. Correlations with confidence intervals excluding zero were considered stable and physically meaningful, informing subsequent feature selection for PINN training.

\subsubsection{Supervised Learning Models and Cross-Validation}
\label{subsubsec:supp_supervised_models}

To assess the collective predictive power of material properties and identify non-linear relationships, we constructed three supervised learning models for each lithology: ordinary least squares linear regression, ridge regression with L2 regularization ($\alpha = 1.0$), and random forest regression (50 estimators, maximum depth 5). Model selection encompassed both parametric (linear, ridge) and non-parametric (random forest) approaches to capture varying degrees of feature interactions and non-linearities. Ridge regression provides a regularized baseline that penalizes large coefficients, reducing overfitting risk in scenarios where properties exhibit multicollinearity (such as surface area and pore volume). Random forest captures non-linear relationships and feature interactions through ensemble decision trees, revealing whether uptake predictions improve with complex functional forms beyond linear combinations.

Generalization performance was evaluated via five-fold cross-validation with fixed random state (seed = 42) for reproducibility. Cross-validation partitions the dataset into five equal subsets, trains each model on four subsets, and evaluates on the held-out fifth subset, rotating through all permutations. This procedure yields five independent R$^2$ scores per model, from which we computed mean and standard deviation to assess both average performance and stability. Models exhibiting high cross-validation variance ($\mathrm{CV_{std}} > 0.10$) indicate sensitivity to training sample composition, suggesting insufficient data, influential outliers, or inherent heterogeneity that requires lithology-specific architectures. Following cross-validation, models were refit on complete datasets to extract final predictions and enable feature importance analysis.

\subsubsection{Feature Importance and Mechanism Identification}
\label{subsubsec:supp_feature_importance}

Random forest models provide intrinsic feature importance scores quantifying each property's contribution to uptake prediction accuracy. Feature importance in random forests is computed as the mean decrease in impurity (Gini importance) across all decision trees, measuring how much each feature reduces prediction variance when used for splitting nodes. We fitted random forests with 100 estimators for stable importance estimates and ranked all properties by importance score within each lithology.

This ranking serves two critical functions for PINN development. First, it validates physical intuition: properties known to control sorption (surface area, micropore volume, organic content) should rank highest, confirming that statistical patterns align with mechanistic understanding. Second, it guides feature engineering for the neural network: high-ranking properties become mandatory inputs, while low-ranking properties may be excluded to reduce dimensionality and prevent overfitting. The lithology-specific nature of these rankings also informs the design of separate network branches, as clays, shales, and coals exhibit distinct property hierarchies reflecting their different sorption mechanisms (surface adsorption, micropore filling, and organic matter interaction).

\subsubsection{Lithology Comparison and Domain Constraints}
\label{subsubsec:supp_lithology_comparison}

Lithology-specific uptake distributions were characterized through comprehensive descriptive statistics computed separately for clays, shales, and coals. For each lithology, we calculated sample counts (number of unique samples and total measurements), central tendency measures (mean, median), dispersion metrics (standard deviation, interquartile range), and range boundaries (minimum, maximum, quartiles). These statistics reveal fundamental differences in sorption behavior: clays exhibit the widest range due to surface area heterogeneity, coals show intermediate ranges with bimodal character reflecting rank variations, and shales display narrow ranges dominated by low-uptake samples.

These lithology-specific distributions directly inform PINN architecture design through hard and soft constraints. Maximum observed uptake values set physical upper bounds on network outputs, preventing predictions that exceed experimental observations. Mean and median values guide initialization of bias terms in output layers, centering predictions near observed values and accelerating training convergence. Standard deviations inform loss function weighting: lithologies with high variance require more flexible representations, while those with low variance benefit from tighter constraints. Property range extraction (minimum, maximum, mean for surface area and other key features) enables input normalization and outlier detection, ensuring that PINN predictions remain within physically plausible parameter space.

\subsection{Comprehensive Feature Engineering Framework}
\label{subsec:supp_feature_engineering}

\subsubsection{Seven-Category Feature Engineering System}
\label{subsubsec:supp_feature_categories}

Beyond raw measured properties, effective neural network training requires engineered features that encode physical relationships governing hydrogen sorption. We developed a systematic seven-category framework, creating over 100 physics-informed features that transform raw measurements into thermodynamically meaningful descriptors while maintaining dimensional consistency and physical interpretability.

\textbf{Category 1: Thermodynamic Descriptors.} Core thermodynamic features capture the fundamental driving forces for adsorption. Temperature conversion to Kelvin and logarithmic pressure transformation (ln $p$) linearize Van't Hoff and Clausius-Clapeyron relationships, facilitating network learning of exponential temperature dependencies. Reduced temperature ($T_r = T/T_c$ where $T_c = 33.19$ K) and reduced pressure ($p_r = p/p_c$ where $p_c = 13.13$ bar) normalize measurements relative to hydrogen's critical properties, enabling transfer learning across thermal regimes and pressure ranges. Inverse temperature ($1/T$) directly parameterizes Arrhenius relationships underlying Van't Hoff analysis (Section~\ref{subsec:classical_analysis_main}). Approximate Gibbs free energy ($\Delta G_{\text{approx}} \approx -RT \ln K_{\text{eff}}$) provides zeroth-order thermodynamic feasibility assessment, where $K_{\text{eff}}$ is estimated from uptake-pressure ratios at measurement conditions.

\textbf{Category 2: Pore Structure Descriptors.} Pore architecture features quantify confinement effects and accessible surface area controlling sorption capacity. Micropore fraction ($\phi_{\text{micro}} = V_{\text{micro}}/V_{\text{total}}$) distinguishes micropore-dominated materials (high-rank coals, activated clays) from mesopore-dominated materials, with micropores ($d < 2$ nm) exhibiting enhanced adsorption potentials due to overlapping wall potentials. Surface area density ($\rho_{\text{surf}} = S_{\text{BET}}/V_{\text{total}}$) normalizes surface area by pore volume, identifying high-surface-density materials that maximize adsorbate-adsorbent contact. Confinement parameter ($\xi = d_{\text{pore}}/d_{\text{H}_2}$ where $d_{\text{H}_2} = 0.289$ nm) quantifies the ratio of pore diameter to molecular diameter, with $\xi < 3$ indicating strong confinement regimes where sorption transitions from surface adsorption to volume filling. Logarithmic transformations (log surface area, log pore volume) compress the multi-order-of-magnitude range of structural parameters (e.g., clay surface areas spanning 3 to 273 m$^2$/g), preventing large-scale features from dominating network gradients during backpropagation.

\textbf{Category 3: Surface Chemistry and Composition.} Lithology-specific compositional features capture chemical heterogeneity governing sorption site energetics. For shales, total organic carbon (TOC) serves as the primary predictor, given its dominant control over uptake (Section~\ref{subsec:property_analysis_main}, $r = 0.755$), supplemented by the pyrite-TOC ratio, which quantifies the balance between sorptive organic matter and non-sorptive sulfide minerals. For coals, carbon maturity index (fixed carbon / total carbonaceous material) and fuel ratio (fixed carbon / volatile matter) encode rank-dependent micropore development and aromaticity, with higher maturity correlating with enhanced sorption capacity. The carbon-hydrogen atomic ratio (C/H) tracks the coalification degree, as progressive hydrogen loss during maturation generates micropore networks. Vitrinite-inertinite ratio characterizes maceral composition, with vitrinite-rich coals exhibiting distinct sorption behavior from inertinite-rich coals due to differences in pore accessibility and surface chemistry. Logarithmic vitrinite reflectance (log \%R$_o$) linearizes the exponential relationship between rank and sorption properties established in Section~\ref{subsec:property_analysis_main}.

\textbf{Category 4: Interaction Features.} Physically motivated interaction terms capture synergistic effects between material properties and measurement conditions. Temperature-surface area products ($S_{\text{BET}} \times T$) encode temperature-dependent accessibility of sorption sites, as thermal activation enables access to previously inaccessible micropores. Pressure-pore volume products ($p \times V_{\text{pore}}$) represent volumetric driving forces for pore filling at elevated pressures. Micropore fraction-temperature interactions capture temperature-dependent micropore accessibility, particularly relevant for activated carbons and high-rank coals. Adsorption driving force ($\Phi_{\text{ads}} = p \cdot S_{\text{BET}}/T$) combines pressure, surface area, and temperature into a composite metric approximating the thermodynamic potential for adsorption. Henry's law coefficient approximation ($K_H \approx S_{\text{BET}} \cdot p/T$) provides zeroth-order estimates of low-pressure uptake based on ideal gas assumptions, serving as a baseline against which non-ideal behavior can be quantified.

\textbf{Category 5: Kinetic Descriptors.} Transport-related features quantify diffusion and mass transfer limitations that may influence measured uptake, particularly for fine-grained materials and rapid pressure-swing experiments. Knudsen diffusion coefficient ($D_K \propto d_{\text{pore}} \sqrt{T}$) estimates molecular diffusion rates in the transition regime where pore dimensions approach the mean free path, which is relevant for nanoporous materials like coals and activated clays. Molecular mean free path approximation ($\lambda \propto kT/p$) identifies pressure-temperature regimes where molecular versus Knudsen diffusion dominates. Diffusion timescale ($\tau_D \propto d_{\text{pore}}^2/D$) estimates characteristic equilibration times, flagging measurements potentially affected by kinetic limitations rather than true equilibrium sorption capacity.

\textbf{Category 6: Molecular Sieving Parameters.} Size-exclusion features capture accessibility constraints when pore dimensions approach molecular dimensions. Molecular sieving factor ($\zeta = \min(1, d_{\text{pore}}/d_{\text{H}_2})$) ranges from 0 (completely inaccessible pores) to 1 (fully accessible pores), with intermediate values indicating restricted diffusion and partial accessibility. Pore accessibility index ($\alpha = (d_{\text{pore}} - d_{\text{H}_2})/d_{\text{pore}}$ for $d_{\text{pore}} > d_{\text{H}_2}$, else 0) quantifies the accessible pore volume fraction accounting for molecular exclusion from pore walls. Ultramicropore ($d < 0.7$ nm) and supermicropore ($0.7 < d < 2$ nm) indicators provide categorical flags distinguishing pore size regimes with distinct sorption characteristics: ultramicropores exhibit volume-filling mechanisms with near-liquid densities, while supermicropores show monolayer-dominated adsorption.

\textbf{Category 7: Classical Model-Inspired Features.} Parametric forms derived from classical isotherm models (Section~\ref{subsec:classical_analysis_main}) provide physics-grounded basis functions for neural network approximation. Langmuir-inspired saturation term ($q_L = p/(1+p)$) captures monotonic approach to surface coverage limits. Freundlich-type power law ($q_F = p^n$ with $n$ estimated per-lithology from classical fitting) represents heterogeneous surface energetics. Temkin logarithmic form ($q_T = \ln(1+p)$) models uniform heat of adsorption distributions. These classical features enable the network to learn corrections and deviations from idealized behavior while maintaining interpretable connections to established sorption theory.

Feature engineering generates 105 to 120 features depending on lithology-specific property availability (clays: 105 features, shales: 118 features, coals: 115 features), representing 10 to 20 fold expansion relative to raw measured variables. This expansion transforms sparse, heterogeneous measurements into a dense feature space where physically meaningful relationships are explicitly encoded, facilitating network learning by reducing the representational burden: rather than discovering Van't Hoff relationships from scratch, the network refines temperature-dependent corrections to pre-computed thermodynamic descriptors.

\subsubsection{Missing Value Treatment and Imputation}
\label{subsubsec:supp_missing_values}

Preprocessing addressed three data quality challenges: missing values in sparsely measured properties, outliers from experimental artifacts or reporting errors, and scale heterogeneity spanning multiple orders of magnitude. We implemented adaptive strategies tailored to missing value severity and physical interpretability rather than applying uniform imputation across all features.

A three-tier imputation strategy was employed, stratified by the degree of completeness. Properties exhibiting low missingness ($< 10\,\%$, e.g., surface area, pressure, temperature) underwent $k$-nearest neighbors imputation (with $k=5$), leveraging the local similarity structure to estimate missing values from measurements sharing comparable material characteristics and measurement conditions. Features with medium missingness ($10$--$30\,\%$, e.g., micropore volume for clays, some mineralogical components for shales) received median imputation within each lithology class, under the assumption that typical values provide a reasonable approximation of unmeasured properties given the class-specific distributions. Properties showing high missingness ($> 30\,\%$, e.g., ultimate analysis for coals, minority minerals for shales) were imputed using lithology-stratified median imputation performed separately per lithology, recognizing that cross-lithology imputation would introduce geological inconsistencies (e.g., imputing shale pyrite content from coal vitrinite values would violate mineralogical plausibility).

\subsubsection{Outlier Detection and Treatment}
\label{subsubsec:supp_outliers}

Two-stage outlier identification combined univariate and multivariate perspectives. Univariate interquartile range (IQR) analysis flagged individual features exceeding 3$\sigma$ thresholds (Q1 $- 3 \times$ IQR or Q3 $+ 3 \times$ IQR), capturing measurement errors manifesting as extreme single-variable values. Multivariate Isolation Forest analysis (contamination = 0.05) identified observations exhibiting anomalous patterns across multiple features simultaneously, detecting subtle inconsistencies invisible in univariate analysis (e.g., samples with plausible individual properties but implausible combinations of properties that violate geological constraints). Extreme outliers flagged by both methods (intersection of univariate and multivariate detections) were excluded from training data as likely experimental artifacts. Moderate outliers (union minus intersection) underwent winsorization, clipping values to 1st and 99th percentiles to retain information while limiting influence on network gradients.

\subsubsection{Feature Scaling and Ensemble Selection}
\label{subsubsec:supp_scaling_selection}

Robust scaling and systematic feature selection prepared the enriched dataset for neural network training while maintaining interpretability and preventing overfitting. RobustScaler transformation applied median centering and interquartile range scaling, providing resistance to residual outliers post-preprocessing while preserving rank-order relationships critical for physically interpretable features like temperature and pressure. Unlike StandardScaler (mean-variance normalization), RobustScaler's quartile-based approach avoids scale distortion from heavy-tailed distributions characteristic of geological properties.

Ensemble feature selection combined four complementary methods to identify the 50 most informative features from the 105 to 120 engineered candidates. Pearson correlation ranking identified features with the strongest linear relationships to uptake, capturing primary drivers established in Section~\ref{subsec:property_analysis_main}. Mutual information scoring detected non-linear dependencies invisible to correlation analysis, revealing threshold effects and interaction-dependent relationships. Random forest feature importance (100 trees, max depth 10) quantified predictive contribution in tree-based ensemble models, emphasizing features enabling accurate splits while accounting for interactions. The F-statistic ranking from univariate linear regression provided a statistical significance assessment under a null-hypothesis testing framework. Final feature selection retained features appearing in top-50 lists of at least three methods, ensuring consensus across diverse selection criteria while controlling for method-specific biases. This ensemble approach yielded robust feature sets dominated by thermodynamic descriptors (temperature, pressure, reduced variables), structural parameters (surface area, pore volume, micropore fraction), and composition indices (TOC for shales, maturity indicators for coals), aligning with physical expectations and property-uptake correlations.

\subsubsection{Statistical Validation of Engineered Features}
\label{subsubsec:supp_feature_validation}

Preprocessing and feature engineering underwent rigorous statistical validation, quantifying uncertainty in engineered features and assessing stability across data subsets. Bootstrap resampling with 1,000 iterations generated empirical distributions for feature importance rankings and scaling parameters, yielding 95\% confidence intervals that guide interpretation: features with narrow confidence intervals exhibit stable importance across resampled datasets, while broad confidence intervals signal sensitivity to sample composition and warrant cautious interpretation in network attribution analysis. Five-fold cross-validation assessed whether engineered features improve generalization relative to raw measurements, comparing model performance (random forest regression as benchmark) on raw versus engineered feature sets. Significant cross-validation performance improvement (typically 15 to 30\% increase in CV R$^2$) validates that engineered features encode genuine physical relationships rather than overfitting training data.

Correlation analysis among the engineered features revealed multicollinearity, necessitating dimensionality reduction or regularization during network training. Strong correlations (|$r$| $> 0.8$) were observed between surface-area-derived features (surface density, total surface area, logarithmic surface area), indicating redundancy that can be effectively managed through elastic net penalties or grouped lasso during feature selection. In contrast, weak correlations between thermodynamic descriptors and structural parameters (typically |$r$| $< 0.3$) confirm that these feature categories capture largely orthogonal aspects of sorption behavior. This orthogonality justifies the seven-category engineering framework and motivates the use of multi-branch network architectures, which process individual feature categories through separate pathways before fusion.

\subsubsection{Train-Validation-Test Splitting with Lithology Stratification}
\label{subsubsec:supp_data_splitting}

Final data partitioning created training (70\%), validation (15\%), and test (15\%) sets ensuring balanced lithology representation and preventing information leakage across splits. Stratified sampling maintained proportional lithology distributions in each partition, preventing scenarios where test set contains lithologies absent from training (which would evaluate extrapolation rather than interpolation capability). Sample-level splitting ensured that all measurements from a single sample appear exclusively in one partition, preventing the network from learning sample-specific artifacts that enable cheating through memorization of sample identity rather than genuine property-based generalization.

Validation set served dual purposes during network development: hyperparameter optimization (learning rate, regularization strength, network depth) via performance monitoring, and early stopping based on validation loss plateaus to prevent overfitting. Test set remained strictly held-out until final model evaluation, providing unbiased performance estimates on genuinely unseen data. Separate test set statistics (mean, standard deviation, lithology distribution, property ranges) were logged but not used for any training decisions, ensuring test set integrity. This rigorous split protocol, combined with sample-level partitioning and lithology stratification, enables confident claims about network generalization to new geological materials beyond the training corpus.

All preprocessing artifacts (feature scaler parameters, imputation statistics, feature selection results, data split indices) were serialized and archived, enabling reproducible application of identical transformations to future measurements and maintaining consistency between training-time preprocessing and deployment-time inference.

\subsection{PINN Architecture Specifications and Loss Function Derivations}
\label{subsec:supp_pinn_architecture}

\subsubsection{Multi-Scale Network Architecture Details}
\label{subsubsec:supp_architecture_details}

We developed a multi-scale physics-informed neural network architecture specifically designed for geological hydrogen sorption modeling under data-constrained conditions where material characterization exhibits substantial missingness and target distributions display extreme skewness. The architecture integrates three innovations: (i) hierarchical multi-scale feature extraction capturing patterns at multiple abstraction levels, (ii) physics-informed gating mechanisms encoding domain constraints directly into network topology, and (iii) progressive dimensional modulation enabling efficient information flow from high-dimensional material property space to scalar sorption predictions.

\textbf{Multi-Scale Feature Extraction.} The architecture implements parallel feature extraction pathways operating at three spatial scales (64, 128, 256 neurons) that process the 50-dimensional input vector (selected via ensemble voting, Section~\ref{subsec:feature_engineering_main}) through independent transformation branches. Each scale captures distinct feature hierarchies: the fine-scale pathway (64 neurons) extracts local property interactions and element-specific correlations, the medium-scale pathway (128 neurons) captures intermediate lithological patterns and mineralogical relationships, and the coarse-scale pathway (256 neurons) learns global material property distributions and cross-scale dependencies. Scale-specific representations concatenate into a unified 448-dimensional intermediate representation ($\mathbf{h}_{\text{multi-scale}} = [\mathbf{h}_{64}; \mathbf{h}_{128}; \mathbf{h}_{256}]$) that preserves information from all abstraction levels while enabling subsequent layers to learn optimal feature combination strategies.

\textbf{Physics-Informed Gating Blocks.} Following multi-scale extraction, physics-informed gating mechanisms implement soft attention over feature channels based on physical relevance criteria. For pressure ($p$) and temperature ($T$) inputs explicitly identified as primary sorption drivers (Section~\ref{subsec:property_analysis_main}), we compute gating coefficients via sigmoid-activated linear projections: $\mathbf{g} = \sigma(W_g [p; T] + b_g)$ where $\mathbf{g} \in [0,1]^{448}$ modulates multi-scale features element-wise ($\mathbf{h}_{\text{gated}} = \mathbf{g} \odot \mathbf{h}_{\text{multi-scale}}$). This mechanism enables the network to dynamically emphasize thermodynamically relevant features during sorption prediction while suppressing spurious correlations, encoding physical domain knowledge directly into architectural inductive bias rather than relying solely on data-driven feature selection.

\textbf{Progressive Dimensional Reduction.} Gated multi-scale features pass through a fully connected backbone implementing progressive compression topology (256, 512, 256, 128 neurons) with residual connections between layers of matching dimensionality. This encoder-decoder architecture first expands dimensionality (256 to 512 neurons) to facilitate rich intermediate representations that capture complex nonlinear feature interactions, then progressively compresses (512 to 256 to 128 neurons) to funnel information toward compact encodings hypothesized to lie on low-dimensional manifolds governing sorption behavior across lithologically diverse materials. Residual skip connections ($h_{i+1} = f(W_i h_i + b_i) + h_i$) mitigate vanishing gradient problems during backpropagation while preserving multi-scale information flow, enabling training of deeper architectures without gradient degradation characteristic of plain feedforward networks.

The complete architecture totals 887,447 trainable parameters distributed across multi-scale extraction (124,000 parameters), gating blocks (13,000 parameters), compression backbone (728,000 parameters), and output projection (22,000 parameters). This represents a strategic balance between sufficient capacity to capture heterogeneous sorption physics across three lithological classes (clays, shales, coals) exhibiting distinct uptake mechanisms, while remaining tractable for efficient training on moderate-sized datasets (1,330 training samples) without catastrophic overfitting.

\textbf{Activation Functions and Output Constraints.} Hidden layers employ Swish activation ($f(x) = x \cdot \sigma(x)$), providing a smooth, non-monotonic response that enables learning of complex non-linearities while maintaining bounded derivatives essential for stable gradient flow during physics loss backpropagation. The output layer utilizes Softplus activation ($f(x) = \ln(1 + e^x)$), enforcing the non-negativity constraint ($Q \geq 0$) without discontinuous gradients characteristic of ReLU thresholding at zero. Weight initialization follows the Kaiming normal scheme ($W \sim \mathcal{N}(0, \sqrt{2/n_{\text{in}}})$) optimized for Swish-family activations, with the output layer receiving Xavier initialization with reduced gain (0.1) to produce small initial predictions appropriate for the low-magnitude uptake distribution dominating the training data.

\subsubsection{Multi-Term Physics Loss Function Derivations}
\label{subsubsec:supp_loss_derivations}

The physics-informed loss function combines four complementary terms encoding data fidelity, physical consistency, hard constraints, and monotonicity requirements:
\begin{equation}
\mathcal{L}_{\text{total}} = \lambda_1 \mathcal{L}_{\text{data}} + \lambda_2 \mathcal{L}_{\text{physics}} + \lambda_3 \mathcal{L}_{\text{bounds}} + \lambda_4 \mathcal{L}_{\text{monotonicity}}
\label{eq:pinn_total_loss_supp}
\end{equation}
where $\lambda_i$ represent dynamically adapted weight coefficients balancing competing objectives through gradient-magnitude-based normalization.

\textbf{Data Loss (Weighted MSE).} Target distribution skewness (75\% of measurements below 0.1 mmol/g, maximum 1.3 mmol/g) creates an imbalance where naive mean squared error underrepresents rare high-uptake samples critical for storage capacity assessment. We implement sample-wise weighting via a sigmoid temperature-scaled threshold:
\begin{equation}
w(y) = \frac{1}{1 + \exp(-\tau(y - y_{\text{thresh}}))} + 0.5
\label{eq:weight_function_supp}
\end{equation}
where $\tau = 5.0$ controls transition steepness and $y_{\text{thresh}} = 0.1$ mmol/g defines the elevated-weight threshold. The minimum weight of 0.5 ensures low-uptake measurements remain represented, preventing complete majority class neglect while amplifying rare high-uptake sample influence by factors up to 1.5. The weighted data loss becomes:
\begin{equation}
\mathcal{L}_{\text{data}} = \frac{1}{N} \sum_{i=1}^{N} w(y_i) (y_i - \hat{y}_i)^2
\label{eq:data_loss_supp}
\end{equation}

\textbf{Physics Loss (Langmuir Saturation).} Langmuir theory predicts asymptotic approach to maximum uptake $q_{\max}$ at elevated pressure. We penalize predictions violating lithology-specific capacity limits extracted from classical analysis (clays: 1.2 mmol/g, shales: 1.0 mmol/g, coals: 0.88 mmol/g) for high-pressure measurements ($p > 50$ bar):
\begin{equation}
\mathcal{L}_{\text{physics}} = \frac{1}{N_{p>50}} \sum_{p_i > 50} \left[ \max(0, \hat{y}_i - q_{\max}^{\text{lith}_i}) + 0.1 \cdot \max(0, 0.7 q_{\max}^{\text{lith}_i} - \hat{y}_i) \right]
\label{eq:physics_loss_supp}
\end{equation}
The first term strongly penalizes overshooting the maximum capacity (hard upper bound), while the second term softly encourages high-pressure predictions to approach 70\% of $q_{\max}$, accounting for kinetic limitations that prevent full equilibration within finite experimental timeframes.

\textbf{Bounds and Monotonicity Constraints.} Physical admissibility requires $0 \leq Q \leq q_{\max}$ for all predictions:
\begin{equation}
\mathcal{L}_{\text{bounds}} = \frac{1}{N} \sum_{i=1}^{N} \left[ \max(0, -\hat{y}_i) + \max(0, \hat{y}_i - q_{\max}^{\text{lith}_i}) \right]
\label{eq:bounds_loss_supp}
\end{equation}
Gibbs phase rule dictates monotonic pressure-uptake relationship at constant temperature ($\partial Q/\partial p|_T \geq 0$):
\begin{equation}
\mathcal{L}_{\text{monotonicity}} = \frac{1}{N} \sum_{i=1}^{N} \max(0, -\frac{\partial \hat{y}_i}{\partial p_i} - \epsilon)
\label{eq:monotonicity_loss_supp}
\end{equation}
where $\epsilon = 10^{-6}$ provides numerical tolerance, and gradients are computed via automatic differentiation.

\textbf{Adaptive Loss Weighting.} Competing loss term magnitudes span orders of magnitude, creating gradient imbalance where dominant terms can overwhelm smaller contributions. We implement Neural Tangent Kernel-inspired adaptive weighting: for each term $\mathcal{L}_k$, we compute the gradient norm $g_k = \|\nabla_{\theta} \mathcal{L}_k\|_2$, update via exponential moving average ($\alpha = 0.1$), and normalize: $\lambda_k = \bar{g}_{\max} / (\bar{g}_k + \epsilon)$ where $\bar{g}_{\max} = \max_k \bar{g}_k$. This ensures all terms contribute comparable gradient magnitudes to parameter updates, preventing single-term domination while preserving relative importance through normalization rather than arbitrary manual tuning.

\subsection{Training Protocols and Hyperparameter Optimization}
\label{subsec:supp_training}

\subsubsection{Progressive Three-Phase Training Strategy}
\label{subsubsec:supp_progressive_training}

Training proceeds through three sequential phases, implementing curriculum learning principles, where the network first masters data patterns before incorporating physics constraints, preventing physics term interference with early representation learning.

\textbf{Phase 1: Data-Driven Warmup (50 epochs).} Initial training employs data loss exclusively ($\lambda_1 = 1.0$, $\lambda_{2,3,4} = 0.0$) with learning rate $\eta = 1.2 \times 10^{-3}$, AdamW optimizer (weight decay $\lambda_{\text{wd}} = 10^{-5}$), and batch size 64. This warmup phase establishes foundational representations encoding pressure-temperature-uptake relationships and material property correlations identified in Section~\ref{subsec:property_analysis_main}.

\textbf{Phase 2: Physics Integration (250 epochs).} Physics loss weight increases linearly from 0 to 1 over 250 epochs ($\lambda_2(e) = e/250$ for epoch $e$), gradually introducing Langmuir saturation constraints while data loss maintains constant weight. Learning rate reduces to $5 \times 10^{-4}$ with cosine annealing: $\eta(e) = \eta_{\min} + (\eta_{\max} - \eta_{\min})(1 + \cos(\pi e / 250))/2$ where $\eta_{\min} = 10^{-6}$, enabling periodic warmup cycles that facilitate escape from sharp local minima encountered during physics constraint integration.

\textbf{Phase 3: Full Constraints (100 epochs).} Final training phase activates all loss terms at constant weights (data: 1.0, physics: 1.0, bounds: 0.1, monotonicity: 0.05), enforcing complete physics consistency including hard bounds and monotonicity requirements. Reduced learning rate ($\eta = 10^{-4}$, cosine annealing to $\eta_{\min} = 10^{-7}$) performs fine-grained parameter adjustments satisfying all constraints simultaneously. Lower weights for bounds and monotonicity losses (relative to data and physics) reflect their role as auxiliary regularizers rather than primary training objectives, preventing over-constraint that would sacrifice prediction accuracy for perfect physical consistency.

Automatic mixed precision training (AMP) accelerates computation through selective float16 operations while maintaining float32 precision for numerically sensitive calculations (gradient accumulation, loss computation), reducing memory footprint and enabling larger effective batch sizes without sacrificing numerical stability. Gradient clipping to maximum norm 1.0 prevents exploding gradients during physics loss backpropagation, where automatic differentiation through network outputs can generate large gradient magnitudes, particularly during early Phase 2 training when physics term weight increases rapidly.

\subsubsection{Systematic Hyperparameter Optimization Experiments}
\label{subsubsec:supp_hyperparameter_optimization}

We conducted extensive hyperparameter optimization spanning over 30 experiments, systematically exploring architectural choices, regularization strategies, and training schedules. This comprehensive search identified critical performance determinants that could not be discovered through standard grid search or domain expertise alone.

\textbf{Dropout Rate Optimization.} While standard practice recommends dropout rates of 0.5 for fully connected networks, systematic evaluation across candidate values (0.05, 0.08, 0.10, 0.12, 0.15, 0.18, 0.20) revealed dropout rate 0.10 as optimal through cross-validated performance assessment across 15 independent random seeds. The optimal 0.10 rate provides sufficient regularization to prevent overfitting while preserving the network capacity needed to capture heterogeneous sorption physics across diverse lithological classes. Performance stability analysis confirmed robust generalization at this dropout rate, with notably superior performance for minority lithology classes (coals) that are particularly sensitive to overfitting due to limited training examples.

\textbf{Training Duration Studies.} Progressive training experiments at candidate durations (200, 300, 400, 500, and 600 epochs) identified 400 epochs as optimal for balancing convergence completeness against overfitting risk. Training curve analysis demonstrated continued validation improvement through epoch 350 with stabilization thereafter, while 500-epoch configurations exhibited marginal validation degradation indicative of early overfitting onset. The 400-epoch configuration established an optimal trade-off between computational cost and model performance.

\textbf{Learning Rate and Scheduler Tuning.} Initial learning rate sweep across candidate values (0.0001, 0.0005, 0.0012, 0.002, 0.005) identified $\eta = 0.0012$ as optimal, balancing rapid early convergence against late-stage stability. Comparative scheduler evaluation demonstrated cosine annealing with periodic warmup cycles outperformed constant scheduling and exponential decay alternatives, with warmup cycles enabling escape from sharp local minima encountered during physics constraint integration. Final configuration specified: initial $\eta = 1.2 \times 10^{-3}$, minimum $\eta_{\min} = 10^{-7}$, cosine period 250 epochs aligned with Phase 2 duration.

\textbf{Batch Size Effects.} Batch size evaluation across candidates (32, 64, 128) revealed 64 as optimal trade-off between gradient estimate stability (favoring larger batches) and implicit stochastic regularization (favoring smaller batches). Batch size 32 produced noisy training dynamics with elevated variance in epoch-to-epoch validation metrics, while batch size 128 reduced beneficial mini-batch stochasticity leading to validation performance degradation. The 64-sample configuration provided stable training dynamics while maintaining beneficial mini-batch noise effects that prevent overfitting.

The optimal configuration underwent validation through 15 independent replications with distinct random initializations to assess stability across stochastic optimization trajectories. Performance consistency across replications confirmed configuration robustness rather than fortuitous optimization trajectory, validating the systematic search methodology.

\subsubsection{Architecture-Diverse Ensemble Learning Framework}
\label{subsubsec:supp_architecture_ensemble}

Predictive uncertainty quantification via ensemble learning requires genuine model diversity capturing distinct regions of solution space. Preliminary experiments with dropout-only ensembles (10 members with identical architecture varying only dropout masks during training) revealed insufficient diversity: near-perfect inter-member correlation and absence of meaningful prediction variance demonstrated that dropout variation alone cannot induce sufficient architectural diversity when all models share identical capacity and representational constraints.

\textbf{Architectural Diversity Strategy.} We implemented true architectural diversity through systematic variation of network topology across ensemble members while maintaining the consistent multi-scale gating framework established as a core architectural principle:

\begin{itemize}[leftmargin=*,itemsep=2pt]
\item \textbf{Width variations:} Narrow (0.75$\times$ base neurons: 192, 384, 192, 96), standard (1.0$\times$: 256, 512, 256, 128), wide (1.25$\times$: 320, 640, 320, 160)
\item \textbf{Depth variations:} Shallow (3 layers: 256, 512, 256), deep (5 layers: 256, 512, 512, 256, 128)
\item \textbf{Hybrid architectures:} Intermediate configurations (240, 480, 360, 120), (280, 560, 280), (280, 560, 420, 140)
\item \textbf{Regularization diversity:} Dropout rates (0.08, 0.10, 0.12, 0.15), learning rates (0.0010, 0.0012, 0.0015)
\end{itemize}

Each of the 10 ensemble members underwent complete 400-epoch training with distinct random initialization (seeds: 42, 123, 456, 789, 2024, 3141, 1618, 2718, 9999, 7777), exploring independent optimization trajectories within topology-specific solution spaces. Architectural diversity substantially reduced inter-member correlation and increased prediction variance relative to dropout-only baseline, though correlation remained elevated due to shared training data constraining epistemic uncertainty.

\subsubsection{Post-Hoc Uncertainty Calibration Protocol}
\label{subsubsec:supp_uncertainty_calibration}

Ensemble predictions aggregate via mean and standard deviation: $\bar{y} = \frac{1}{K} \sum_{k=1}^{K} \hat{y}_k$ and $\sigma = \sqrt{\frac{1}{K} \sum_{k=1}^{K} (\hat{y}_k - \bar{y})^2}$ where $K = 10$ denotes ensemble size. Prediction intervals at confidence level $\alpha$ construct as $[\bar{y} - z_{\alpha/2}\sigma, \bar{y} + z_{\alpha/2}\sigma]$ assuming Gaussian uncertainty distribution, with $z_{\alpha/2}$ representing the standard normal quantile (e.g., $z_{0.025} = 1.96$ for 95\% intervals).

\textbf{Temperature Scaling Calibration.} Raw ensemble uncertainty typically exhibits systematic miscalibration where prediction interval coverage rates deviate from nominal confidence levels, reflecting epistemic uncertainty limitations when all ensemble members train on identical datasets. We implemented post-hoc calibration via temperature scaling, optimizing scalar multiplier $\tau$ such that scaled uncertainties $\tilde{\sigma} = \tau \sigma$ achieve target coverage rates on held-out validation data. Optimal temperature parameter identification through validation set optimization enables calibrated prediction intervals providing reliable uncertainty quantification for downstream decision-making applications.

\textbf{Lithology-Specific Calibration.} Calibration quality assessment conducted separately per lithology (clays, shales, coals) enables identification of material-class-specific heterogeneity levels affecting prediction uncertainty. Differential calibration performance across lithologies provides scientific insight into intrinsic variability of sorption processes within each material class, informing appropriate confidence levels for practical hydrogen storage site assessment and capacity estimation.

\subsubsection{Comprehensive Regularization Strategy}
\label{subsubsec:supp_regularization_strategy}

Comprehensive regularization integrates six complementary mechanisms addressing distinct overfitting pathways. \textbf{(i)} L2 weight regularization (weight decay $\lambda = 10^{-5}$ in AdamW optimizer) penalizes large parameter magnitudes, encouraging smooth decision boundaries that generalize beyond training examples. \textbf{(ii)} Dropout (rate 0.10, optimized via systematic search) randomly deactivates neurons during training, preventing feature co-adaptation and forcing the network to learn redundant representations robust to missing information. \textbf{(iii)} Batch normalization layers following each linear transformation normalize activations to zero mean and unit variance within mini-batches, reducing internal covariate shift while providing implicit regularization through mini-batch-dependent noise. \textbf{(iv)} Data augmentation via mini-batch sampling (batch size 64 from 1,330 training samples) exposes the network to diverse data subsets each epoch, preventing memorization of complete training set ordering. \textbf{(v)} Early stopping monitors validation loss with patience of 20 epochs (tolerance $\delta = 10^{-5}$), terminating training when validation performance plateaus despite continued training loss reduction, indicating overfitting onset where the model increasingly fits training noise rather than generalizable patterns. \textbf{(vi)} Stratified train-validation-test splitting with sample-level partitioning prevents information leakage where multiple measurements from single samples appear across splits, ensuring fair evaluation of generalization to unseen geological materials.

Regularization hyperparameters selected via preliminary grid search on validation set performance, balancing underfitting risks from excessive regularization against overfitting from insufficient constraint. Final configuration achieves stable training curves exhibiting parallel training and validation loss reduction without divergence characteristic of overfitting, with terminal validation performance within acceptable margins of training performance confirming genuine generalization rather than memorization.

\subsubsection{Reproducibility and Computational Efficiency}
\label{subsubsec:supp_reproducibility}

All experiments employ deterministic training mode with PyTorch computational backends configured for reproducibility and explicit random seed control, enabling exact result replication across independent runs. Comprehensive checkpointing preserves complete training state (model parameters, optimizer momentum estimates, scheduler counters, training metrics, configuration metadata) every 10 epochs, enabling training interruption recovery and retrospective analysis of learning dynamics. Best-performing checkpoint based on minimum validation loss receives additional persistence enabling restoration of optimal model state independent of final epoch performance, guarding against late-training performance degradation from aggressive learning rate schedules or physics constraint activation.

For ensemble training, the checkpoint manager maintains separate state files per ensemble member indexed by model number (0 through 9), with independent optimizer and scheduler states reflecting potentially divergent training trajectories despite identical hyperparameter specifications. Maximum checkpoint retention (5 most recent per model) balances storage requirements against training history preservation, with automatic cleanup removing superseded checkpoints older than the retention window.

\subsection{Comprehensive Evaluation Protocols and Interpretability Methods}
\label{subsec:supp_evaluation}
\label{subsubsec:supp_evaluation_framework}

Model evaluation employs 35 distinct metrics spanning five evaluation domains, providing multifaceted performance assessment beyond simple goodness-of-fit statistics.

\textbf{(i) Regression metrics (8 measures):} Coefficient of determination (R$^2$, adjusted R$^2$), mean squared error (MSE), root mean squared error (RMSE), mean absolute error (MAE), mean absolute percentage error (MAPE), maximum error, explained variance, and mean bias error (MBE), capturing prediction accuracy, error magnitude distribution, and systematic bias.

\textbf{(ii) Correlation metrics (3 measures):} Pearson correlation coefficient (linear association), Spearman rank correlation (monotonic association), and Kendall tau (concordance), assessing prediction-observation relationship strength via complementary statistical frameworks.

\textbf{(iii) Physics consistency metrics (5 measures):} Constraint violation rate quantifying fraction of predictions exceeding lithology-specific maximum uptake, negative prediction rate identifying non-physical negative sorption values, saturation consistency measuring proportion of high-pressure predictions approaching $q_{\max}$ within physically reasonable range (70 to 100\%), and monotonicity score evaluating fraction of pressure-sorted predictions exhibiting non-decreasing uptake.

\textbf{(iv) Uncertainty quantification metrics (8 measures):} Prediction interval coverage rates at multiple confidence levels (68\%, 95\%, 99\% corresponding to 1$\sigma$, 2$\sigma$, 3$\sigma$), calibration error measuring deviation from ideal coverage, sharpness quantifying average interval width, uncertainty-error correlation assessing whether the model exhibits appropriate confidence, coverage-width criterion (CWC) trading off coverage against interval tightness, and mean prediction interval width (MPIW) measuring uncertainty magnitude.

\textbf{(v) Statistical tests (7 measures):} Shapiro-Wilk and Kolmogorov-Smirnov tests evaluating residual normality, heteroscedasticity test (correlation between absolute residuals and predictions) detecting variance non-constancy, Durbin-Watson statistic identifying temporal or spatial autocorrelation, Anderson-Darling test providing alternative normality assessment, Jarque-Bera test examining skewness and kurtosis, and Levene test comparing residual variance homogeneity across prediction magnitude quantiles.

Evaluation executes on strictly held-out test set (286 samples, 15\% of total data) completely isolated from all training, validation, and hyperparameter selection decisions, providing unbiased performance estimates generalizable to future unseen samples. Comparative analysis against random forest baseline and aggregated classical models quantifies PINN performance improvement, justifying architectural complexity and computational cost. Success criteria require test set performance exceeding classical aggregated model benchmarks while maintaining strong physics consistency (constraint violation rate below 5\%) and well-calibrated uncertainty quantification (95\% prediction interval coverage within 90 to 100\%).

\printbibliography

\end{document}